\documentclass[11pt]{article}

\usepackage[preprint]{acl}

\usepackage{blindtext}
\usepackage{hyperref}
\usepackage{enumitem}
\usepackage{times}
\usepackage{latexsym}
\usepackage{subcaption}
\usepackage{multirow}
\usepackage{multicol}
\usepackage{subcaption}

\usepackage[T1]{fontenc}

\usepackage[utf8]{inputenc}

\usepackage{microtype}

\usepackage{inconsolata}

\usepackage{graphicx}
\usepackage{booktabs}
\usepackage{mathtools} 
\usepackage{amsmath}
\usepackage{xcolor}
\usepackage{enumitem}
\usepackage[table]{xcolor}
\usepackage{colortbl}
\definecolor{cbBlue}{HTML}{0072B2}
\definecolor{cbOrange}{HTML}{D55E00}
\usepackage{tabularx}
\usepackage{array}

\title{Want Better Synthetic Data? Steer It: Activation Steering for Low-Resource Language Generation}

\author{Jan Cegin$^{\spadesuit}$, Daniil Gurgurov$^\dagger$, Yusser Al Ghussin$^\dagger$, Simon Ostermann$^{\dagger}$ \\
  $^{\spadesuit}$ Kempelen Institute of Intelligent Technologies, Bratislava, Slovakia\\
  \texttt{jan.cegin@kinit.sk} \\
  $^\dagger$ German Research Institute for Artificial Intelligence (DFKI), Saarbr\"{u}cken, Germany \\
    \texttt{\{daniil.gurgurov, yusser.al\_ghussin, simon.ostermann\}}@dfki.de \\}

\begin{document}
\maketitle
\begin{abstract}
Large language models (LLMs) have become an effective tool for synthetic data generation, including for low-resource languages, where generated data can improve downstream task performance. Current best-performing approaches typically rely on few-shot prompting with target-language examples, which increases inference costs and may reduce diversity through lexical anchoring. In this work, we investigate activation steering as an alternative for low-resource synthetic data generation. We study two steering strategies: \textit{Language} Steering, which targets the linguistic identity of a language, and \textit{Quality} Steering, which captures well-formedness by contrasting human-written and backtranslated text representations. We evaluate these methods across four open-source LLMs, multiple layers, and 11 typologically diverse languages by generating sentiment and topic classification data and finetuning smaller classifiers. Steering is applied in both zero-shot and few-shot prompting settings and compared against non-steered counterparts. Our results show that steering on early layers consistently improves the diversity of generated data while often yielding stronger downstream model performance, particularly for low-resource languages.
\end{abstract}

\section{Introduction}

With the emergence of LLMs, various studies have demonstrated their ability to generate label-adherent and well-formed text~\cite{cegin-etal-2023-chatgpt, ubani2023zeroshotdataaug}. These two capabilities have made them an ideal tool for data \textit{augmentation} and synthetic data \textit{generation}, as demonstrated across domains such as sentiment analysis~\cite{ONAN2023101611}, intent recognition~\cite{cegin2024effectsdiversityincentivessample}, and topic classification~\cite{piedboeuf-langlais-2023-chatgpt}. For synthetic data generation, the workflow usually consists of prompting an LLM to produce a set number of samples with given labels in a particular language. These are then often used for fine-tuning of downstream encoder models.

Recent works have extended these approaches to low-resource languages, often improving downstream model performance~\cite{anikina-etal-2025-rigorous, pranida2025syntheticdatagenerationculturally, cegin-etal-2026-rose}, where data scarcity is prominent~\cite{joshi2020state}. The strongest results generally come from few-shot prompting with examples from the target language~\cite{anikina-etal-2025-rigorous}. However, this increases inference costs and may reduce diversity through lexical anchoring.

In parallel, research on representation engineering or activation steering has shown that modifying internal model activations can steer LLM behavior toward desired semantic or stylistic properties~\cite{zou2023representation, turner2024steeringlanguagemodelsactivation}. By injecting steering vectors into hidden states, prior work has controlled attributes such as truthfulness~\cite{li2023inference}, sentiment~\cite{rimsky-etal-2024-steering}, and toxicity~\cite{turner2024steeringlanguagemodelsactivation} without additional fine-tuning. Compared to few-shot prompting, activation steering provides an efficient alternative that may better capture various abstract properties \cite{ostermann2026weights}. Additionally, it can be combined with any type of prompt, zero- or few-shot \cite{radevski2026compositional}.

\begin{figure}[t!]
    \centering
    \includegraphics[width=0.45\textwidth]{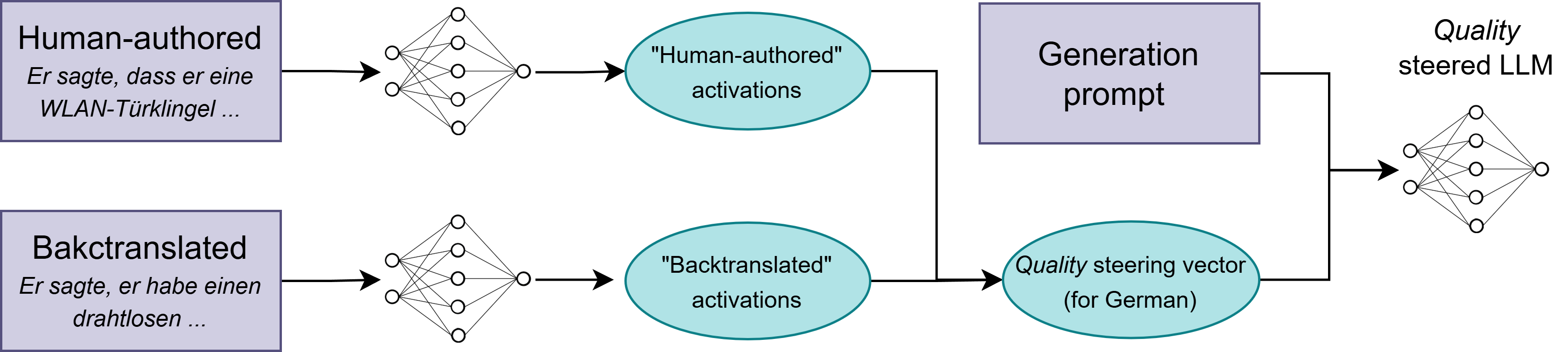}
    \caption{Example showing how \textit{Quality} steering vectors are created. This example shows a contrastive collection of activations for German, from which a steering vector is created. This steering vector is used before generating data from the target LLM and language.}
    \label{fig:steer_qual_example}
\end{figure}

In this paper, we investigate activation steering for low-resource synthetic data generation. We generate synthetic datasets for sentiment and topic classification tasks across 11 typologically diverse languages and evaluate them through downstream fine-tuning performance. We study two steering strategies: \textit{Language} Steering, which targets linguistic identity per previous studies~\cite{gurgurov2026clas,gurgurov-etal-2025-language, ghussin2026multilingualsteeringdesignmultilingual, ghussin2026dfkimltsemeval2026task7}, and \textit{Quality} Steering, which isolates well-formedness by contrasting human-written and backtranslated text. To our knowledge, this is the first work to derive and use \textit{Quality} steering vectors. Across four open-source LLMs and multiple layers, we apply these steering vectors to both zero- and few-shot prompts~\cite{anikina-etal-2025-rigorous} and analyze the resulting data in terms of diversity and representational properties, compared to no-steering baselines. Code can be found at~\url{https://github.com/kinit-sk/steering-synth-gen}.


Our main contributions are:
\begin{itemize}
    \item For the first time, we apply \textit{Language} steering vectors for synthetic data generation and introduce and utilize \textit{Quality} steering vectors by contrasting human-authored texts with backtranslated texts.
    \item  We analyze the cosine similarity between \textit{Quality} and \textit{Language} steering vectors, demonstrating that their alignment is highly LLM-dependent and most polarized in earlier layers. Crucially, the majority of evaluated languages display strong negative similarity, implying that, in general, \textit{Quality} steering vectors are not tied to a specific language. 
    \item We show that applying \textit{Quality} and \textit{Language} steering to early layers improves LLM performance in generating synthetic data, as measured by downstream performance, while increasing the diversity of the generated data for both zero- and few-shot prompting.
\end{itemize}

\section{Related Work}
\paragraph{Synthetic data.}
LLMs are increasingly being used to create semantically new samples adhering to a given label~\cite{ubani2023zeroshotdataaug, cegin2024userandomselectionnow}. LLM-based augmentation and data generation have been used for a variety of tasks such as automated scoring~\cite{fang2023using}, low-resource language generation~\cite{ghosh-etal-2023-dale}, intent classification~\cite{sahu-etal-2022-data}, sentiment analysis ~\cite{piedboeuf-langlais-2023-chatgpt, ONAN2023101611}, content recommendation~\cite{contect-based-recom}, and health symptoms classifications~\cite{dai2023auggpt}. As LLMs are better generators than classifiers of data in low-resource languages~\cite{pecher-etal-2026-better}, recent studies have focused on generating such label-adhering data in a variety of low-resource languages like Vietnamese~\cite{feng-etal-2021-survey}, Marathi~\cite{tran-etal-2026-representation}, or various African languages~\cite{belay2026afrilangtutoradvancinglanguagetutoring}, and for a variety of tasks and domains such as QA~\cite{namboori2023gemquad}, fact-checking~\cite{chung2025beyond}, NER~\cite{liu-etal-2021-mulda}, or text classification~\cite{glenn-etal-2023-jetsons}. A recent study has demonstrated that generating data rather than translating it leads to better model performance~\cite{pranida2025syntheticdatagenerationculturally}. Other studies focused on finding the best approaches for generating data in low-resource languages, such as enhancing generator selection~\cite{cegin-etal-2026-rose} or finding the best prompting techniques~\cite{anikina-etal-2025-rigorous}. As such, the current state-of-the-art technique~\cite{anikina-etal-2025-rigorous} uses few-shot examples in the prompt itself, leading to increased performance at the cost of additional inference.

\begin{figure*}[t!]
\centering
\begin{subfigure}{0.49\textwidth}
    \centering
    \includegraphics[width=\textwidth]{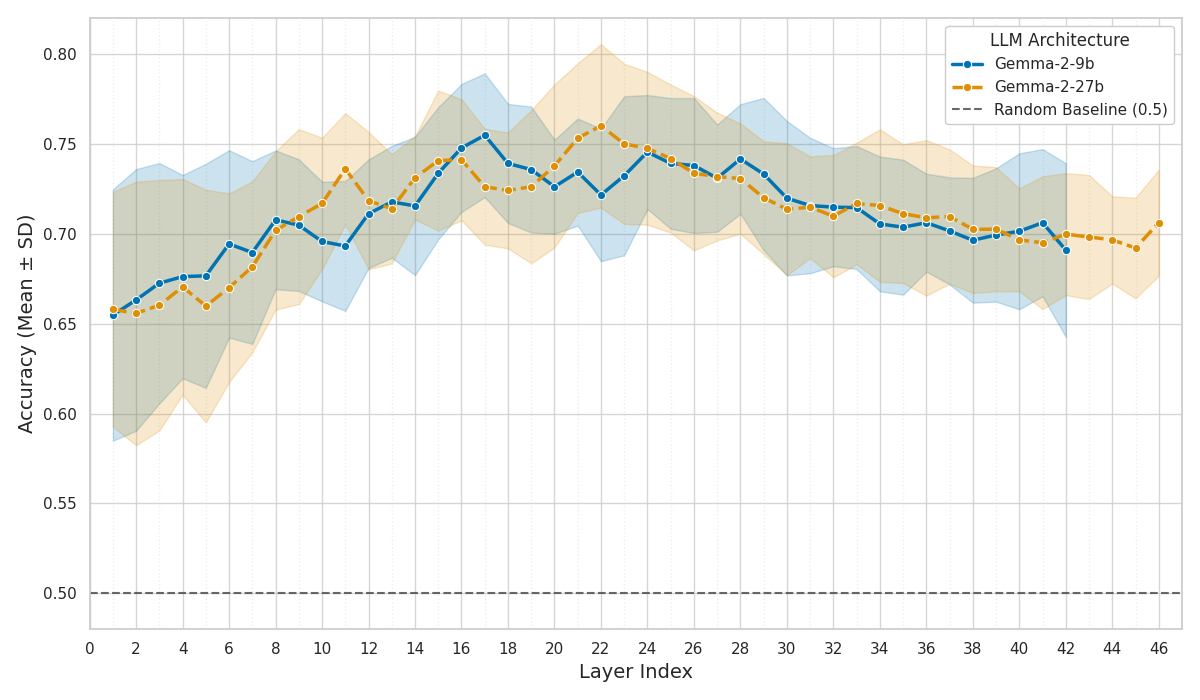}
    \caption{Gemma 2 model results.}
    \label{fig:multi_llm_probe_gemma}
\end{subfigure}
\hfill
\begin{subfigure}{0.49\textwidth}
    \centering
    \includegraphics[width=\textwidth]{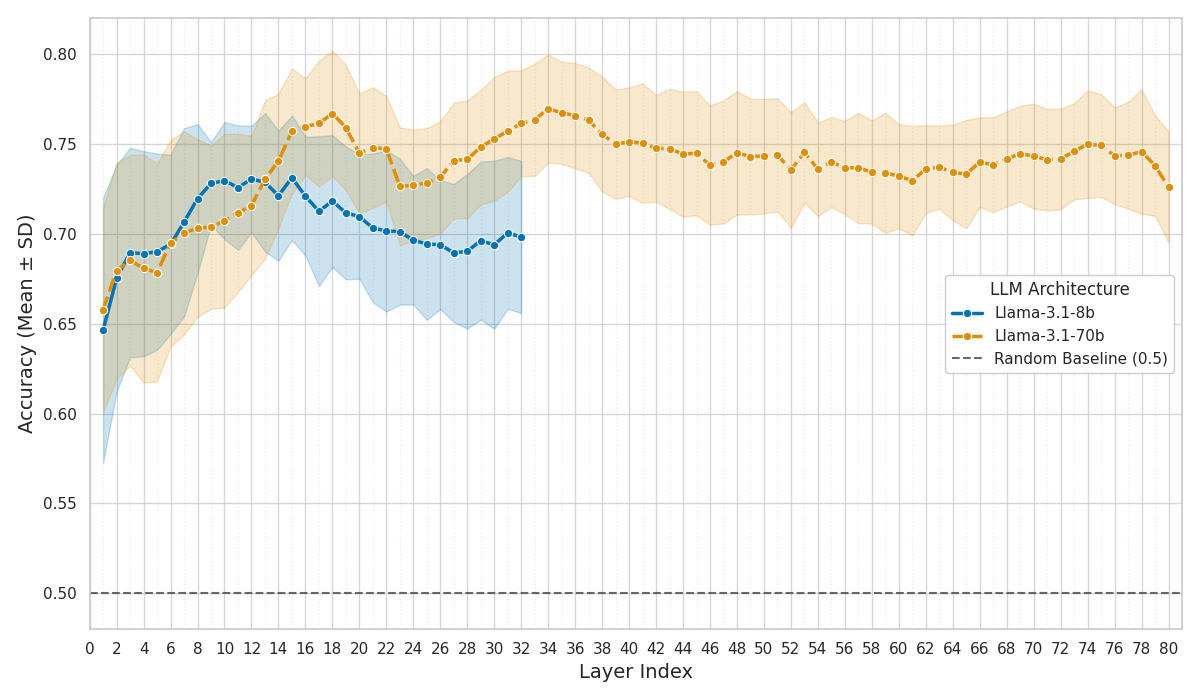}
    \caption{Llama 3.1 model results.}
    \label{fig:multi_llm_probe_llama}
\end{subfigure}
\caption{Linear probing results for human-authored vs. backtranslated textual pairs aggregated over languages.}
\label{fig:linear_probe_res}
\end{figure*}

\paragraph{Activation steering.}
A separate line of work has explored activation steering, where model behavior is modified by intervening directly on intermediate representations during the forward pass rather than through prompting or fine-tuning \cite{zou2023representation, turner2024steeringlanguagemodelsactivation, ostermann2026weights}. The most widely used family of methods is difference-based steering, which derives steering vectors from contrastive examples \cite{turner2024steeringlanguagemodelsactivation, rimsky-etal-2024-steering, marks2023geometry}. 
Beyond this, optimization-based approaches such as ReFT \citep{wu2024reft} and its rank-1 variant \citep{wuaxbench} learn low-rank interventions on hidden states, while dictionary-learning methods use sparse autoencoders to decompose activations into interpretable, individually steerable features \citep{gao2024scaling, huben2024sparse}. These techniques have been shown to reliably control a wide range of behaviors, such as sentiment, topic, and style \cite{turner2024steeringlanguagemodelsactivation, konen2024stylevectorssteeringgenerative}, truthfulness and sycophancy \cite{li2023inference, panickssery2023steering}, and refusal \cite{arditi2024refusal}. Particularly relevant to our setting, steering has also been applied to multilingual control: Previous work has identified language-specific neurons \cite{zhao2024large, tang2024language, gurgurov-etal-2025-language}, SAE features \citep{chou2025causallanguagecontrolmultilingual, deng2025unveilinglanguagespecificfeatureslarge}, and language vectors \citep{wong2026langfir, gurgurov2026clas, ghussin2026multilingualsteeringdesignmultilingual} that can be activated or ablated to change the output language while preserving semantics quality. Recent work has further shown that language-vector steering can be used for culturally aware multilingual inference: \citet{ghussin2026dfkimltsemeval2026task7} reports improvements for cultural knowledge retrieval.  While these studies establish that linguistic identity is encoded as a steerable direction in activation space, they focus on language control or downstream improvement on cultural benchmarks rather than the \textit{generation quality} of the produced text. Our work addresses this gap by using both language and quality steering vectors to produce better synthetic data for low-resource languages.

\section{Methodology and Experiments}
\label{sec:method_exp}
Our methodology consists of a simple procedure where we (1) collect activations from the residual stream from given layers and create the steering vectors from the collected activations, (2) apply the steering vector to the LLM, (3) generate data with labels in a given language, and (4) finetune a downstream model and evaluate it on human test data. This is done for 11 different languages.

These 11 languages in our study are selected to cover diverse typological properties and ensure overlap with the evaluation datasets. They span Indo-European (Germanic: Danish, German; Slavic: Czech, Slovak, Slovenian), Afro-Asiatic (Semitic: Amharic, Hebrew, Maltese), and Austronesian (Indonesian, Javanese, Sundanese) families \cite{nordhoff2011glottolog}. This design allows us to test steering robustness across varied morphological systems, from agglutinative to fusional structures, as well as across different writing systems (Ge'ez, Hebrew, Latin). This diversity supports evaluating whether the learned steering manifold generalizes beyond orthographic form.

\subsection{Computing Steering Vectors}
\paragraph{Language Vectors.} We use the FLORES~\cite{nllb-24} and BOUQUET~\cite{andrews-etal-2025-bouquet} datasets, both of which contain human-authored multilingual texts. FLORES provides professionally translated parallel sentences across all 11 target languages listed in Appendix~\ref{sec:lang_abbreviations}, enabling control over semantic content while isolating linguistic identity \cite{ghussin2026dfkimltsemeval2026task7}. BOUQUET is added to increase lexical diversity and include more naturalistic language. Concatenating both datasets also ensures sufficient token coverage for stable mean activations. 

To construct \textit{Language} steering vectors, we employ a one-vs-rest contrastive approach over the residual stream activations of all transformer blocks \cite{ghussin2026dfkimltsemeval2026task7}. For each of the 11 target languages, we compute a mean activation vector from the dataset by averaging across all non-padding tokens, yielding a language-specific mean representation in the model’s latent space for every layer. The steering vector is then defined as the difference between the target language mean representation and the mean representation of all remaining languages, followed by normalization.

\begin{table*}
\caption{Mean F1 difference across languages, models, steering methods, and layers aggregated over tasks, with \textcolor{cbBlue}{positive} and \textcolor{cbOrange}{negative} differences compared to a \textbf{zero-shot prompt} with no steering. The best alpha values and results were selected based on validation set results per layer, task, and language combination.}
\label{tab:steering_results}
\tiny
\resizebox{\textwidth}{!}{%
\begin{tabular}{lllccccccccccc||c}
\toprule
Method & LLM & Layer & am & cs & da & de & he & id & jv & mt & sk & sl & su & Average \\
\midrule
\midrule
\multirow[c]{12}{*}{\textit{Language}} & \multirow[c]{3}{*}{Gemma 2 27b} & L10 & {\cellcolor{cbBlue!24}} 2.74 & {\cellcolor{cbBlue!15}} 0.97 & {\cellcolor{cbOrange!13}} -0.65 & {\cellcolor{cbBlue!29}} 3.71 & {\cellcolor{cbOrange!11}} -0.30 & {\cellcolor{cbBlue!23}} 2.71 & {\cellcolor{cbBlue!19}} 1.85 & {\cellcolor{cbBlue!12}} 0.45 & {\cellcolor{cbBlue!29}} 3.70 & {\cellcolor{cbBlue!25}} 3.06 & {\cellcolor{cbBlue!45}} 6.86 & {\cellcolor{cbBlue!21}} 2.28 \\
 &  & L22 & {\cellcolor{cbBlue!26}} 3.22 & {\cellcolor{cbBlue!13}} 0.75 & {\cellcolor{cbOrange!10}} -0.08 & {\cellcolor{cbBlue!15}} 1.01 & {\cellcolor{cbBlue!11}} 0.36 & {\cellcolor{cbBlue!19}} 1.76 & {\cellcolor{cbOrange!11}} -0.33 & {\cellcolor{cbBlue!15}} 1.16 & {\cellcolor{cbBlue!30}} 3.90 & {\cellcolor{cbBlue!20}} 2.01 & {\cellcolor{cbBlue!36}} 5.11 & {\cellcolor{cbBlue!18}} 1.72 \\
 &  & L34 & {\cellcolor{cbBlue!32}} 4.34 & {\cellcolor{cbBlue!14}} 0.86 & {\cellcolor{cbBlue!11}} 0.23 & {\cellcolor{cbBlue!13}} 0.71 & {\cellcolor{cbBlue!14}} 0.83 & {\cellcolor{cbBlue!23}} 2.67 & {\cellcolor{cbBlue!24}} 2.78 & {\cellcolor{cbOrange!15}} -1.10 & {\cellcolor{cbBlue!26}} 3.24 & {\cellcolor{cbBlue!25}} 3.06 & {\cellcolor{cbBlue!41}} 6.20 & {\cellcolor{cbBlue!21}} 2.17 \\
 \cmidrule(lr){2-15}
 & \multirow[c]{3}{*}{Gemma 2 9b} & L9 & {\cellcolor{cbOrange!19}} -1.91 & {\cellcolor{cbBlue!32}} 4.28 & {\cellcolor{cbBlue!20}} 2.07 & {\cellcolor{cbBlue!20}} 1.99 & {\cellcolor{cbBlue!39}} 5.80 & {\cellcolor{cbBlue!11}} 0.28 & {\cellcolor{cbOrange!16}} -1.26 & {\cellcolor{cbBlue!28}} 3.64 & {\cellcolor{cbBlue!44}} 6.71 & {\cellcolor{cbBlue!59}} 9.51 & {\cellcolor{cbBlue!23}} 2.61 & {\cellcolor{cbBlue!25}} 3.07 \\
 &  & L20 & {\cellcolor{cbOrange!18}} -1.58 & {\cellcolor{cbBlue!28}} 3.59 & {\cellcolor{cbBlue!22}} 2.41 & {\cellcolor{cbBlue!12}} 0.53 & {\cellcolor{cbBlue!23}} 2.62 & {\cellcolor{cbBlue!14}} 0.88 & {\cellcolor{cbBlue!11}} 0.21 & {\cellcolor{cbBlue!27}} 3.38 & {\cellcolor{cbBlue!51}} 8.02 & {\cellcolor{cbBlue!40}} 5.98 & {\cellcolor{cbBlue!30}} 3.96 & {\cellcolor{cbBlue!24}} 2.73 \\
 &  & L31 & {\cellcolor{cbBlue!26}} 3.22 & {\cellcolor{cbBlue!18}} 1.71 & {\cellcolor{cbBlue!21}} 2.17 & {\cellcolor{cbBlue!17}} 1.49 & {\cellcolor{cbBlue!25}} 2.99 & {\cellcolor{cbBlue!13}} 0.70 & {\cellcolor{cbBlue!12}} 0.40 & {\cellcolor{cbBlue!25}} 2.96 & {\cellcolor{cbBlue!43}} 6.56 & {\cellcolor{cbBlue!42}} 6.27 & {\cellcolor{cbBlue!22}} 2.52 & {\cellcolor{cbBlue!24}} 2.82 \\
 \cmidrule(lr){2-15}
 & \multirow[c]{3}{*}{Llama3.1 70b} & L17 & {\cellcolor{cbBlue!27}} 3.47 & {\cellcolor{cbOrange!13}} -0.62 & {\cellcolor{cbBlue!16}} 1.27 & {\cellcolor{cbBlue!23}} 2.62 & {\cellcolor{cbBlue!16}} 1.32 & {\cellcolor{cbBlue!24}} 2.74 & {\cellcolor{cbBlue!28}} 3.53 & {\cellcolor{cbBlue!19}} 1.91 & {\cellcolor{cbBlue!60}} 9.83 & {\cellcolor{cbBlue!27}} 3.33 & {\cellcolor{cbOrange!12}} -0.46 & {\cellcolor{cbBlue!23}} 2.63 \\
 &  & L37 & {\cellcolor{cbBlue!33}} 4.65 & {\cellcolor{cbBlue!16}} 1.29 & {\cellcolor{cbBlue!19}} 1.88 & {\cellcolor{cbBlue!28}} 3.61 & {\cellcolor{cbBlue!22}} 2.40 & {\cellcolor{cbBlue!16}} 1.35 & {\cellcolor{cbBlue!37}} 5.43 & {\cellcolor{cbBlue!12}} 0.53 & {\cellcolor{cbBlue!47}} 7.35 & {\cellcolor{cbBlue!22}} 2.48 & {\cellcolor{cbBlue!25}} 2.96 & {\cellcolor{cbBlue!25}} 3.08 \\
 &  & L57 & {\cellcolor{cbBlue!27}} 3.43 & {\cellcolor{cbBlue!11}} 0.34 & {\cellcolor{cbBlue!19}} 1.77 & {\cellcolor{cbBlue!23}} 2.61 & {\cellcolor{cbBlue!19}} 1.82 & {\cellcolor{cbBlue!22}} 2.52 & {\cellcolor{cbBlue!31}} 4.19 & {\cellcolor{cbBlue!17}} 1.49 & {\cellcolor{cbBlue!30}} 3.92 & {\cellcolor{cbBlue!29}} 3.78 & {\cellcolor{cbBlue!33}} 4.60 & {\cellcolor{cbBlue!24}} 2.77 \\
 \cmidrule(lr){2-15}
 & \multirow[c]{3}{*}{Llama3.1 8b} & L7 & {\cellcolor{cbBlue!53}} 8.50 & {\cellcolor{cbBlue!13}} 0.74 & {\cellcolor{cbBlue!16}} 1.18 & {\cellcolor{cbBlue!17}} 1.48 & {\cellcolor{cbBlue!14}} 0.81 & {\cellcolor{cbBlue!22}} 2.47 & {\cellcolor{cbBlue!22}} 2.47 & {\cellcolor{cbBlue!33}} 4.49 & {\cellcolor{cbBlue!30}} 4.01 & {\cellcolor{cbBlue!30}} 4.01 & {\cellcolor{cbOrange!10}} -0.14 & {\cellcolor{cbBlue!24}} 2.73 \\
 &  & L15 & {\cellcolor{cbBlue!57}} 9.28 & {\cellcolor{cbBlue!25}} 2.94 & {\cellcolor{cbBlue!27}} 3.40 & {\cellcolor{cbBlue!21}} 2.27 & {\cellcolor{cbOrange!10}} -0.05 & {\cellcolor{cbBlue!34}} 4.78 & {\cellcolor{cbBlue!14}} 0.81 & {\cellcolor{cbBlue!31}} 4.25 & {\cellcolor{cbBlue!29}} 3.84 & {\cellcolor{cbBlue!33}} 4.59 & {\cellcolor{cbOrange!16}} -1.17 & {\cellcolor{cbBlue!26}} 3.18 \\
 &  & L23 & {\cellcolor{cbBlue!64}} 10.56 & {\cellcolor{cbBlue!21}} 2.20 & {\cellcolor{cbBlue!26}} 3.16 & {\cellcolor{cbOrange!13}} -0.63 & {\cellcolor{cbOrange!15}} -1.13 & {\cellcolor{cbBlue!19}} 1.77 & {\cellcolor{cbOrange!10}} -0.08 & {\cellcolor{cbBlue!25}} 3.08 & {\cellcolor{cbBlue!35}} 4.86 & {\cellcolor{cbBlue!23}} 2.66 & {\cellcolor{cbBlue!10}} 0.14 & {\cellcolor{cbBlue!22}} 2.42 \\
 \midrule
 \midrule
\multirow[c]{12}{*}{\textit{Quality}} & \multirow[c]{3}{*}{Gemma 2 27b} & L10 & {\cellcolor{cbBlue!25}} 3.01 & {\cellcolor{cbBlue!14}} 0.86 & {\cellcolor{cbOrange!10}} -0.02 & {\cellcolor{cbBlue!23}} 2.71 & {\cellcolor{cbBlue!14}} 0.93 & {\cellcolor{cbBlue!21}} 2.30 & {\cellcolor{cbBlue!17}} 1.38 & {\cellcolor{cbBlue!17}} 1.40 & {\cellcolor{cbBlue!25}} 3.02 & {\cellcolor{cbBlue!30}} 3.92 & {\cellcolor{cbBlue!39}} 5.65 & {\cellcolor{cbBlue!21}} 2.29 \\
 &  & L22 & {\cellcolor{cbBlue!20}} 2.06 & {\cellcolor{cbBlue!12}} 0.56 & {\cellcolor{cbOrange!12}} -0.55 & {\cellcolor{cbBlue!15}} 0.98 & {\cellcolor{cbBlue!16}} 1.21 & {\cellcolor{cbBlue!22}} 2.36 & {\cellcolor{cbBlue!19}} 1.79 & {\cellcolor{cbBlue!18}} 1.65 & {\cellcolor{cbBlue!28}} 3.57 & {\cellcolor{cbBlue!27}} 3.49 & {\cellcolor{cbBlue!37}} 5.35 & {\cellcolor{cbBlue!20}} 2.04 \\
 &  & L34 & {\cellcolor{cbBlue!24}} 2.82 & {\cellcolor{cbBlue!11}} 0.26 & {\cellcolor{cbOrange!12}} -0.54 & {\cellcolor{cbBlue!17}} 1.45 & {\cellcolor{cbBlue!11}} 0.20 & {\cellcolor{cbBlue!21}} 2.31 & {\cellcolor{cbBlue!27}} 3.44 & {\cellcolor{cbBlue!11}} 0.28 & {\cellcolor{cbBlue!34}} 4.68 & {\cellcolor{cbBlue!24}} 2.77 & {\cellcolor{cbBlue!32}} 4.28 & {\cellcolor{cbBlue!20}} 2.00 \\
 \cmidrule(lr){2-15}
 & \multirow[c]{3}{*}{Gemma 2 9b} & L9 & {\cellcolor{cbBlue!11}} 0.34 & {\cellcolor{cbBlue!51}} 8.04 & {\cellcolor{cbBlue!21}} 2.19 & {\cellcolor{cbBlue!30}} 3.90 & {\cellcolor{cbBlue!33}} 4.63 & {\cellcolor{cbBlue!19}} 1.83 & {\cellcolor{cbBlue!11}} 0.26 & {\cellcolor{cbBlue!19}} 1.76 & {\cellcolor{cbBlue!70}} 11.77 & {\cellcolor{cbBlue!62}} 10.25 & {\cellcolor{cbBlue!36}} 5.12 & {\cellcolor{cbBlue!33}} 4.55 \\
 &  & L20 & {\cellcolor{cbBlue!27}} 3.34 & {\cellcolor{cbBlue!28}} 3.66 & {\cellcolor{cbBlue!27}} 3.31 & {\cellcolor{cbBlue!23}} 2.69 & {\cellcolor{cbBlue!16}} 1.34 & {\cellcolor{cbBlue!23}} 2.66 & {\cellcolor{cbBlue!12}} 0.52 & {\cellcolor{cbBlue!24}} 2.82 & {\cellcolor{cbBlue!45}} 6.81 & {\cellcolor{cbBlue!37}} 5.42 & {\cellcolor{cbBlue!32}} 4.39 & {\cellcolor{cbBlue!27}} 3.36 \\
 &  & L31 & {\cellcolor{cbBlue!21}} 2.25 & {\cellcolor{cbBlue!31}} 4.22 & {\cellcolor{cbBlue!21}} 2.22 & {\cellcolor{cbBlue!20}} 2.10 & {\cellcolor{cbBlue!19}} 1.81 & {\cellcolor{cbBlue!19}} 1.77 & {\cellcolor{cbOrange!11}} -0.26 & {\cellcolor{cbBlue!26}} 3.27 & {\cellcolor{cbBlue!80}} 13.58 & {\cellcolor{cbBlue!35}} 4.96 & {\cellcolor{cbBlue!25}} 3.01 & {\cellcolor{cbBlue!28}} 3.54 \\
 \cmidrule(lr){2-15}
 & \multirow[c]{3}{*}{Llama3.1 70b} & L17 & {\cellcolor{cbBlue!37}} 5.41 & {\cellcolor{cbBlue!16}} 1.26 & {\cellcolor{cbBlue!15}} 1.02 & {\cellcolor{cbBlue!19}} 1.87 & {\cellcolor{cbBlue!18}} 1.56 & {\cellcolor{cbBlue!15}} 1.00 & {\cellcolor{cbBlue!15}} 1.08 & {\cellcolor{cbBlue!22}} 2.46 & {\cellcolor{cbBlue!47}} 7.30 & {\cellcolor{cbBlue!19}} 1.89 & {\cellcolor{cbBlue!25}} 2.95 & {\cellcolor{cbBlue!23}} 2.53 \\
 &  & L37 & {\cellcolor{cbBlue!34}} 4.83 & {\cellcolor{cbBlue!12}} 0.44 & {\cellcolor{cbBlue!13}} 0.60 & {\cellcolor{cbBlue!13}} 0.69 & {\cellcolor{cbBlue!17}} 1.53 & {\cellcolor{cbBlue!18}} 1.71 & {\cellcolor{cbOrange!19}} -1.84 & {\cellcolor{cbBlue!20}} 2.12 & {\cellcolor{cbBlue!48}} 7.51 & {\cellcolor{cbBlue!18}} 1.71 & {\cellcolor{cbBlue!24}} 2.73 & {\cellcolor{cbBlue!20}} 2.00 \\
 &  & L57 & {\cellcolor{cbBlue!35}} 4.96 & {\cellcolor{cbBlue!16}} 1.17 & {\cellcolor{cbBlue!16}} 1.31 & {\cellcolor{cbBlue!25}} 2.93 & {\cellcolor{cbBlue!14}} 0.92 & {\cellcolor{cbBlue!14}} 0.92 & {\cellcolor{cbBlue!11}} 0.26 & {\cellcolor{cbBlue!20}} 2.03 & {\cellcolor{cbBlue!41}} 6.08 & {\cellcolor{cbBlue!18}} 1.68 & {\cellcolor{cbBlue!14}} 0.90 & {\cellcolor{cbBlue!20}} 2.11 \\
 \cmidrule(lr){2-15}
 & \multirow[c]{3}{*}{Llama3.1 8b} & L7 & {\cellcolor{cbBlue!77}} 13.13 & {\cellcolor{cbBlue!25}} 2.97 & {\cellcolor{cbBlue!30}} 3.89 & {\cellcolor{cbBlue!15}} 1.04 & {\cellcolor{cbBlue!16}} 1.26 & {\cellcolor{cbBlue!18}} 1.63 & {\cellcolor{cbBlue!19}} 1.79 & {\cellcolor{cbBlue!37}} 5.38 & {\cellcolor{cbBlue!32}} 4.31 & {\cellcolor{cbBlue!32}} 4.46 & {\cellcolor{cbBlue!14}} 0.83 & {\cellcolor{cbBlue!29}} 3.70 \\
 &  & L15 & {\cellcolor{cbBlue!68}} 11.36 & {\cellcolor{cbBlue!20}} 1.98 & {\cellcolor{cbBlue!21}} 2.17 & {\cellcolor{cbOrange!10}} -0.15 & {\cellcolor{cbBlue!11}} 0.24 & {\cellcolor{cbBlue!19}} 1.87 & {\cellcolor{cbBlue!18}} 1.60 & {\cellcolor{cbBlue!35}} 4.96 & {\cellcolor{cbBlue!27}} 3.34 & {\cellcolor{cbBlue!24}} 2.85 & {\cellcolor{cbBlue!17}} 1.49 & {\cellcolor{cbBlue!24}} 2.88 \\
 &  & L23 & {\cellcolor{cbBlue!65}} 10.85 & {\cellcolor{cbBlue!24}} 2.73 & {\cellcolor{cbBlue!19}} 1.85 & {\cellcolor{cbBlue!11}} 0.32 & {\cellcolor{cbOrange!23}} -2.65 & {\cellcolor{cbBlue!23}} 2.68 & {\cellcolor{cbBlue!13}} 0.74 & {\cellcolor{cbBlue!31}} 4.09 & {\cellcolor{cbBlue!27}} 3.41 & {\cellcolor{cbBlue!26}} 3.13 & {\cellcolor{cbBlue!17}} 1.43 & {\cellcolor{cbBlue!23}} 2.60 \\
\bottomrule
\end{tabular}
}
\end{table*}

\paragraph{Quality Vectors.} We follow the contrastive activation extraction approach of \citet{rimsky-etal-2024-steering}, which requires paired examples representing opposing properties (in our case: human vs. generated samples). To construct these pairs, we generate backtranslations for FLORES and BOUQUET using the distilled NLLB model~\cite{koishekenov-etal-2023-memory}. We treat the original human-authored text as the desired representation and the backtranslated text as its contrastive counterpart, following prior findings that synthetic text is generally lower quality than human-authored data~\cite{schaffelder2026syntheticeggsbasketsimpact}. Each \textit{Quality} steering vector is created language-specifically from the contrastive activations of human-authored and backtranslated texts from that given language. The aim is to capture subtle representational differences between human-written and synthetic text. An example is shown in Figure~\ref{fig:steer_qual_example}. All steering vectors are derived from task-agnostic data. Additional details are provided in Appendix~\ref{sec:appendix_backtranslation}.

To construct \textit{Quality} steering vectors, we use TransformerLens \citep{nanda2022transformerlens} to extract residual stream activations from transformer blocks without further training. We compute token-wise global averages for each example and define the steering vector as the difference between the mean activations of the contrastive pairs. Each of the \textit{Quality} steering vectors is computed for a specific language (from data from that specific language). Details can be found in Appendix~\ref{appendix:sec_steer_vec_comp}.

\subsection{Using Steering Vectors for Generation}
To evaluate the effectiveness of steering vectors for synthetic data generation, we apply them at equivalent relative depths across four instruction-tuned LLMs: Gemma-2-9B, Gemma-2-27B, Llama-3.1-8B, and Llama-3.1-70B~\cite{gemmateam2024gemma2improvingopen, llama3modelcard}. Steering is applied at approximately $21\%$, $48\%$, and $74\%$ of model depth, corresponding to early, middle, and late processing stages in the Transformer hierarchy~\cite{bartoszcze2025representationengineeringlargelanguagemodels}. Early layers should be primarily associated with the consolidation of syntactic and linguistic identity~\cite{tenney-etal-2019-bert}; the middle layers should represent the peak of conceptual abstraction~\cite{geva-etal-2021-transformer}; and the later layers, where the model transitions from abstract conceptual processing toward task-specific refinement and next-token probability mapping~\cite{belrose2025elicitinglatentpredictionstransformers}. These layers were additionally selected based on stable linear probe performance in distinguishing human-authored from synthetic text (Section~\ref{sec:linear_probe}). Further details are in Appendix~\ref{sec:appendix_prompt_templates}.

We further investigate the effect of steering strength $\alpha$. For Gemma models, we evaluate $\alpha \in \{25, 50, 75, 100\}$, while for Llama models we use $\alpha \in \{1, 2, 3, 4\}$. We observe substantial differences in sensitivity between the model families: Llama models frequently collapse at higher $\alpha$ values (e.g., repetition or empty outputs), whereas Gemma models require larger intervention strengths, remaining stable even at $\alpha=100$. We attribute this to architectural differences, particularly Gemma-2's use of logit soft-capping and Query-Key normalization~\cite{gemmateam2024gemma2improvingopen}, as well as its larger residual stream norms.

\begin{table*}
\caption{Mean F1 difference across languages, models, steering methods, and layers aggregated over tasks, with \textcolor{cbBlue}{positive} and \textcolor{cbOrange}{negative} differences compared to a \textbf{few-shot prompt} with no steering.  The best alpha values and results were selected based on validation set results per layer, task, and language combination.}
\label{tab:steering_results_icl}
\tiny
\resizebox{\textwidth}{!}{%
\begin{tabular}{lllccccccccccc||c}
\toprule
Method & LLM & Layer & am & cs & da & de & he & id & jv & mt & sk & sl & su & Average \\
\midrule
\midrule
\multirow[c]{12}{*}{\textit{Language}} & \multirow[c]{3}{*}{Gemma 2 27b} & L10 & {\cellcolor{cbBlue!22}} 1.58 & {\cellcolor{cbBlue!29}} 2.48 & {\cellcolor{cbBlue!26}} 2.04 & {\cellcolor{cbBlue!25}} 1.91 & {\cellcolor{cbBlue!39}} 3.71 & {\cellcolor{cbBlue!37}} 3.48 & {\cellcolor{cbBlue!13}} 0.41 & {\cellcolor{cbBlue!36}} 3.37 & {\cellcolor{cbBlue!13}} 0.49 & {\cellcolor{cbBlue!49}} 5.09 & {\cellcolor{cbOrange!12}} -0.30 & {\cellcolor{cbBlue!27}} 2.20 \\
 &  & L22 & {\cellcolor{cbOrange!14}} -0.64 & {\cellcolor{cbBlue!45}} 4.51 & {\cellcolor{cbBlue!12}} 0.37 & {\cellcolor{cbBlue!24}} 1.86 & {\cellcolor{cbBlue!24}} 1.87 & {\cellcolor{cbBlue!23}} 1.66 & {\cellcolor{cbOrange!10}} -0.02 & {\cellcolor{cbBlue!31}} 2.68 & {\cellcolor{cbOrange!11}} -0.20 & {\cellcolor{cbBlue!55}} 5.75 & {\cellcolor{cbBlue!14}} 0.55 & {\cellcolor{cbBlue!23}} 1.67 \\
 &  & L34 & {\cellcolor{cbBlue!17}} 0.97 & {\cellcolor{cbBlue!45}} 4.47 & {\cellcolor{cbBlue!23}} 1.72 & {\cellcolor{cbBlue!23}} 1.73 & {\cellcolor{cbBlue!16}} 0.80 & {\cellcolor{cbBlue!28}} 2.29 & {\cellcolor{cbBlue!13}} 0.46 & {\cellcolor{cbBlue!26}} 2.05 & {\cellcolor{cbBlue!24}} 1.85 & {\cellcolor{cbBlue!40}} 3.94 & {\cellcolor{cbBlue!11}} 0.16 & {\cellcolor{cbBlue!24}} 1.86 \\
  \cmidrule(lr){2-15}
 & \multirow[c]{3}{*}{Gemma 2 9b} & L9 & {\cellcolor{cbOrange!36}} -3.42 & {\cellcolor{cbBlue!43}} 4.31 & {\cellcolor{cbBlue!23}} 1.70 & {\cellcolor{cbBlue!30}} 2.63 & {\cellcolor{cbBlue!19}} 1.22 & {\cellcolor{cbOrange!21}} -1.49 & {\cellcolor{cbBlue!32}} 2.84 & {\cellcolor{cbBlue!25}} 1.93 & {\cellcolor{cbBlue!52}} 5.40 & {\cellcolor{cbOrange!19}} -1.16 & {\cellcolor{cbOrange!13}} -0.49 & {\cellcolor{cbBlue!19}} 1.22 \\
 &  & L20 & {\cellcolor{cbOrange!10}} -0.07 & {\cellcolor{cbBlue!47}} 4.79 & {\cellcolor{cbBlue!54}} 5.63 & {\cellcolor{cbBlue!23}} 1.70 & {\cellcolor{cbBlue!63}} 6.82 & {\cellcolor{cbBlue!22}} 1.56 & {\cellcolor{cbBlue!42}} 4.07 & {\cellcolor{cbOrange!14}} -0.52 & {\cellcolor{cbBlue!36}} 3.36 & {\cellcolor{cbBlue!20}} 1.39 & {\cellcolor{cbBlue!13}} 0.46 & {\cellcolor{cbBlue!30}} 2.65 \\
 &  & L31 & {\cellcolor{cbOrange!15}} -0.70 & {\cellcolor{cbBlue!27}} 2.24 & {\cellcolor{cbBlue!45}} 4.56 & {\cellcolor{cbBlue!30}} 2.57 & {\cellcolor{cbBlue!22}} 1.56 & {\cellcolor{cbBlue!10}} 0.12 & {\cellcolor{cbBlue!44}} 4.41 & {\cellcolor{cbBlue!18}} 1.03 & {\cellcolor{cbBlue!37}} 3.50 & {\cellcolor{cbBlue!14}} 0.57 & {\cellcolor{cbBlue!14}} 0.53 & {\cellcolor{cbBlue!24}} 1.85 \\
  \cmidrule(lr){2-15}
 & \multirow[c]{3}{*}{Llama3.1 70b} & L17 & {\cellcolor{cbBlue!26}} 2.10 & {\cellcolor{cbOrange!10}} -0.12 & {\cellcolor{cbOrange!11}} -0.13 & {\cellcolor{cbBlue!39}} 3.75 & {\cellcolor{cbBlue!37}} 3.55 & {\cellcolor{cbBlue!51}} 5.29 & {\cellcolor{cbBlue!12}} 0.38 & {\cellcolor{cbBlue!18}} 1.12 & {\cellcolor{cbBlue!62}} 6.67 & {\cellcolor{cbBlue!17}} 0.98 & {\cellcolor{cbBlue!26}} 2.04 & {\cellcolor{cbBlue!28}} 2.33 \\
 &  & L37 & {\cellcolor{cbBlue!12}} 0.32 & {\cellcolor{cbBlue!20}} 1.37 & {\cellcolor{cbBlue!20}} 1.35 & {\cellcolor{cbBlue!19}} 1.25 & {\cellcolor{cbBlue!46}} 4.67 & {\cellcolor{cbBlue!34}} 3.09 & {\cellcolor{cbBlue!17}} 0.98 & {\cellcolor{cbBlue!35}} 3.23 & {\cellcolor{cbBlue!38}} 3.65 & {\cellcolor{cbBlue!16}} 0.82 & {\cellcolor{cbBlue!45}} 4.48 & {\cellcolor{cbBlue!28}} 2.29 \\
 &  & L57 & {\cellcolor{cbBlue!19}} 1.22 & {\cellcolor{cbBlue!21}} 1.44 & {\cellcolor{cbBlue!23}} 1.67 & {\cellcolor{cbBlue!21}} 1.44 & {\cellcolor{cbBlue!44}} 4.41 & {\cellcolor{cbBlue!23}} 1.72 & {\cellcolor{cbBlue!23}} 1.68 & {\cellcolor{cbBlue!21}} 1.42 & {\cellcolor{cbBlue!23}} 1.73 & {\cellcolor{cbBlue!23}} 1.73 & {\cellcolor{cbBlue!43}} 4.26 & {\cellcolor{cbBlue!26}} 2.07 \\
  \cmidrule(lr){2-15}
 & \multirow[c]{3}{*}{Llama3.1 8b} & L7 & {\cellcolor{cbBlue!22}} 1.62 & {\cellcolor{cbBlue!25}} 2.01 & {\cellcolor{cbBlue!36}} 3.43 & {\cellcolor{cbBlue!27}} 2.26 & {\cellcolor{cbBlue!18}} 1.06 & {\cellcolor{cbBlue!24}} 1.80 & {\cellcolor{cbOrange!13}} -0.47 & {\cellcolor{cbBlue!33}} 2.95 & {\cellcolor{cbOrange!14}} -0.52 & {\cellcolor{cbBlue!24}} 1.80 & {\cellcolor{cbOrange!14}} -0.59 & {\cellcolor{cbBlue!20}} 1.40 \\
 &  & L15 & {\cellcolor{cbBlue!20}} 1.39 & {\cellcolor{cbBlue!27}} 2.23 & {\cellcolor{cbBlue!26}} 2.15 & {\cellcolor{cbBlue!23}} 1.75 & {\cellcolor{cbOrange!10}} -0.10 & {\cellcolor{cbBlue!29}} 2.46 & {\cellcolor{cbBlue!20}} 1.34 & {\cellcolor{cbBlue!32}} 2.84 & {\cellcolor{cbBlue!37}} 3.45 & {\cellcolor{cbBlue!28}} 2.30 & {\cellcolor{cbOrange!11}} -0.23 & {\cellcolor{cbBlue!24}} 1.78 \\
 &  & L23 & {\cellcolor{cbBlue!24}} 1.89 & {\cellcolor{cbBlue!32}} 2.84 & {\cellcolor{cbBlue!28}} 2.41 & {\cellcolor{cbBlue!27}} 2.19 & {\cellcolor{cbBlue!13}} 0.43 & {\cellcolor{cbBlue!30}} 2.61 & {\cellcolor{cbBlue!27}} 2.28 & {\cellcolor{cbBlue!27}} 2.23 & {\cellcolor{cbBlue!32}} 2.91 & {\cellcolor{cbBlue!28}} 2.30 & {\cellcolor{cbBlue!16}} 0.81 & {\cellcolor{cbBlue!26}} 2.08 \\
 \midrule
 \midrule
\multirow[c]{12}{*}{\textit{Quality}} & \multirow[c]{3}{*}{Gemma 2 27b} & L10 & {\cellcolor{cbOrange!11}} -0.13 & {\cellcolor{cbBlue!27}} 2.23 & {\cellcolor{cbBlue!29}} 2.52 & {\cellcolor{cbBlue!19}} 1.25 & {\cellcolor{cbBlue!27}} 2.22 & {\cellcolor{cbBlue!27}} 2.26 & {\cellcolor{cbBlue!10}} 0.10 & {\cellcolor{cbBlue!38}} 3.68 & {\cellcolor{cbBlue!28}} 2.31 & {\cellcolor{cbBlue!51}} 5.33 & {\cellcolor{cbBlue!15}} 0.70 & {\cellcolor{cbBlue!26}} 2.04 \\
 &  & L22 & {\cellcolor{cbBlue!43}} 4.25 & {\cellcolor{cbBlue!41}} 4.07 & {\cellcolor{cbBlue!24}} 1.81 & {\cellcolor{cbBlue!26}} 2.06 & {\cellcolor{cbBlue!27}} 2.20 & {\cellcolor{cbBlue!33}} 3.01 & {\cellcolor{cbOrange!13}} -0.46 & {\cellcolor{cbBlue!33}} 2.97 & {\cellcolor{cbBlue!19}} 1.26 & {\cellcolor{cbBlue!55}} 5.80 & {\cellcolor{cbBlue!13}} 0.45 & {\cellcolor{cbBlue!29}} 2.49 \\
 &  & L34 & {\cellcolor{cbBlue!19}} 1.17 & {\cellcolor{cbBlue!39}} 3.71 & {\cellcolor{cbBlue!19}} 1.22 & {\cellcolor{cbBlue!18}} 1.10 & {\cellcolor{cbBlue!38}} 3.66 & {\cellcolor{cbBlue!24}} 1.84 & {\cellcolor{cbBlue!12}} 0.26 & {\cellcolor{cbBlue!37}} 3.55 & {\cellcolor{cbBlue!24}} 1.80 & {\cellcolor{cbBlue!37}} 3.47 & {\cellcolor{cbBlue!15}} 0.67 & {\cellcolor{cbBlue!26}} 2.04 \\
  \cmidrule(lr){2-15}
 & \multirow[c]{3}{*}{Gemma 9b} & L9 & {\cellcolor{cbBlue!15}} 0.74 & {\cellcolor{cbBlue!35}} 3.24 & {\cellcolor{cbBlue!31}} 2.75 & {\cellcolor{cbBlue!41}} 3.95 & {\cellcolor{cbBlue!80}} 8.91 & {\cellcolor{cbBlue!18}} 1.10 & {\cellcolor{cbBlue!41}} 4.03 & {\cellcolor{cbBlue!35}} 3.19 & {\cellcolor{cbBlue!26}} 2.08 & {\cellcolor{cbBlue!21}} 1.45 & {\cellcolor{cbBlue!21}} 1.50 & {\cellcolor{cbBlue!33}} 2.99 \\
 &  & L20 & {\cellcolor{cbOrange!12}} -0.34 & {\cellcolor{cbBlue!22}} 1.57 & {\cellcolor{cbBlue!27}} 2.26 & {\cellcolor{cbBlue!23}} 1.66 & {\cellcolor{cbBlue!69}} 7.62 & {\cellcolor{cbBlue!20}} 1.31 & {\cellcolor{cbBlue!21}} 1.52 & {\cellcolor{cbBlue!15}} 0.68 & {\cellcolor{cbBlue!22}} 1.62 & {\cellcolor{cbOrange!10}} -0.02 & {\cellcolor{cbBlue!28}} 2.31 & {\cellcolor{cbBlue!24}} 1.84 \\
 &  & L31 & {\cellcolor{cbOrange!11}} -0.19 & {\cellcolor{cbBlue!27}} 2.21 & {\cellcolor{cbBlue!39}} 3.77 & {\cellcolor{cbBlue!28}} 2.30 & {\cellcolor{cbBlue!42}} 4.08 & {\cellcolor{cbBlue!16}} 0.87 & {\cellcolor{cbBlue!14}} 0.52 & {\cellcolor{cbOrange!18}} -1.14 & {\cellcolor{cbBlue!44}} 4.38 & {\cellcolor{cbBlue!12}} 0.36 & {\cellcolor{cbBlue!18}} 1.03 & {\cellcolor{cbBlue!22}} 1.65 \\
  \cmidrule(lr){2-15}
 & \multirow[c]{3}{*}{Llama3 70b} & L17 & {\cellcolor{cbBlue!22}} 1.55 & {\cellcolor{cbBlue!22}} 1.65 & {\cellcolor{cbBlue!18}} 1.14 & {\cellcolor{cbBlue!25}} 1.97 & {\cellcolor{cbBlue!52}} 5.42 & {\cellcolor{cbBlue!33}} 3.02 & {\cellcolor{cbBlue!21}} 1.50 & {\cellcolor{cbBlue!37}} 3.46 & {\cellcolor{cbBlue!61}} 6.52 & {\cellcolor{cbBlue!27}} 2.16 & {\cellcolor{cbBlue!38}} 3.63 & {\cellcolor{cbBlue!32}} 2.91 \\
 &  & L37 & {\cellcolor{cbBlue!17}} 0.96 & {\cellcolor{cbBlue!18}} 1.11 & {\cellcolor{cbBlue!19}} 1.16 & {\cellcolor{cbBlue!24}} 1.90 & {\cellcolor{cbBlue!34}} 3.12 & {\cellcolor{cbBlue!15}} 0.66 & {\cellcolor{cbBlue!22}} 1.56 & {\cellcolor{cbOrange!10}} -0.07 & {\cellcolor{cbBlue!32}} 2.89 & {\cellcolor{cbBlue!28}} 2.34 & {\cellcolor{cbBlue!47}} 4.77 & {\cellcolor{cbBlue!24}} 1.86 \\
 &  & L57 & {\cellcolor{cbOrange!10}} -0.09 & {\cellcolor{cbBlue!21}} 1.46 & {\cellcolor{cbBlue!17}} 0.97 & {\cellcolor{cbBlue!16}} 0.88 & {\cellcolor{cbBlue!28}} 2.39 & {\cellcolor{cbBlue!22}} 1.56 & {\cellcolor{cbBlue!19}} 1.26 & {\cellcolor{cbBlue!29}} 2.44 & {\cellcolor{cbBlue!31}} 2.75 & {\cellcolor{cbBlue!22}} 1.65 & {\cellcolor{cbBlue!53}} 5.48 & {\cellcolor{cbBlue!24}} 1.89 \\
  \cmidrule(lr){2-15}
 & \multirow[c]{3}{*}{Llama3 8b} & L7 & {\cellcolor{cbBlue!28}} 2.39 & {\cellcolor{cbBlue!27}} 2.18 & {\cellcolor{cbBlue!41}} 3.95 & {\cellcolor{cbBlue!28}} 2.37 & {\cellcolor{cbBlue!17}} 0.91 & {\cellcolor{cbBlue!17}} 1.00 & {\cellcolor{cbBlue!14}} 0.62 & {\cellcolor{cbBlue!34}} 3.13 & {\cellcolor{cbBlue!16}} 0.79 & {\cellcolor{cbBlue!14}} 0.54 & {\cellcolor{cbBlue!12}} 0.37 & {\cellcolor{cbBlue!23}} 1.66 \\
 &  & L15 & {\cellcolor{cbBlue!21}} 1.49 & {\cellcolor{cbBlue!17}} 0.93 & {\cellcolor{cbBlue!12}} 0.31 & {\cellcolor{cbBlue!23}} 1.75 & {\cellcolor{cbBlue!13}} 0.48 & {\cellcolor{cbBlue!21}} 1.47 & {\cellcolor{cbOrange!11}} -0.23 & {\cellcolor{cbBlue!35}} 3.21 & {\cellcolor{cbOrange!18}} -1.08 & {\cellcolor{cbOrange!10}} -0.04 & {\cellcolor{cbOrange!12}} -0.28 & {\cellcolor{cbBlue!15}} 0.73 \\
 &  & L23 & {\cellcolor{cbBlue!48}} 4.89 & {\cellcolor{cbBlue!17}} 1.00 & {\cellcolor{cbOrange!17}} -0.89 & {\cellcolor{cbBlue!26}} 2.06 & {\cellcolor{cbOrange!13}} -0.49 & {\cellcolor{cbBlue!19}} 1.18 & {\cellcolor{cbBlue!11}} 0.23 & {\cellcolor{cbBlue!33}} 2.97 & {\cellcolor{cbOrange!11}} -0.25 & {\cellcolor{cbBlue!25}} 1.99 & {\cellcolor{cbBlue!12}} 0.31 & {\cellcolor{cbBlue!19}} 1.18 \\
\bottomrule
\end{tabular}
}
\end{table*}

\subsection{Finetuning Downstream Models}
In our experiments, downstream model performance serves as the primary indicator of synthetic data quality, following prior work~\cite{anikina-etal-2025-rigorous, cegin-etal-2026-rose}. We evaluate the effect of \textit{Language} and \textit{Quality} steering on synthetic data for topic classification (multi-class) and sentiment analysis (binary). Due to the limited availability of multilingual benchmarks, we use SIB-200~\cite{adelani-etal-2024-sib} and the sentiment dataset collection of~\cite{Brychcin.Habernal.2013, gurgurov-etal-2024-adapting, gurgurov-etal-2025-gremlin}. For each combination of LLM, language, layer, $\alpha$, and task, we generate synthetic datasets, producing 50 samples per label for sentiment and 20 samples per label for topic classification, similar to prior setups~\cite{anikina-etal-2025-rigorous}.

For downstream evaluation, we fine-tune XLM-R~\cite{DBLP:journals/corr/abs-1911-02116} (\textit{facebookAI/xlm-roberta-base}) with early stopping. We compare steering-based generation against two baselines: (1) zero-shot prompting without demonstrations, and (2) state-of-the-art few-shot prompting with human demonstrations~\cite{anikina-etal-2025-rigorous}. Prompt templates are found in Appendix~\ref{sec:appendix_diversity_metrics}, with further implementation details provided in Appendix~\ref{sec:appendix_downstream_finetuning}. For each combination of language, prompt, steering method, layer, and model, we report the downstream model performance achieved by the model with the best validation set loss. This led to various alphas for each cell in Tables~\ref{tab:steering_results} and~\ref{tab:steering_results_icl}, due to which we report their corresponding alpha values in Tables~\ref{tab:selected_alpha_zero_shot} and~\ref{tab:selected_alpha_few_shot}. We also investigate a case where we choose the same alpha per layer, steering method, and LLM generator to see if steering is applicable in cases where there is no validation set in Section~\ref{sec:only_best_alpha}.

\begin{table*}[t!]
\centering
\caption{Relative diversity metric differences across models, steering methods, and layers, aggregated over tasks and different alphas. Relative \textcolor{cbBlue}{positive} and \textcolor{cbOrange}{negative} changes.}
\begin{subtable}[t]{0.48\textwidth}
\centering
\tiny
\caption{\textbf{Zero-shot} setup}
\label{tab:diversity_results}
\begin{tabular}{llcccc}
\toprule
 &  & LexDiv & EmbDiv & IsoRad. & Homogen. \\
Model + Steering & Layer &  &  &  &  \\
\midrule
\midrule
\multirow[c]{3}{*}{Gemma 2 27b (lang.)} & L10 & {\cellcolor{cbBlue!10}} 0.06 & {\cellcolor{cbOrange!11}} -1.35 & {\cellcolor{cbOrange!11}} -1.02 & {\cellcolor{cbOrange!11}} -1.24 \\
 & L22 & {\cellcolor{cbOrange!10}} -0.00 & {\cellcolor{cbOrange!12}} -2.16 & {\cellcolor{cbOrange!10}} -0.98 & {\cellcolor{cbBlue!10}} 0.55 \\
 & L34 & {\cellcolor{cbBlue!10}} 0.42 & {\cellcolor{cbOrange!11}} -1.17 & {\cellcolor{cbOrange!10}} -0.26 & {\cellcolor{cbBlue!13}} 3.06 \\
\midrule
\multirow[c]{3}{*}{Gemma 2 27b (qual.)} & L10 & {\cellcolor{cbBlue!10}} 0.83 & {\cellcolor{cbOrange!10}} -0.01 & {\cellcolor{cbOrange!10}} -0.62 & {\cellcolor{cbBlue!11}} 1.45 \\
 & L22 & {\cellcolor{cbBlue!10}} 0.30 & {\cellcolor{cbOrange!11}} -1.55 & {\cellcolor{cbOrange!10}} -0.92 & {\cellcolor{cbOrange!10}} -0.51 \\
 & L34 & {\cellcolor{cbOrange!10}} -0.16 & {\cellcolor{cbOrange!12}} -2.38 & {\cellcolor{cbOrange!10}} -0.60 & {\cellcolor{cbBlue!11}} 1.96 \\
\midrule
\multirow[c]{3}{*}{Gemma 2 9b (lang.)} & L9 & {\cellcolor{cbBlue!12}} 2.12 & {\cellcolor{cbBlue!33}} 23.51 & {\cellcolor{cbBlue!11}} 1.81 & {\cellcolor{cbBlue!22}} 12.50 \\
 & L20 & {\cellcolor{cbBlue!20}} 10.72 & {\cellcolor{cbBlue!23}} 13.57 & {\cellcolor{cbBlue!11}} 1.24 & {\cellcolor{cbBlue!16}} 6.07 \\
 & L31 & {\cellcolor{cbBlue!17}} 7.68 & {\cellcolor{cbBlue!26}} 16.54 & {\cellcolor{cbBlue!11}} 0.99 & {\cellcolor{cbBlue!14}} 4.80 \\
\midrule
\multirow[c]{3}{*}{Gemma 2 9b (qual.)} & L9 & {\cellcolor{cbBlue!32}} 21.78 & {\cellcolor{cbBlue!63}} 52.83 & {\cellcolor{cbBlue!14}} 4.09 & {\cellcolor{cbBlue!21}} 11.33 \\
 & L20 & {\cellcolor{cbBlue!27}} 17.07 & {\cellcolor{cbBlue!28}} 18.22 & {\cellcolor{cbBlue!10}} 0.10 & {\cellcolor{cbBlue!15}} 5.52 \\
 & L31 & {\cellcolor{cbBlue!32}} 22.15 & {\cellcolor{cbBlue!42}} 32.59 & {\cellcolor{cbBlue!10}} 0.66 & {\cellcolor{cbBlue!16}} 6.80 \\
\midrule
\multirow[c]{3}{*}{Llama3.1 70b (lang.)} & L17 & {\cellcolor{cbOrange!39}} -29.03 & {\cellcolor{cbBlue!65}} 55.32 & {\cellcolor{cbBlue!11}} 1.93 & {\cellcolor{cbBlue!23}} 13.13 \\
 & L37 & {\cellcolor{cbOrange!12}} -2.43 & {\cellcolor{cbOrange!12}} -2.94 & {\cellcolor{cbOrange!10}} -0.85 & {\cellcolor{cbBlue!11}} 1.60 \\
 & L57 & {\cellcolor{cbOrange!10}} -0.97 & {\cellcolor{cbOrange!14}} -4.14 & {\cellcolor{cbOrange!11}} -1.61 & {\cellcolor{cbOrange!10}} -0.47 \\
\midrule
\multirow[c]{3}{*}{Llama3.1 70b (qual.)} & L17 & {\cellcolor{cbBlue!25}} 14.98 & {\cellcolor{cbBlue!38}} 28.64 & {\cellcolor{cbBlue!11}} 1.90 & {\cellcolor{cbBlue!16}} 6.91 \\
 & L37 & {\cellcolor{cbBlue!20}} 10.33 & {\cellcolor{cbBlue!18}} 8.37 & {\cellcolor{cbOrange!10}} -0.06 & {\cellcolor{cbBlue!17}} 7.26 \\
 & L57 & {\cellcolor{cbBlue!12}} 2.40 & {\cellcolor{cbOrange!16}} -6.37 & {\cellcolor{cbOrange!11}} -1.57 & {\cellcolor{cbBlue!11}} 1.18 \\
\midrule
\multirow[c]{3}{*}{Llama3.1 8b (lang.)} & L7 & {\cellcolor{cbBlue!80}} 69.28 & {\cellcolor{cbBlue!78}} 68.10 & {\cellcolor{cbBlue!13}} 3.24 & {\cellcolor{cbBlue!15}} 5.77 \\
 & L15 & {\cellcolor{cbBlue!54}} 44.03 & {\cellcolor{cbBlue!69}} 58.97 & {\cellcolor{cbBlue!12}} 2.45 & {\cellcolor{cbBlue!13}} 3.34 \\
 & L23 & {\cellcolor{cbBlue!43}} 32.80 & {\cellcolor{cbBlue!30}} 20.41 & {\cellcolor{cbBlue!12}} 2.22 & {\cellcolor{cbBlue!15}} 5.27 \\
\midrule
\multirow[c]{3}{*}{Llama3.1 8b (qual.)} & L7 & {\cellcolor{cbBlue!53}} 43.08 & {\cellcolor{cbBlue!46}} 35.96 & {\cellcolor{cbBlue!13}} 3.35 & {\cellcolor{cbBlue!19}} 8.95 \\
 & L15 & {\cellcolor{cbBlue!44}} 33.73 & {\cellcolor{cbOrange!24}} -14.71 & {\cellcolor{cbBlue!11}} 1.21 & {\cellcolor{cbBlue!15}} 5.63 \\
 & L23 & {\cellcolor{cbBlue!46}} 36.48 & {\cellcolor{cbOrange!12}} -2.27 & {\cellcolor{cbBlue!12}} 2.28 & {\cellcolor{cbBlue!17}} 7.73 \\
\bottomrule
\end{tabular}
\end{subtable}
\hfill
\begin{subtable}[t]{0.48\textwidth}
\centering
\tiny
\caption{\textbf{Few-shot} setup}
\label{tab:diversity_results_icl}
\begin{tabular}{llcccc}
\toprule
 &  & LexDiv & EmbDiv & IsoRad. & Homogen.\\
Model + Steering & Layer &  &  &  &  \\
\midrule
\midrule
\multirow[c]{3}{*}{Gemma 2 27b (lang.)} & L10 & {\cellcolor{cbBlue!10}} 0.33 & {\cellcolor{cbOrange!14}} -3.66 & {\cellcolor{cbBlue!10}} 0.38 & {\cellcolor{cbBlue!11}} 1.22 \\
 & L22 & {\cellcolor{cbBlue!11}} 0.92 & {\cellcolor{cbOrange!16}} -4.92 & {\cellcolor{cbBlue!10}} 0.15 & {\cellcolor{cbBlue!10}} 0.26 \\
 & L34 & {\cellcolor{cbBlue!11}} 0.99 & {\cellcolor{cbOrange!13}} -2.61 & {\cellcolor{cbBlue!10}} 0.25 & {\cellcolor{cbBlue!10}} 0.11 \\
\midrule
\multirow[c]{3}{*}{Gemma 2 27b (qual.)} & L10 & {\cellcolor{cbBlue!11}} 1.37 & {\cellcolor{cbOrange!14}} -3.48 & {\cellcolor{cbBlue!10}} 0.33 & {\cellcolor{cbBlue!11}} 1.00 \\
 & L22 & {\cellcolor{cbBlue!11}} 1.51 & {\cellcolor{cbOrange!15}} -4.45 & {\cellcolor{cbBlue!10}} 0.32 & {\cellcolor{cbBlue!10}} 0.43 \\
 & L34 & {\cellcolor{cbBlue!10}} 0.75 & {\cellcolor{cbOrange!16}} -5.62 & {\cellcolor{cbBlue!10}} 0.49 & {\cellcolor{cbBlue!10}} 0.65 \\
\midrule
\multirow[c]{3}{*}{Gemma 2 9b (lang.)} & L9 & {\cellcolor{cbOrange!22}} -10.57 & {\cellcolor{cbBlue!24}} 11.69 & {\cellcolor{cbBlue!10}} 0.30 & {\cellcolor{cbOrange!15}} -4.56 \\
 & L20 & {\cellcolor{cbBlue!18}} 6.77 & {\cellcolor{cbBlue!24}} 11.91 & {\cellcolor{cbBlue!10}} 0.09 & {\cellcolor{cbOrange!18}} -7.19 \\
 & L31 & {\cellcolor{cbBlue!15}} 4.13 & {\cellcolor{cbBlue!13}} 2.74 & {\cellcolor{cbBlue!10}} 0.24 & {\cellcolor{cbOrange!13}} -2.80 \\
\midrule
\multirow[c]{3}{*}{Gemma 2 9b (qual.)} & L9 & {\cellcolor{cbBlue!23}} 10.82 & {\cellcolor{cbBlue!38}} 23.05 & {\cellcolor{cbBlue!11}} 1.29 & {\cellcolor{cbBlue!10}} 0.19 \\
 & L20 & {\cellcolor{cbBlue!17}} 5.98 & {\cellcolor{cbBlue!37}} 22.17 & {\cellcolor{cbBlue!11}} 0.95 & {\cellcolor{cbBlue!10}} 0.59 \\
 & L31 & {\cellcolor{cbBlue!15}} 4.15 & {\cellcolor{cbBlue!27}} 14.53 & {\cellcolor{cbBlue!10}} 0.20 & {\cellcolor{cbOrange!12}} -1.90 \\
\midrule
\multirow[c]{3}{*}{Llama3.1 70b (lang.)} & L17 & {\cellcolor{cbOrange!46}} -29.66 & {\cellcolor{cbBlue!63}} 43.96 & {\cellcolor{cbBlue!11}} 1.37 & {\cellcolor{cbBlue!16}} 5.36 \\
 & L37 & {\cellcolor{cbOrange!11}} -1.56 & {\cellcolor{cbBlue!15}} 4.48 & {\cellcolor{cbBlue!10}} 0.52 & {\cellcolor{cbBlue!12}} 2.40 \\
 & L57 & {\cellcolor{cbOrange!11}} -1.35 & {\cellcolor{cbBlue!15}} 4.83 & {\cellcolor{cbBlue!10}} 0.29 & {\cellcolor{cbBlue!11}} 1.21 \\
\midrule
\multirow[c]{3}{*}{Llama3.1 70b (qual.)} & L17 & {\cellcolor{cbBlue!20}} 8.15 & {\cellcolor{cbBlue!29}} 15.55 & {\cellcolor{cbBlue!11}} 0.94 & {\cellcolor{cbBlue!13}} 2.62 \\
 & L37 & {\cellcolor{cbBlue!22}} 10.10 & {\cellcolor{cbBlue!25}} 12.51 & {\cellcolor{cbBlue!11}} 1.56 & {\cellcolor{cbBlue!16}} 5.66 \\
 & L57 & {\cellcolor{cbBlue!11}} 1.39 & {\cellcolor{cbBlue!10}} 0.10 & {\cellcolor{cbBlue!10}} 0.16 & {\cellcolor{cbBlue!10}} 0.37 \\
\midrule
\multirow[c]{3}{*}{Llama3.1 8b (lang.)} & L7 & {\cellcolor{cbBlue!63}} 43.19 & {\cellcolor{cbBlue!80}} 57.04 & {\cellcolor{cbBlue!11}} 1.45 & {\cellcolor{cbOrange!18}} -6.83 \\
 & L15 & {\cellcolor{cbBlue!16}} 5.40 & {\cellcolor{cbBlue!58}} 39.75 & {\cellcolor{cbBlue!10}} 0.53 & {\cellcolor{cbOrange!11}} -0.90 \\
 & L23 & {\cellcolor{cbOrange!17}} -6.11 & {\cellcolor{cbBlue!23}} 11.37 & {\cellcolor{cbBlue!10}} 0.01 & {\cellcolor{cbOrange!11}} -1.06 \\
\midrule
\multirow[c]{3}{*}{Llama3.1 8b (qual.)} & L7 & {\cellcolor{cbBlue!27}} 14.00 & {\cellcolor{cbBlue!12}} 1.84 & {\cellcolor{cbOrange!10}} -0.08 & {\cellcolor{cbOrange!13}} -2.83 \\
 & L15 & {\cellcolor{cbOrange!31}} -17.68 & {\cellcolor{cbOrange!64}} -44.81 & {\cellcolor{cbOrange!11}} -1.61 & {\cellcolor{cbBlue!14}} 3.75 \\
 & L23 & {\cellcolor{cbOrange!23}} -11.14 & {\cellcolor{cbOrange!47}} -30.27 & {\cellcolor{cbOrange!11}} -1.09 & {\cellcolor{cbOrange!10}} -0.16 \\
\bottomrule
\end{tabular}
\end{subtable}
\label{tab:both}
\end{table*}
\section{Results and Discussion}

\subsection{Pre-Experiment: Linear Probing Results}
\label{sec:linear_probe}
We first conduct a linear probe analysis of human-authored (Flores + BOUQET) vs. back-translated pairs. This diagnostic step is done to justify our use of linear steering vectors for the \textit{quality} approach; if a simple logistic regression can distinguish between these distributions with high accuracy, it confirms that the ``quality'' concept may be encoded as a specific direction in the latent space that we can manipulate \cite{park2024linearrepresentationhypothesisgeometry}. The details on how the linear probe is computed can be found in Appendix~\ref{sec:appendix_linear_probe}.

As seen in the aggregated results over all languages in Figure~\ref{fig:linear_probe_res}, all tested models across both the Gemma-2 and Llama-3.1 families exhibit accuracies significantly above the 0.5 random baseline from the very first layer. While the probe accuracy peaks in the middle-to-late layers, the "quality" signal is already well-defined by the earlier layers. The Gemma-2-9b model peaks at Layer 17 with an accuracy of approximately 0.75, while the larger Gemma-2-27b model peaks at Layer 22 with a slightly higher accuracy of approximately 0.76. The Llama-3.1-70b model demonstrates superior separability compared to its 8b counterpart, reaching a sustained plateau of approximately 0.77 accuracy between layers 35 and 40.

\subsection{Downstream Model Performance Evaluation}
\label{sec:downstream_model_res}

We report the results for only the best alpha per cell in Tables~\ref{tab:steering_results} and~\ref{tab:steering_results_icl} in terms of downstream model performance over both tasks combined. Additional results about how different alpha values affect downstream model performance can be found in the Appendix~\ref{sec:appendix_alpha_analysis}.  For statistical tests, we used Mann-Whitney-U \cite{mann1947test} with \textit{p=0.05}. Full F1 baseline results for both zero- and few-shot settings can be found in Appendix~\ref{sec:appendix_baseline_f1}. 

\paragraph{Steering vectors applied with zero-shot prompts}
Aggregated results across tasks are shown in Table~\ref{tab:steering_results}, with per-task results provided in Appendix~\ref{sec:appendix_per_task_f1}. Overall, \textit{Quality} steering consistently outperformed \textit{Language} steering in terms of downstream F1 gains, suggesting that steering toward a ``human-authored'' manifold is more beneficial than steering toward generic linguistic identity alone.

Earlier layers proved the most effective intervention points, particularly for \textit{Quality} steering. Across all task-aggregated experiments, \textit{Quality} steering improved performance in 97.73\% of cases for early layers, compared to 95.45\% and 93.18\% for middle and late layers. \textit{Language} steering showed an inverse trend, with gains in 86.36\%, 88.64\%, and 90.91\% of cases, respectively. Statistically significant improvements for \textit{Quality} steering occurred in 45.46\% of cases for early layers, versus 45.46\% and 36.36\% for middle and late layers. For \textit{Language} steering, statistically significant gains were observed in 34.09\%, 36.36\%, and 35.23\% of cases, respectively. Per-task results are provided in Appendix~\ref{sec:appendix_per_task_f1}.

Across languages, Slovak, and Slovenian showed the largest gains, while Javanese and Danish were comparatively resistant to steering, particularly for Gemma models. Both were languages with a considerable number of negative outcomes. Among models, Gemma-2-9B and Llama-3.1-8B were the most responsive to steering, producing the largest absolute F1 improvements. Overall, Llama models benefited more consistently from steering, whereas Gemma-2-27B appeared more conservative, suggesting that larger models may require stronger or more precise interventions.

\begin{figure*}[t!]
\centering
\begin{subfigure}{0.4\textwidth}
    \centering
    \includegraphics[width=\textwidth]{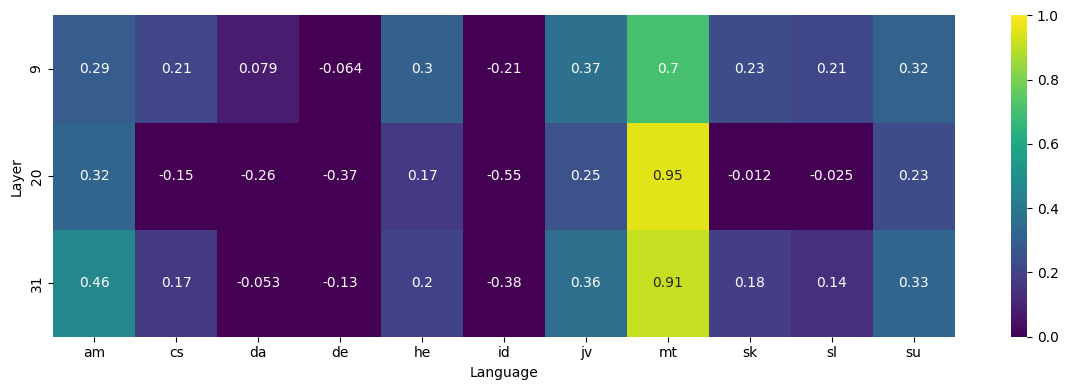}
    \caption{Gemma 2 9b}
    \label{fig:results_corrs}
\end{subfigure}
\hfill
\begin{subfigure}{0.4\textwidth}
    \centering
    \includegraphics[width=\textwidth]{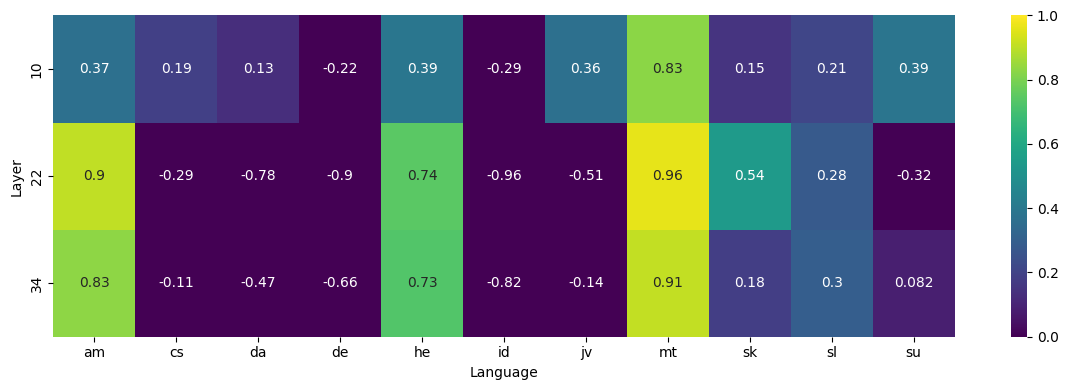}
    \caption{Gemma 2 27b}
    \label{fig:results_corrs_kendal}
\end{subfigure}


\begin{subfigure}{0.4\textwidth}
    \centering
    \includegraphics[width=\textwidth]{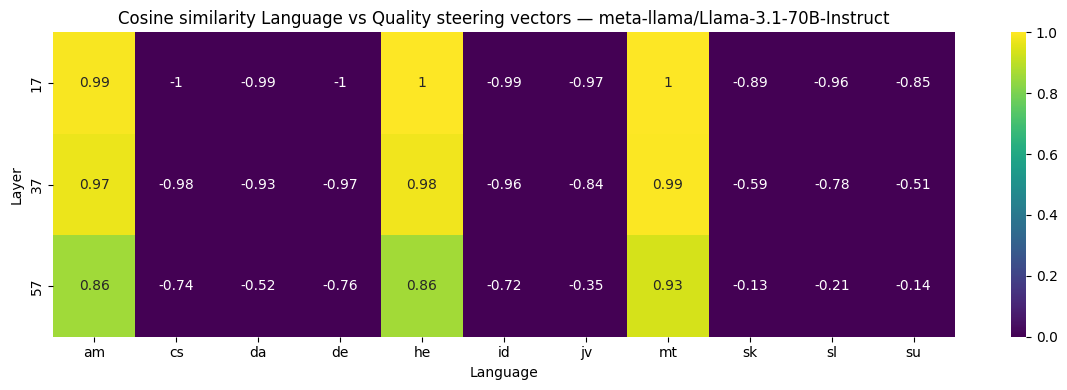}
    \caption{Llama 3.1 70b}
    \label{fig:another1}
\end{subfigure}
\hfill
\begin{subfigure}{0.4\textwidth}
    \centering
    \includegraphics[width=\textwidth]{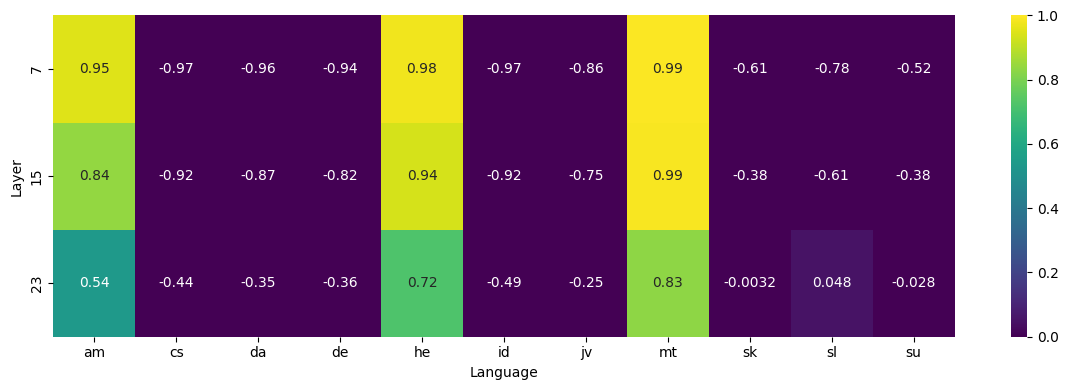}
    \caption{Llama 3.1 8b}
    \label{fig:another2}
\end{subfigure}

\caption{Cosine similarity between quality and language vectors for different LLMs used in this study.}
\label{fig:combined-corrs}
\end{figure*}
\paragraph{Steering vectors applied with few-shot prompts}
The aggregated few-shot results are shown in Table~\ref{tab:steering_results_icl}, with task-specific results provided in Appendix~\ref{sec:appendix_per_task_f1}. Compared to the zero-shot setting, the advantage of \textit{Quality} steering over \textit{Language} steering becomes less pronounced, and the strong preference for early-layer interventions weakens when demonstrations are included in the prompt. Nevertheless, early-layer \textit{Quality} steering remains the most consistent approach overall. Notably, \textit{Language} steering has no clear winner for layer selection, while \textit{Quality} steering yields best performance gains at earlier layers.

\textit{Quality} steering was again most effective at early layers, improving downstream performance in 97.73\% of cases, compared to 81.82\% and 86.36\% for middle and late layers. \textit{Language} steering showed an inverse trend, with gains in 77.27\%, 84.09\%, and 97.73\% of cases, respectively. Average gains decreased for most models, except for Llama-3.1-70B and Gemma 2 27B, which appeared to benefit from combining steering with few-shot demonstrations. Statistically significant improvements for \textit{Quality} steering occurred in 36.36\% of cases for early layers, versus 21.59\% and 25\% for middle and late layers. For \textit{Language} steering, statistically significant gains were observed in 29.55\%, 31.82\%, and 26.14\% of cases, respectively. Together with the zero-shot results, these findings indicate that early-layer steering consistently outperforms deeper interventions. Per-task results are provided in Appendix~\ref{sec:appendix_per_task_f1}.

Among models, Gemma-2-27B and Llama-3.1-70B were the most robust, showing predominantly positive performance shifts. In contrast, Gemma-2-9B behaved more inconsistently in the few-shot setting, exhibiting substantially more negative outcomes than in the zero-shot experiments.

\paragraph{Comparison of zero-shot vs. few-shot}
Comparing zero-shot and few-shot prompting reveals a clear saturation effect. In the zero-shot setting, steering often produced substantial gains, including double-digit F1 improvements, whereas the same configurations yielded smaller improvements under few-shot prompting. This suggests that few-shot demonstrations already shift the model toward the ``human-authored'' manifold, reducing the additional effect of steering. Nevertheless, \textit{Quality} steering applied to early layers continued to provide consistent improvements.

The strong ``early layers are best'' trend observed in zero-shot generation also becomes less rigid in the few-shot setting. Larger models, in particular, showed greater robustness and occasionally benefited from steering at deeper layers.

Overall, our results indicate that activation steering is most impactful in zero-shot settings, where it can compensate for the absence of stylistic guidance in the prompt. In few-shot scenarios, steering remains beneficial, though with smaller gains. This shows that, regardless of the original prompt used, steering is very often beneficial for downstream model performance when applied to the used prompting technique, with \textit{Quality} steering offering higher and more consistent increases.

\subsection{Using Only Best Alpha Per Layer}\label{sec:only_best_alpha}
Our evaluation setup uses a validation split of the given dataset and language to select the best alpha on the given layer via validation loss. To investigate whether we can select a single alpha value per layer, steering method, and generator LLM, we use the same alpha value for the entire layer (e.g., all data generated using Llama3 8b had the same alpha applied for the \textit{Quality} steering vector at layer 7). The results for the zero-shot and few-shot settings can be found in Tables~\ref{tab:steering_results_best_alpha_per_layer} and~\ref{tab:steering_results_icl_alpha_per_layer}, respectively.

For the zero-shot setting, early layers proved the most effective intervention points again. Across all task-aggregated experiments, \textit{Quality} steering improved performance in 79.54\% of cases for early layers, compared to 68.18\% and 70.45\% for middle and late layers. \textit{Language} steering showed a similar trend, with gains in 72.73\%, 70.45\%, and 63.64\% of cases, respectively. This is an absolute decrease in the number of positive cases by approximately 20\% to 28\% compared to using the best alpha per case. Statistically significant improvements have dropped by approximately 10\% on average.

We can see a similar trend in the few-shot setting, where the absolute number of positive increases drops by 25\% to 30\%. However, the number of statistically significant cases remains similar. For languages, we can see that using the correct alpha for generation becomes even more important — languages like Danish, Hebrew, or Javanese yield overall slight decreases in model performance. This underscores the importance of validation splits during evaluation for selecting the best alpha.

\section{Evaluating Diversity of Generated Data}

As shown in Section~\ref{sec:downstream_model_res}, both \textit{Quality} and \textit{Language} steering generally improve downstream model performance. To better understand these gains, we additionally analyze the diversity of the generated data, which has previously been linked to improved downstream robustness and performance~\cite{cegin-etal-2026-rose}.

We evaluate four diversity metrics: (1) \textit{Lexical diversity}, measuring the ratio of unique character 3-grams; (2) \textit{Embedding diversity}, capturing the average pairwise cosine distance between text embeddings; (3) \textit{Homogeneity}, measuring the structural consistency of the embedding distribution; and (4) \textit{Isocontour radius}, estimating the overall spread of embedding representations. Details of the computation are provided in Appendix~\ref{sec:appendix_diversity_metrics}.

We compute these metrics across all datasets. The aggregated results are in Table~\ref{tab:diversity_results} for zero-shot and in Table~\ref{tab:diversity_results_icl} for few-shot prompting, with additional visualizations in Appendix~\ref{sec:appendix_detailed_div}.

In the zero-shot setting, steering generally increases output diversity, particularly for smaller models and when applied to early layers. Most configurations show gains across all diversity metrics, with the notable exception of Gemma-2-27B, which also showed limited responsiveness in downstream evaluation. Increases in lexical and embedding diversity indicate that steering reduces repetitive generations and broadens semantic coverage, while higher homogeneity suggests that this increased diversity remains structurally coherent rather than noisy. Early-layer steering appears especially effective because it influences representations before higher-level semantic processing has stabilized, allowing the model to propagate the intervention throughout the generation. This aligns with the stronger downstream performance improvements observed for earlier-layer steering.

In the few-shot setting, diversity gains become smaller and less consistent, suggesting that demonstrations already constrain the model’s latent space. In some cases, deeper-layer \textit{Quality} steering even reduces diversity, particularly for Gemma-2-27B. Smaller models also tend to show decreases in homogeneity under few-shot steering, whereas larger models such as Llama-3.1-70B often maintain or improve structural consistency. Interestingly, \textit{Language} steering frequently produces larger increases in embedding diversity than \textit{Quality} steering in Llama models, possibly because language vectors are less aligned with the representations already induced by few-shot examples.

Although increased diversity does not always guarantee better downstream performance, our results show that steering vectors consistently encourage the generation of more varied data without explicit diversity optimization. This likely contributes to the improved robustness and downstream effectiveness that we observed.

\section{Similarity of \textit{Language} and \textit{Quality} Vectors}
We provide cosine similarity between \textit{Language} and \textit{Quality} vectors per LLM in Figure~\ref{fig:combined-corrs}.

We see a distinct difference between LLMs in Gemma and Llama families. For Gemma 2 9b, most languages show low to moderate positive similarity, except Maltese. The polarity is much stronger in Gemma 2 27b. In contrast, both Llama models show a remarkably similar, highly polarized pattern for the same languages. Amharic, Hebrew, and Maltese are all nearly perfectly positively correlated, while other languages are nearly perfectly negatively correlated.

The languages appear to cluster into two distinct groups based on vector alignment. For the positive cluster, the \textit{Quality} and \textit{Language} are essentially in the same direction in the model's latent space. This explains why Amharic showed strong F1 gains under both steering types in most of our experiments. Notably, all of these languages are Semitic, which could indicate that for specific language groups the model sees \textit{Quality} and \textit{Language} as nearly identical. For the majority of languages, especially European and Southeast Asian ones, the \textit{Quality} direction is often the mathematical opposite of \textit{Language} direction in Llama models.

The polarities ($-1.0$ or $1.0$) are most consistent in the early layers across all models. This reinforces our earlier conclusion that early-layer interventions are more effective because they target these highly defined, clear directions before the representations become more ``mixed'' or diffuse in later layers. The polarities indicate that the \textit{Quality} direction is often the inverse of the \textit{Language} direction. While both of these steering vectors can help in downstream model performance, this indicates that \textit{Quality} steering vectors are (in general) not tied to a specific language and seem to be language-agnostic. This also explains the volatility of improvements from \textit{Language} steering, as for most of the languages, it is steering the model in the exact opposite direction of the \textit{Quality} steering.

\section{LLM-as-Judge Evaluation}
We also evaluated the generated data in terms of label adherence, language adherence, and fluency via LLM-as-Judge using Gemma 3 27b. The aggregated results and details can be found in Appendix~\ref{sec:appendix_llm_as_juge}.

In general, we see only slight changes for \textit{Quality} steering for both zero- and few-shot setups. We see that the early layers lead to often smaller relative decreases in all categories (in 61.1 \% of cases), as such steering seems to produce only noisier, but still valid, samples that are beneficial for downstream model performance. However, the decreases were not significant.

\textit{Language} steering had even more negative cases (66.66\%) and produced stronger outliers in the Llama 3.1 70b model in particular, where early layers lead to an approximate 10\% relative decrease in all 3 metrics. Similar changes can also be seen in other models, indicating that \textit{Language} steering can hurt the perceived validity (by the LLM) of the data.

\section{Conclusion}
In this work, we investigated activation steering for synthetic data generation in low-resource languages. Across four open-source LLMs, 11 languages, and two classification tasks, we showed that both \textit{Language} and \textit{Quality} steering can improve downstream performance while consistently increasing the diversity of generated data. Our results further demonstrate that steering is most effective when applied to earlier transformer layers, particularly in zero-shot settings. This finding is consistent with the cultural steering results of \citet{ghussin2026dfkimltsemeval2026task7}, who likewise observe that the effectiveness of vectors is strongly layer-dependent, with earlier-layer steering yielding the most reliable gains.

Overall, \textit{Quality} steering proved to be the more reliable approach, suggesting that steering models toward a ``human-authored'' notion is especially beneficial for synthetic data generation. We additionally found that the relationship between Language and \textit{Quality} vectors is highly model- and language-dependent, highlighting structured multilingual representations in LLM latent spaces.

These findings position activation steering as an efficient alternative or complement to few-shot prompting for multilingual synthetic data generation, especially in low-resource settings.

\section*{Acknowledgments}
This work was partially supported by AI-CODE
(GA No. 101135437), by the European Union under the project lorAI - Low Resource Artificial Intelligence, GA No. 101136646, by NextGenerationEU through the Recovery and Resilience Plan for Slovakia under the project No. 09I01-03-V04-00068 (GEPERO), and by the German Federal Ministry of Research, Technology and Space (BMFTR) as part of the project TRAILS (01IW24005).

This work was computationally supported by the Ministry of Education, Youth and Sports of the Czech Republic through the e-INFRA CZ (ID:90254) and by the use of the supercomputer PERUN, with the support of the European Union from the funds of the Recovery and Resilience Plan of the Slovak Republic within the framework of project No. 17I03-04-P03-00001.

\section*{Limitations}
While our work demonstrates the efficacy of activation steering for low-resource synthetic data generation, several limitations remain to be addressed in future work.

\textbf{Task and Generative Scope:} Our evaluation is bounded by two text classification tasks: sentiment analysis and topic classification. While classification is a standard benchmark for evaluating synthetic data utility~\cite{anikina-etal-2025-rigorous, cegin-etal-2026-rose}, it does not fully test the generative boundaries of activation steering.

\textbf{Dependence on Machine Translation Quality:} The extraction of our Quality steering vector relies on constructing contrastive text pairs via two rounds of backtranslation using the NLLB-200-600M distilled model. We did not investigate how many specific rounds or which translation models are the best.

\textbf{Hyperparameter Sensitivity and Model Dependency:} Our results highlight that the optimal steering intensity ($\alpha$) and layer selection are heavily model- and language-dependent. For instance, the optimal scaling factors vary by orders of magnitude between the Gemma and Llama families (e.g., $\alpha=100.0$ for Gemma 2 27B vs. $\alpha=1.0$ for Llama 3.1 70B). At present, we lack a predictive, analytical framework to establish the perfect layer and multiplier combination a priori, limiting the immediate plug-and-play deployability of this method.

\textbf{LLM-as-Judge:} Our evaluation used LLM-as-Judge, which can be misleading and produce invalid annotations.

\bibliography{custom}

\appendix

\section{Ethical Considerations}
\label{sec:ethical}
Based on a thorough ethical assessment performed on the basis of intra-institutional ethical guidelines and checklists tailored to the use of data and algorithms, we see no ethical concerns pertaining directly to the conduct of this research. Although the production of new data through LLMs bears several risks, such as the introduction of biases, the small size of the produced dataset, sufficient for experimentation, is, at the same time, insufficient for any major machine learning endeavours where such biases could be transferred.

We follow the license terms for all the models and datasets we used (such as the one required for the use of the Llama-3.1 model) – all models and datasets allow their use as part of the research.

We used generative AI for grammatical checks and fixes of the paper, and have verified all the texts changed by the generative AI.

\section{Computational Resources}\label{sec:computational_resources}
Our experiments were run on a computational cluster with H200 GPUs, 2x AMD EPYC 9745 CPUs, and 128 GB of RAM. Our generation experiments took approximately 6-8 minutes per one combination of LLM, steering vector, layer, task, language, and alpha value. Given that we had a total of 2112 combinations, our generation experiments took approximately 260 GPU hours. For the fine-tuning experiments, we used approximately 240 GPU hours. In total, for our experiments, we used approximately 500 GPU hours.

\section{Linear Probe Details}
\label{sec:appendix_linear_probe}
To quantify the linear separability of the quality signal within the model's latent representations, we implemented a layer-wise probing diagnostic using logistic regression. For each input sequence in our human-authored and backtranslated distributions, we extracted the hidden states from the model's residual stream. We then applied mean pooling across non-padding tokens to generate representative sequence-level embeddings for every layer. These activations were sanitized by clipping outliers and then L2-normalized to facilitate stable classifier training. Using a stratified 80/20 train-test split, we trained a logistic regression classifier on the embeddings of each layer independently and measured the resulting classification accuracy. As illustrated in Figure~\ref{fig:linear_probe_res}, the layers exhibiting peak accuracy represent the regions where the quality concept is most robustly and linearly encoded, providing the optimal candidates for activation steering interventions.

\section{Steering Vectors Computation Details}\label{appendix:sec_steer_vec_comp}
\subsection{\textit{Language} Steering Vector}
To construct the language steering vectors, we employ a one-vs-rest contrastive methodology across the residual stream of all transformer blocks. This approach ensures that the resulting vector captures features specific to the target language rather than shared cross-lingual properties or general model priors.

For each target language $L \in \mathcal{L}$, where $\mathcal{L}$ is our set of 11 typologically diverse languages, we pass the dataset through the model. Using the TransformerLens framework, we extract the residual stream activations for every layer $l$. We compute the mean activation vector $\mu_{l,L}$ by averaging across all non-padding tokens $N_L$ in the language-specific dataset:$$\mu_{l,L} = \frac{1}{N_L} \sum_{i,j} \mathbf{a}_{l,i,j} \cdot m_{i,j}$$. This process results in a language-specific centroid in the latent space for every layer of the model. 

To isolate the linguistic identity of language $L$, we define the steering vector $\mathbf{v}_{l,L}$ as the difference between the language's mean activation and the mean activation of all other languages in our study:$$\mathbf{v}_{l,L} = \mu_{l,L} - \frac{1}{|\mathcal{L}| - 1} \sum_{L' \in \mathcal{L}, L' \neq L} \mu_{l,L'}$$ By contrasting the target language against a global multilingual mean, we filter out common semantic and structural activations, focusing the vector on the unique "fingerprint" of the target language.

Finally, we apply layer-wise L2 normalization to the difference vectors:$$\hat{\mathbf{v}}_{l,L} = \frac{\mathbf{v}_{l,L}}{\|\mathbf{v}_{l,L}\|_2}$$This normalization step is critical for maintaining stability during the generation phase, as it ensures that the steering magnitude $\alpha$ remains consistent across different layers and languages regardless of the original activation scales.

\subsection{\textit{Quality} Steering Vector}
For each target language $L \in \mathcal{L}$, we utilize two matching corpora representing a binary quality dimension $\mathcal{D} = \{d_{\text{good}}, d_{\text{bad}}\}$, where $d_{\text{good}}$ consists of authentic, human-authored text and $d_{\text{bad}}$ comprises corresponding backtranslations. We process each corpus through the model independently to capture the token-level post-residual block hidden states $\mathbf{a}_{l,i,j}$ across all layers $l$. The direct mean activation vector $\mu_{l,L,d}$ for an attribute dimension $d \in \mathcal{D}$ within language $L$ is accumulated over its respective unpadded token count $N_{L,d}$:
$$\mu_{l,L,d} = \frac{1}{N_{L,d}} \sum_{i,j} \mathbf{a}_{l,i,j} \cdot m_{i,j}$$

To extract a clean signal, the raw quality direction $\mathbf{v}_{l,L,\text{qual}}$ is isolated by subtracting the centroid of the corrupted dataset from the centroid of the human-authored dataset:
$$\mathbf{v}_{l,L,\text{qual}} = \mu_{l,L,d_{\text{good}}} - \mu_{l,L,d_{\text{bad}}}$$
By establishing this explicit pairwise vector difference within the same language, task-neutral semantic properties cancel out, leaving a vector that should isolate the geometric directional shift toward more human-like data.

Following the same stability protocol as our language vectors, we enforce a layer-wise L2 normalization step to yield the final quality steering direction:
$$\hat{\mathbf{v}}_{l,L,\text{qual}} = \frac{\mathbf{v}_{l,L,\text{qual}}}{\|\mathbf{v}_{l,L,\text{qual}}\|_2}$$

\section{Language Abbreviations}\label{sec:lang_abbreviations}
Language abbreviations for each language used in the study can be found in Table~\ref{tab:lang_abbreviations}.
\begin{table}[h]
\centering
\footnotesize
\begin{tabular}{ll}
\toprule
Code &  Language \\
\midrule
    am & Amharic \\
    cs & Czech \\
    de & German \\
    da & Danish \\
    he & Hebrew \\
    id & Indonesian \\
    jv & Javanese \\
    mt & Maltese \\
    sk & Slovak \\
    sl & Slovenian \\
    su & Sundanese \\
\bottomrule
\end{tabular}
\caption{Language abbreviations.}
\label{tab:lang_abbreviations}
\end{table}

\section{Backtranslation Details}\label{sec:appendix_backtranslation}
For the backtranslation and creation of parallel synthetic data to human-authored data, we employ the NLLB 600M distilled model~\footnote{\url{https://huggingface.co/facebook/nllb-200-distilled-600M}}. To ensure enough distinction between the human-authored and synthetic data, we use two rounds of backtranslation from the source language to target language 1, then to target language 2, back to target language 1, and then back to the source language. We use English as target language 1 and Chinese as target language 2 as the two target languages, due to their resourcefulness. We also investigated additional corruptions of the synthetic data (e.g., random token swaps), but decided against it given the sufficient linear probe results from Section~\ref{sec:linear_probe}.

\section{Downstream Fine-tuning of XLM-R}\label{sec:appendix_downstream_finetuning}
For the downstream evaluation, we fine-tune the XLM-R \cite{DBLP:journals/corr/abs-1911-02116} \textit{FacebookAI/xlm-roberta-base} model with a batch size of 16 and employ early stopping with a patience of 10 epochs to prevent overfitting. We perform hyperparameter optimisation to determine the optimal learning rate and set it to 1e-5. \textit{AdamW} is used as an optimiser. We balance the generated datasets to have the same number of samples per label. We perform finetuning 20 times with different seeds for each generated data. We normalise all inputs by converting them to lowercase and removing punctuation.

\section{Diversity Metrics}\label{sec:appendix_diversity_metrics}
To quantitatively evaluate the geometric and linguistic characteristics of the generated text, we employ four distinct diversity metrics. Before metric computation, all generated texts undergo a standardized normalization pipeline, which normalizes text into the NFKD variant, strips diacritics and non-textual symbols (such as emojis), and collapses consecutive whitespaces.

For dense vector calculations, text sequences are mapped into a latent embedding space using the Qwen3-Embedding-4B~\footnote{\url{https://huggingface.co/Qwen/Qwen3-Embedding-4B}} encoder model.

\paragraph{Lexical diversity:} Calculated as the ratio of unique character-level trigrams to total trigrams across the corpus. It captures surface-level variety; higher scores indicate diverse, natural phrasing, while lower scores signify repetitive or formulaic text.

\paragraph{Embedding diversity:} Computed by grouping the dataset by task labels, calculating the mean pairwise cosine distance within each group $g$, and taking the macro-average across all unique labels $G$. It measures categorical semantic dispersion, showing whether a model generates a broad range of distinct concepts or collapses into a narrow thematic subspace.

\paragraph{Isocontour radius}~\cite{lai-etal-2020-diversity}: Determines the geometric boundary of the generated dataset by measuring the Euclidean distance of all individual embeddings to the global dataset centroid $\mu$. The metric outputs the 90th percentile of this distance distribution. It represents the total volume of the latent operational envelope; a larger radius indicates that the steering intervention has pushed the generations into wider, peripheral regions of the latent space.

\paragraph{Homogeneity}~\cite{lai-etal-2020-diversity}:  Evaluates structural density uniformity using a similarity-based Kernel Density Estimation (KDE). It applies an RBF kernel ($\sigma = 0.5$) to the pairwise cosine similarities to compute a localized density score for each point. It tracks structural consistency across the manifold; a lower score indicates a highly uniform, evenly distributed layout, whereas a higher score signals structural imbalance and tight data clustering.

\section{Prompt Templates and Generation Details}\label{sec:appendix_prompt_templates}

For generating synthetic data, we used this general prompt template for the \textbf{zero-shot} setting:

\textbf{Topic:} \textit{"Generate a long wiki-style sentence on the topic of {label} in {language} language. Output only the text in {language}."}

\textbf{Sentiment:} \textit{"Generate a review with {label} sentiment in {language} language. Output only the text and nothing else."}

For the \textbf{few-shot} setting, we used 5 shots in the target language and label, with these prompts:

\textbf{Topic:} \textit{"Generate a long wiki style sentence on the topic of {label} in {language} language based on examples. Output only the text in {language}. Examples: {examples}"}

\textbf{Sentiment:} \textit{"Generate a review with {label} sentiment in {language} language based on examples. Output only the text and nothing else. Examples: {examples}"}

Sampling parameters used for generation were: \textit{temperature=0.8,  top\_p=0.9, max\_tokens=512, freq\_penalty=0.0}. We excluded duplicates to ensure unique samples were collected.

Specific versions of LLMs used for generations were: Llama-3.1-70b-instruct~\footnote{\url{https://huggingface.co/meta-llama/Llama-3.1-70B-Instruct}}, Llama-3.1-8b-instruct~\footnote{\url{https://huggingface.co/meta-llama/Llama-3.1-8B-Instruct}}, gemma-2-27b-it~\footnote{\url{https://huggingface.co/google/gemma-2-27b-it}}, gemma-2-9b-it~\footnote{\url{https://huggingface.co/google/gemma-2-9b-it}}. The LLMs were used with \textit{bfloat16} precision.

\section{Baseline Results for Zero- and Few-shot Setting}
\label{sec:appendix_baseline_f1}
We provide the baseline result for topic and sentiment detection for both zero-shot in Table~\ref{tab:baseline_zero_shot} and for the few-shot setting in Table~\ref{tab:baseline_few_shot}.

\section{Alpha Values Analysis}
\label{sec:appendix_alpha_analysis}

We provide an analysis of alpha values in all settings in Table~\ref{tab:results_zero_shot_alphas} for zero-shot and in Table~\ref{tab:results_few_shot_alphas} for the few-shot setting.

For generic language vectors, increasing alpha often behaves like a poison pill. As alpha increases, variance widens significantly, and medians frequently trend downward. \textit{Quality} vectors handle high alpha values much better. Instead of collapsing, higher alphas either monotonically improve performance or reach a safe plateau with tight variances. We see the best performance for alpha values at \textit{50} or \textit{75} for Gemma 2 models and \textit{2} or \textit{3} for Llama3.1 models, indicating that the best downstream model performance comes at moderate-to-strong steering. 

Larger models are more brittle to bad steering, as for both Llama 3.1 70b and Gemma 2 27b, choosing the wrong vector (\textit{Language}) or the wrong layer/alpha combination leads to dramatic, sweeping performance drops, most notably for high alpha values in the Llama3.1 70b model. They require precise steering. On the other hand, Llama 3.1 8b and Gemma 2 9b show remarkably clean, uniform positive shifts across almost all alpha values in their early and middle layers, indicating that smaller models might have more centralized "quality pathways" that are easier to uniformly amplify.

\section{Downstream Model Performance Per Task}
\label{sec:appendix_per_task_f1}
We provide a per-task breakdown for the zero-shot setting for the topic detection task in Table~\ref{tab:steering_results_topic} and for the sentiment detection task in Table~\ref{tab:steering_results_sent}. For the few-shot setting, the topic detection task can be found in Table~\ref{tab:steering_results_icl_topic} and the sentiment detection task in Table~\ref{tab:steering_results_icl_sent}. All of these tables also contain visualizations showing which specific cases had statistically significant increases in downstream model performance.

In terms of the zero-shot setting, we see similar results to those observed in Section~\ref{sec:downstream_model_res}, as the early layers \textit{Quality} steering vector outperforms other approaches, except for the Llama 3.1 70b model, where \textit{Language} vectors applied to earlier layers slightly outperform it. The best performing method is steering at earlier layers, as it offers an increase in downstream model performance in 79.55\% of cases for both topic and sentiment detection. Gemma models see a larger increase in sentiment detection task, while Llama models see a larger increase in downstream model performance in the topic detection task.

For the few-shot setting, again, the \textit{Quality} steering vectors outperform the \textit{Language} steering vectors, in particular at the earlier layers. For topic detection, steering via early layer \textit{Quality} vectors results in increased downstream performance in 70.45\% of cases for topic detection and 84.09\% of cases for sentiment detection. The noisier sentiment detection seems to benefit more from the "human-likeness" introduced by the \textit{Quality} steering vectors. 

\section{Downstream Model Performance With Alpha Selected Per Layer}
\label{sec:appendix_alpha_per_layer}
We provide results for cases where the best alpha is selected not per case, but per layer of the given LLM generator and steering method. Results can be found in Table~\ref{tab:steering_results_best_alpha_per_layer} for zero-shot and Table~\ref{tab:steering_results_icl_alpha_per_layer}.

\section{Detailed Downstream Model Visualizations}
\label{sec:appendix_detailed_downstream_f1}
We provide detailed visualizations for different languages, LLMs, steering vectors, and layers in Table~\ref{tab:results_zero_shot_all} for zero-shot and in Table~\ref{tab:results_few_shot_all} for few-shot.

\section{Detailed Diversity Visualizations}
\label{sec:appendix_detailed_div}
We provide a detailed diversity visualization for each of our 4 metrics and approaches for both the zero-shot baseline and the few-shot baseline in Tables~\ref{tab:diversity_all_zero} and~\ref{tab:diversity_all_few}, respectively.

\section{LLM-as-Judge Setup}
\label{sec:appendix_llm_as_juge}
For the LLM-as-a-judge setup, we leveraged the Gemma-3-27b-it model~\footnote{\url{https://huggingface.co/google/gemma-3-27b-it}}, with temperature set to 0 and maximum new tokens set to 64. Aggregated results per language and task can be found in Table~\ref{tab:both_perf}. We used the following prompt:

\textit{You are an expert linguistic evaluator. Analyze the following synthetic text sample and grade it strictly according to the defined 1-5 scale for Fluency, Label Adherence, and Language Adherence.}

\textit{EVALUATION CRITERIA:
1. Fluency: Grammatical correctness and natural flow.
2. Label Adherence: How accurately the text matches the intended target label, prompt, or concept.
3. Language Adherence: How purely the text uses the designated target language and its correct script.}

\textit{INPUT DATA:
- Target Language: {language}
- Target Label/Concept: {label description in natural language}
- Generated Text: {sample}}

\textit{Output your scores strictly as a JSON object matching the requested schema. Return ONLY the raw JSON object. Do not include markdown code block formatting or any introductory/concluding text.}

\begin{table*}[t!]
\centering
\caption{Relative performance metric differences in \% (Fluency, Language Adherence, Label Adherence) across models, steering methods, and layers, aggregated over tasks and languages compared to the baseline of no steering. Using the best alpha from downstream model performance. Relative \textcolor{cbBlue}{positive} and \textcolor{cbOrange}{negative} changes.}
\begin{subtable}[t]{0.48\textwidth}
\centering
\tiny
\caption{\textbf{Zero-shot} setup}
\label{tab:perf_results_zero}
\begin{tabular}{llccc}
\toprule
 &  & Fluency & LangAdh. & LabelAdh. \\
Model + Steering & Layer &  &  &  \\
\midrule
\midrule
 \multirow[c]{3}{*}{Gemma 2 27b (lang.)} & L10 & {\cellcolor{cbBlue!10}} 0.34 & {\cellcolor{cbOrange!10}} -0.25 & {\cellcolor{cbOrange!10}} -0.04 \\
  & L22 & {\cellcolor{cbBlue!10}} 0.28 & {\cellcolor{cbOrange!10}} -0.26 & {\cellcolor{cbBlue!10}} 0.05 \\
  & L34 & {\cellcolor{cbBlue!10}} 0.47 & {\cellcolor{cbBlue!10}} 0.19 & {\cellcolor{cbBlue!10}} 0.10 \\
\midrule
 \multirow[c]{3}{*}{Gemma 2 27b (qual.)} & L10 & {\cellcolor{cbBlue!11}} 0.69 & {\cellcolor{cbBlue!10}} 0.01 & {\cellcolor{cbBlue!10}} 0.31 \\
  & L22 & {\cellcolor{cbBlue!10}} 0.18 & {\cellcolor{cbBlue!10}} 0.09 & {\cellcolor{cbOrange!10}} -0.13 \\
  & L34 & {\cellcolor{cbBlue!10}} 0.38 & {\cellcolor{cbBlue!10}} 0.10 & {\cellcolor{cbOrange!10}} -0.28 \\
\bottomrule
\end{tabular}
\end{subtable}
\hfill
\begin{subtable}[t]{0.48\textwidth}
\centering
\tiny
\caption{\textbf{Few-shot} setup}
\label{tab:perf_results_icl}
\begin{tabular}{llccc}
\toprule
 &  & Fluency & LangAdh. & LabelAdh. \\
Model + Steering & Layer &  &  &  \\
\midrule
\midrule
 \multirow[c]{3}{*}{Gemma 2 27b (lang.)} & L10 & {\cellcolor{cbBlue!11}} 0.84 & {\cellcolor{cbOrange!10}} -0.08 & {\cellcolor{cbBlue!11}} 0.85 \\
  & L22 & {\cellcolor{cbBlue!11}} 0.70 & {\cellcolor{cbOrange!10}} -0.09 & {\cellcolor{cbOrange!10}} -0.00 \\
  & L34 & {\cellcolor{cbBlue!10}} 0.36 & {\cellcolor{cbOrange!10}} -0.35 & {\cellcolor{cbOrange!10}} -0.17 \\
\midrule
 \multirow[c]{3}{*}{Gemma 2 27b (qual.)} & L10 & {\cellcolor{cbOrange!11}} -0.81 & {\cellcolor{cbOrange!10}} -0.29 & {\cellcolor{cbOrange!11}} -0.56 \\
  & L22 & {\cellcolor{cbOrange!10}} -0.43 & {\cellcolor{cbOrange!10}} -0.14 & {\cellcolor{cbOrange!10}} -0.40 \\
  & L34 & {\cellcolor{cbOrange!11}} -0.51 & {\cellcolor{cbOrange!10}} -0.32 & {\cellcolor{cbBlue!10}} 0.07 \\
\bottomrule
\end{tabular}
\end{subtable}
\label{tab:both_perf}
\end{table*}

\begin{table*}[t!]
\caption{Mean F1 scores across languages and tasks for baseline \textbf{zero-shot} setting with no steering.}
\small
\label{tab:baseline_zero_shot}
\resizebox{\textwidth}{!}{%
\begin{tabular}{llccccccccccc||c}
\toprule
 & Language & am & cs & da & de & he & id & jv & mt & sk & sl & su & Average \\
LLM & Task &  &  &  &  &  &  &  &  &  &  &  &  \\
\midrule
\midrule
\multirow[c]{2}{*}{Gemma 2 27b} & Sentiment & 74.53 & 83.55 & 95.27 & 64.10 & 75.69 & 84.07 & 45.47 & 54.82 & 69.53 & 73.23 & 67.35 & 71.60 \\
 & Topic & 48.28 & 59.26 & 68.93 & 64.82 & 62.99 & 65.99 & 60.43 & 41.33 & 64.46 & 63.84 & 61.01 & 60.12 \\
\midrule
\multirow[c]{2}{*}{Gemma 2 9b} & Sentiment & 79.52 & 80.78 & 94.97 & 71.50 & 71.98 & 87.07 & 62.53 & 53.12 & 66.06 & 68.64 & 73.58 & 73.62 \\
 & Topic & 49.05 & 61.48 & 64.54 & 62.06 & 60.26 & 66.35 & 60.19 & 38.14 & 60.98 & 63.56 & 62.40 & 59.00 \\
\midrule
\multirow[c]{2}{*}{Llama3.1 70b} & Sentiment & 77.21 & 82.94 & 91.72 & 68.91 & 74.46 & 86.87 & 61.79 & 57.17 & 71.04 & 79.69 & 74.29 & 75.10 \\
 & Topic & 50.21 & 75.82 & 76.57 & 73.54 & 67.18 & 70.16 & 49.64 & 39.01 & 71.93 & 76.71 & 61.85 & 64.78 \\
\midrule
\multirow[c]{2}{*}{Llama3.1 8b} & Sentiment & 53.66 & 85.10 & 94.05 & 71.18 & 76.77 & 87.93 & 62.10 & 51.78 & 82.53 & 81.89 & 78.88 & 75.08 \\
 & Topic & 33.82 & 71.06 & 71.55 & 75.84 & 71.24 & 68.04 & 62.72 & 46.41 & 71.65 & 69.79 & 65.18 & 64.30 \\
\bottomrule
\end{tabular}
}
\end{table*}

\begin{table*}[t!]
\caption{Mean F1 scores across languages and tasks for baseline \textbf{few-shot} setting with no steering.}
\small
\label{tab:baseline_few_shot}
\resizebox{\textwidth}{!}{%
\begin{tabular}{llccccccccccc||c}
\toprule
 & Language & am & cs & da & de & he & id & jv & mt & sk & sl & su & Average \\
LLM & Task &  &  &  &  &  &  &  &  &  &  &  &  \\
\midrule
\midrule
\multirow[c]{2}{*}{Gemma 2 27b} & Sentiment & 74.93 & 68.14 & 91.64 & 61.12 & 58.20 & 80.64 & 60.01 & 59.25 & 75.24 & 59.76 & 80.36 & 69.94 \\
 & Topic & 56.70 & 76.64 & 75.06 & 79.39 & 69.56 & 74.96 & 67.42 & 48.30 & 76.30 & 73.02 & 66.14 & 69.41 \\
\midrule
\multirow[c]{2}{*}{Gemma 2 9b} & Sentiment & 81.69 & 75.47 & 87.32 & 59.17 & 55.99 & 87.20 & 54.91 & 57.81 & 72.60 & 82.43 & 78.87 & 72.13 \\
 & Topic & 56.14 & 76.33 & 74.82 & 76.07 & 73.53 & 74.62 & 62.76 & 51.02 & 77.48 & 74.51 & 68.82 & 69.65 \\
\midrule
\multirow[c]{2}{*}{Llama3.1 70b} & Sentiment & 82.72 & 83.30 & 94.68 & 56.36 & 70.67 & 77.89 & 56.34 & 57.13 & 65.10 & 73.90 & 74.65 & 72.07 \\
 & Topic & 62.65 & 77.92 & 75.50 & 80.55 & 74.85 & 81.03 & 69.51 & 48.74 & 78.90 & 76.78 & 68.74 & 72.29 \\
\midrule
\multirow[c]{2}{*}{Llama3.1 8b} & Sentiment & 76.53 & 81.48 & 85.74 & 69.01 & 78.52 & 87.20 & 56.67 & 55.94 & 80.13 & 72.16 & 80.24 & 74.87 \\
 & Topic & 52.06 & 79.78 & 76.74 & 76.04 & 74.93 & 76.01 & 69.14 & 48.15 & 77.94 & 76.30 & 68.11 & 70.47 \\
\bottomrule
\end{tabular}
}
\end{table*}

\begin{table*}
\caption{Mean F1 difference across languages, models, steering methods, and layers for \textbf{topic} detection, with \textcolor{cbBlue}{positive} and \textcolor{cbOrange}{negative} differences compared to \textbf{zero-shot} prompt with no steering. Cells with * denote statistically significant increases over baseline.}
\label{tab:steering_results_topic}
\tiny
\resizebox{\textwidth}{!}{%
\begin{tabular}{lllccccccccccc||c}
\toprule
Method & LLM & Layer & am & cs & da & de & he & id & jv & mt & sk & sl & su & Average \\
\midrule
\midrule
\multirow[c]{12}{*}{\textit{Language}} & \multirow[c]{3}{*}{Gemma 27b} & L10 & {\cellcolor{cbBlue!26}} 3.35* & {\cellcolor{cbBlue!16}} 1.30 & {\cellcolor{cbOrange!16}} -1.23 & {\cellcolor{cbBlue!16}} 1.31 & {\cellcolor{cbOrange!10}} -0.13 & {\cellcolor{cbBlue!28}} 3.69* & {\cellcolor{cbOrange!14}} -0.81 & {\cellcolor{cbBlue!15}} 1.12 & {\cellcolor{cbBlue!14}} 0.81 & {\cellcolor{cbBlue!10}} 0.01 & {\cellcolor{cbBlue!13}} 0.74 & {\cellcolor{cbBlue!14}} 0.92 \\
 &  & L22 & {\cellcolor{cbBlue!31}} 4.32* & {\cellcolor{cbBlue!16}} 1.22 & {\cellcolor{cbBlue!11}} 0.21 & {\cellcolor{cbBlue!16}} 1.24 & {\cellcolor{cbBlue!12}} 0.47 & {\cellcolor{cbBlue!20}} 2.14 & {\cellcolor{cbOrange!13}} -0.76 & {\cellcolor{cbBlue!23}} 2.65 & {\cellcolor{cbBlue!17}} 1.40 & {\cellcolor{cbBlue!16}} 1.28 & {\cellcolor{cbBlue!11}} 0.29 & {\cellcolor{cbBlue!16}} 1.31 \\
 &  & L34 & {\cellcolor{cbBlue!42}} 6.49* & {\cellcolor{cbBlue!14}} 0.84 & {\cellcolor{cbBlue!13}} 0.65 & {\cellcolor{cbBlue!10}} 0.07 & {\cellcolor{cbBlue!17}} 1.40 & {\cellcolor{cbBlue!27}} 3.36* & {\cellcolor{cbOrange!17}} -1.54 & {\cellcolor{cbBlue!11}} 0.23 & {\cellcolor{cbBlue!17}} 1.51 & {\cellcolor{cbBlue!12}} 0.45 & {\cellcolor{cbBlue!18}} 1.69 & {\cellcolor{cbBlue!16}} 1.38 \\
  \cmidrule(lr){2-15}
 & \multirow[c]{3}{*}{Gemma 9b} & L9 & {\cellcolor{cbOrange!32}} -4.49 & {\cellcolor{cbBlue!37}} 5.34* & {\cellcolor{cbBlue!31}} 4.24* & {\cellcolor{cbBlue!24}} 2.80 & {\cellcolor{cbBlue!52}} 8.46* & {\cellcolor{cbOrange!11}} -0.27 & {\cellcolor{cbOrange!24}} -2.92 & {\cellcolor{cbBlue!35}} 5.03* & {\cellcolor{cbBlue!44}} 6.79* & {\cellcolor{cbBlue!34}} 4.82* & {\cellcolor{cbBlue!10}} 0.19 & {\cellcolor{cbBlue!23}} 2.73 \\
\bottomrule
\end{tabular}
}
\end{table*}

\begin{table*}
\caption{Mean F1 difference across languages, models, steering methods, and layers for \textbf{sentiment} detection, with \textcolor{cbBlue}{positive} and \textcolor{cbOrange}{negative} differences compared to a \textbf{zero-shot} prompt with no steering. Cells with * denote statistically significant increases over baseline.}
\label{tab:steering_results_sent}
\tiny
\resizebox{\textwidth}{!}{%
\begin{tabular}{lllccccccccccc||c}
\toprule
Method & LLM & Layer & am & cs & da & de & he & id & jv & mt & sk & sl & su & Average \\
\midrule
\midrule
\multirow[c]{12}{*}{\textit{Language}} & \multirow[c]{3}{*}{Gemma 27b} & L10 & {\cellcolor{cbBlue!18}} 2.13 & {\cellcolor{cbBlue!12}} 0.65 & {\cellcolor{cbOrange!10}} -0.08 & {\cellcolor{cbBlue!35}} 6.11* & {\cellcolor{cbOrange!11}} -0.47 & {\cellcolor{cbBlue!17}} 1.72 & {\cellcolor{cbBlue!29}} 4.51 & {\cellcolor{cbOrange!10}} -0.22 & {\cellcolor{cbBlue!37}} 6.59* & {\cellcolor{cbBlue!35}} 6.12 & {\cellcolor{cbBlue!64}} 12.98* & {\cellcolor{cbBlue!25}} 3.64 \\
 &  & L22 & {\cellcolor{cbBlue!18}} 2.12 & {\cellcolor{cbBlue!11}} 0.28 & {\cellcolor{cbOrange!11}} -0.37 & {\cellcolor{cbBlue!13}} 0.79 & {\cellcolor{cbBlue!11}} 0.26 & {\cellcolor{cbBlue!15}} 1.38 & {\cellcolor{cbBlue!10}} 0.10 & {\cellcolor{cbOrange!11}} -0.32 & {\cellcolor{cbBlue!36}} 6.39* & {\cellcolor{cbBlue!21}} 2.74 & {\cellcolor{cbBlue!51}} 9.94* & {\cellcolor{cbBlue!18}} 2.12 \\
 &  & L34 & {\cellcolor{cbBlue!19}} 2.20 & {\cellcolor{cbBlue!13}} 0.88 & {\cellcolor{cbOrange!10}} -0.20 & {\cellcolor{cbBlue!15}} 1.36 & {\cellcolor{cbBlue!11}} 0.26 & {\cellcolor{cbBlue!18}} 1.98 & {\cellcolor{cbBlue!39}} 7.11* & {\cellcolor{cbOrange!20}} -2.43 & {\cellcolor{cbBlue!30}} 4.96 & {\cellcolor{cbBlue!33}} 5.68* & {\cellcolor{cbBlue!55}} 10.71* & {\cellcolor{cbBlue!22}} 2.96 \\
   \cmidrule(lr){2-15}
 & \multirow[c]{3}{*}{Gemma 9b} & L9 & {\cellcolor{cbBlue!12}} 0.67 & {\cellcolor{cbBlue!23}} 3.21 & {\cellcolor{cbOrange!10}} -0.09 & {\cellcolor{cbBlue!14}} 1.17 & {\cellcolor{cbBlue!23}} 3.13 & {\cellcolor{cbBlue!13}} 0.82 & {\cellcolor{cbBlue!11}} 0.41 & {\cellcolor{cbBlue!19}} 2.26 & {\cellcolor{cbBlue!37}} 6.62* & {\cellcolor{cbBlue!69}} 14.21* & {\cellcolor{cbBlue!31}} 5.03* & {\cellcolor{cbBlue!24}} 3.40 \\
\bottomrule
\end{tabular}
}
\end{table*}

\begin{table*}
\caption{Mean F1 difference across languages, models, steering methods, and layers for \textbf{topic} detection, with \textcolor{cbBlue}{positive} and \textcolor{cbOrange}{negative} differences compared to a \textbf{few-shot} prompt with no steering. Cells with * denote statistically significant increases over baseline.}
\label{tab:steering_results_icl_topic}
\tiny
\resizebox{\textwidth}{!}{%
\begin{tabular}{lllccccccccccc||c}
\toprule
Method & LLM & Layer & am & cs & da & de & he & id & jv & mt & sk & sl & su & Average \\
\midrule
\midrule
\multirow[c]{12}{*}{\textit{Language}} & \multirow[c]{3}{*}{Gemma 27b} & L10 & {\cellcolor{cbBlue!17}} 0.79 & {\cellcolor{cbBlue!25}} 1.67 & {\cellcolor{cbBlue!31}} 2.27 & {\cellcolor{cbOrange!15}} -0.60 & {\cellcolor{cbBlue!11}} 0.11 & {\cellcolor{cbBlue!31}} 2.29 & {\cellcolor{cbOrange!13}} -0.39 & {\cellcolor{cbBlue!36}} 2.85* & {\cellcolor{cbBlue!26}} 1.73 & {\cellcolor{cbBlue!45}} 3.75* & {\cellcolor{cbOrange!11}} -0.12 & {\cellcolor{cbBlue!22}} 1.30 \\
 &  & L22 & {\cellcolor{cbOrange!28}} -1.95 & {\cellcolor{cbBlue!21}} 1.27 & {\cellcolor{cbBlue!21}} 1.23 & {\cellcolor{cbBlue!11}} 0.15 & {\cellcolor{cbBlue!38}} 3.00* & {\cellcolor{cbBlue!20}} 1.09 & {\cellcolor{cbOrange!14}} -0.53 & {\cellcolor{cbBlue!39}} 3.11* & {\cellcolor{cbBlue!11}} 0.20 & {\cellcolor{cbBlue!33}} 2.49* & {\cellcolor{cbBlue!12}} 0.26 & {\cellcolor{cbBlue!18}} 0.94 \\
 &  & L34 & {\cellcolor{cbBlue!18}} 0.94 & {\cellcolor{cbBlue!24}} 1.54 & {\cellcolor{cbBlue!33}} 2.49 & {\cellcolor{cbBlue!10}} 0.03 & {\cellcolor{cbBlue!20}} 1.15 & {\cellcolor{cbBlue!19}} 1.06 & {\cellcolor{cbOrange!11}} -0.20 & {\cellcolor{cbBlue!37}} 2.89 & {\cellcolor{cbBlue!13}} 0.40 & {\cellcolor{cbBlue!41}} 3.40* & {\cellcolor{cbOrange!12}} -0.25 & {\cellcolor{cbBlue!21}} 1.22 \\
  \cmidrule(lr){2-15}
 & \multirow[c]{3}{*}{Gemma 9b} & L9 & {\cellcolor{cbOrange!42}} -3.51 & {\cellcolor{cbOrange!16}} -0.67 & {\cellcolor{cbBlue!37}} 2.92* & {\cellcolor{cbBlue!41}} 3.36* & {\cellcolor{cbBlue!10}} 0.05 & {\cellcolor{cbBlue!10}} 0.09 & {\cellcolor{cbBlue!36}} 2.82 & {\cellcolor{cbBlue!17}} 0.82 & {\cellcolor{cbOrange!18}} -0.88 & {\cellcolor{cbBlue!20}} 1.16 & {\cellcolor{cbOrange!27}} -1.92 & {\cellcolor{cbBlue!13}} 0.39 \\
\bottomrule
\end{tabular}
}
\end{table*}

\begin{table*}
\caption{Mean F1 difference across languages, models, steering methods, and layers for \textbf{sentiment} detection, with \textcolor{cbBlue}{positive} and \textcolor{cbOrange}{negative} differences compared to a \textbf{few-shot} prompt with no steering. Cells with * denote statistically significant increases over baseline.}
\label{tab:steering_results_icl_sent}
\tiny
\resizebox{\textwidth}{!}{%
\begin{tabular}{lllccccccccccc||c}
\toprule
Method & LLM & Layer & am & cs & da & de & he & id & jv & mt & sk & sl & su & Average \\
\midrule
\midrule
\multirow[c]{12}{*}{\textit{Language}} & \multirow[c]{3}{*}{Gemma 27b} & L10 & {\cellcolor{cbBlue!19}} 2.36 & {\cellcolor{cbBlue!23}} 3.30* & {\cellcolor{cbBlue!17}} 1.81 & {\cellcolor{cbBlue!28}} 4.42* & {\cellcolor{cbBlue!39}} 7.30* & {\cellcolor{cbBlue!29}} 4.66* & {\cellcolor{cbBlue!14}} 1.22 & {\cellcolor{cbBlue!25}} 3.89* & {\cellcolor{cbOrange!13}} -0.75 & {\cellcolor{cbBlue!36}} 6.43* & {\cellcolor{cbOrange!11}} -0.47 & {\cellcolor{cbBlue!22}} 3.11 \\
 &  & L22 & {\cellcolor{cbBlue!12}} 0.68 & {\cellcolor{cbBlue!41}} 7.76* & {\cellcolor{cbOrange!12}} -0.49 & {\cellcolor{cbBlue!24}} 3.57* & {\cellcolor{cbBlue!13}} 0.75 & {\cellcolor{cbBlue!19}} 2.24 & {\cellcolor{cbBlue!11}} 0.49 & {\cellcolor{cbBlue!19}} 2.26 & {\cellcolor{cbOrange!12}} -0.60 & {\cellcolor{cbBlue!46}} 9.00* & {\cellcolor{cbBlue!13}} 0.85 & {\cellcolor{cbBlue!19}} 2.41 \\
 &  & L34 & {\cellcolor{cbBlue!14}} 1.00 & {\cellcolor{cbBlue!40}} 7.40* & {\cellcolor{cbBlue!13}} 0.94 & {\cellcolor{cbBlue!24}} 3.42 & {\cellcolor{cbBlue!11}} 0.45 & {\cellcolor{cbBlue!24}} 3.52* & {\cellcolor{cbBlue!14}} 1.13 & {\cellcolor{cbBlue!14}} 1.21 & {\cellcolor{cbBlue!23}} 3.30 & {\cellcolor{cbBlue!28}} 4.48* & {\cellcolor{cbBlue!12}} 0.57 & {\cellcolor{cbBlue!20}} 2.49 \\
 \cmidrule(lr){2-15}
 & \multirow[c]{3}{*}{Gemma 9b} & L9 & {\cellcolor{cbOrange!23}} -3.33 & {\cellcolor{cbBlue!48}} 9.28* & {\cellcolor{cbBlue!11}} 0.49 & {\cellcolor{cbBlue!17}} 1.89 & {\cellcolor{cbBlue!19}} 2.39 & {\cellcolor{cbOrange!22}} -3.06 & {\cellcolor{cbBlue!21}} 2.85 & {\cellcolor{cbBlue!22}} 3.03* & {\cellcolor{cbBlue!57}} 11.69* & {\cellcolor{cbOrange!24}} -3.48 & {\cellcolor{cbBlue!13}} 0.95 & {\cellcolor{cbBlue!18}} 2.06 \\
\bottomrule
\end{tabular}
}
\end{table*}

\begin{table*}
\caption{Selected best $\alpha$ values based on validation sets for each steering method, model, layer, language, and task for the \textbf{zero-shot} setting.}
\label{tab:selected_alpha_zero_shot}
\resizebox{\textwidth}{!}{%
\begin{tabular}{lllccccccccccc||ccccccccccc}
\toprule
 &  & Task & \multicolumn{11}{c}{sent} & \multicolumn{11}{c}{topic} \\
Method & LLM & Language & am & cs & da & de & he & id & jv & mt & sk & sl & su & am & cs & da & de & he & id & jv & mt & sk & sl & su \\
\midrule
\multirow[c]{12}{*}{\textit{Language}} & \multirow[c]{3}{*}{Gemma 27b} & L10 & 75 & 75 & 25 & 75 & 25 & 25 & 100 & 25 & 50 & 100 & 75 & 50 & 75 & 25 & 100 & 50 & 100 & 50 & 25 & 25 & 75 & 75 \\
 &  & L22 & 75 & 100 & 25 & 75 & 50 & 50 & 100 & 50 & 25 & 50 & 75 & 100 & 100 & 75 & 25 & 75 & 75 & 100 & 75 & 25 & 25 & 50 \\
 &  & L34 & 75 & 25 & 100 & 25 & 75 & 25 & 75 & 25 & 50 & 50 & 75 & 75 & 75 & 100 & 50 & 100 & 75 & 100 & 25 & 25 & 100 & 75 \\
  \cmidrule(lr){2-25}
 & \multirow[c]{3}{*}{Gemma 9b} & L9 & 25 & 75 & 50 & 100 & 50 & 75 & 25 & 75 & 75 & 100 & 100 & 50 & 100 & 25 & 75 & 75 & 25 & 25 & 100 & 100 & 100 & 100 \\
 &  & L20 & 50 & 100 & 100 & 100 & 75 & 100 & 25 & 50 & 25 & 100 & 75 & 25 & 75 & 50 & 25 & 100 & 25 & 75 & 25 & 75 & 25 & 50 \\
 &  & L31 & 75 & 100 & 100 & 75 & 100 & 75 & 25 & 25 & 75 & 25 & 25 & 25 & 100 & 75 & 50 & 100 & 75 & 100 & 25 & 100 & 100 & 25 \\
\bottomrule
\end{tabular}
}
\end{table*}

\begin{table*}
\caption{Selected best $\alpha$ values based on validation sets for each steering method, model, layer, language, and task for the \textbf{few-shot} setting.}
\label{tab:selected_alpha_few_shot}
\resizebox{\textwidth}{!}{%
\begin{tabular}{lllccccccccccc||ccccccccccc}
\toprule
 &  & Task & \multicolumn{11}{c}{sent} & \multicolumn{11}{c}{topic} \\
Method & LLM & Language & am & cs & da & de & he & id & jv & mt & sk & sl & su & am & cs & da & de & he & id & jv & mt & sk & sl & su \\
\midrule
\multirow[c]{12}{*}{\textit{Language}} & \multirow[c]{3}{*}{Gemma 27b} & L10 & 25 & 25 & 75 & 25 & 25 & 75 & 100 & 50 & 25 & 75 & 50 & 25 & 25 & 75 & 50 & 50 & 100 & 50 & 50 & 100 & 75 & 100 \\
 &  & L22 & 50 & 25 & 50 & 25 & 100 & 75 & 25 & 100 & 50 & 50 & 25 & 50 & 25 & 75 & 100 & 50 & 25 & 75 & 75 & 50 & 100 & 100 \\
 &  & L34 & 25 & 75 & 25 & 50 & 100 & 25 & 100 & 100 & 75 & 100 & 25 & 25 & 100 & 75 & 100 & 25 & 25 & 50 & 75 & 25 & 100 & 75 \\
  \cmidrule(lr){2-25}
 & \multirow[c]{3}{*}{Gemma 9b} & L9 & 50 & 100 & 25 & 100 & 50 & 25 & 50 & 75 & 100 & 25 & 25 & 50 & 75 & 50 & 25 & 75 & 50 & 50 & 50 & 50 & 25 & 25 \\
 &  & L20 & 25 & 100 & 75 & 50 & 100 & 100 & 25 & 100 & 50 & 75 & 25 & 50 & 100 & 75 & 25 & 100 & 50 & 100 & 25 & 50 & 50 & 75 \\
 &  & L31 & 25 & 25 & 50 & 25 & 25 & 50 & 25 & 25 & 50 & 100 & 75 & 50 & 75 & 75 & 100 & 75 & 75 & 100 & 75 & 25 & 100 & 50 \\
\bottomrule
\end{tabular}
}
\end{table*}

\begin{table*}
\caption{Mean F1 difference across languages, models, steering methods, and layers aggregated over tasks, with \textcolor{cbBlue}{positive} and \textcolor{cbOrange}{negative} differences compared to a \textbf{zero-shot prompt} with no steering. Best alpha selected per layer as the sum of differences.}
\label{tab:steering_results_best_alpha_per_layer}
\tiny
\resizebox{\textwidth}{!}{%
\begin{tabular}{lllccccccccccc||c}
\toprule
Method & LLM & Layer & am & cs & da & de & he & id & jv & mt & sk & sl & su & Average \\
\midrule
\midrule
\multirow{12}{*}{\textit{Language}} & \multirow{3}{*}{Gemma 2 27b} & L10 {\scriptsize$(\alpha=100.0)$} & {\cellcolor{cbBlue!11}} 0.17 & {\cellcolor{cbBlue!11}} 0.28 & {\cellcolor{cbOrange!22}} -2.10 & {\cellcolor{cbBlue!23}} 2.24 & {\cellcolor{cbOrange!16}} -1.16 & {\cellcolor{cbBlue!24}} 2.49 & {\cellcolor{cbBlue!21}} 1.84 & {\cellcolor{cbOrange!13}} -0.56 & {\cellcolor{cbBlue!19}} 1.52 & {\cellcolor{cbBlue!25}} 2.66 & {\cellcolor{cbBlue!23}} 2.33 & {\cellcolor{cbBlue!15}} 0.88 \\
&  & L22 {\scriptsize$(\alpha=75.0)$} & {\cellcolor{cbBlue!24}} 2.36 & {\cellcolor{cbOrange!11}} -0.22 & {\cellcolor{cbOrange!10}} -0.13 & {\cellcolor{cbBlue!14}} 0.73 & {\cellcolor{cbOrange!13}} -0.52 & {\cellcolor{cbBlue!14}} 0.77 & {\cellcolor{cbOrange!25}} -2.67 & {\cellcolor{cbBlue!12}} 0.38 & {\cellcolor{cbBlue!20}} 1.83 & {\cellcolor{cbBlue!10}} 0.07 & {\cellcolor{cbBlue!39}} 4.92 & {\cellcolor{cbBlue!14}} 0.68 \\
&  & L34 {\scriptsize$(\alpha=75.0)$} & {\cellcolor{cbBlue!35}} 4.34 & {\cellcolor{cbOrange!13}} -0.51 & {\cellcolor{cbOrange!12}} -0.43 & {\cellcolor{cbOrange!18}} -1.37 & {\cellcolor{cbOrange!12}} -0.36 & {\cellcolor{cbBlue!23}} 2.17 & {\cellcolor{cbBlue!21}} 1.96 & {\cellcolor{cbOrange!20}} -1.71 & {\cellcolor{cbOrange!20}} -1.72 & {\cellcolor{cbOrange!10}} -0.07 & {\cellcolor{cbBlue!47}} 6.20 & {\cellcolor{cbBlue!14}} 0.77 \\
\cmidrule(lr){2-15}
& \multirow{3}{*}{Gemma 2 9b} & L9 {\scriptsize$(\alpha=50.0)$} & {\cellcolor{cbOrange!29}} -3.25 & {\cellcolor{cbBlue!18}} 1.48 & {\cellcolor{cbBlue!16}} 1.09 & {\cellcolor{cbBlue!12}} 0.43 & {\cellcolor{cbBlue!23}} 2.29 & {\cellcolor{cbBlue!11}} 0.17 & {\cellcolor{cbOrange!31}} -3.63 & {\cellcolor{cbBlue!12}} 0.45 & {\cellcolor{cbBlue!34}} 4.08 & {\cellcolor{cbBlue!51}} 6.88 & {\cellcolor{cbBlue!10}} 0.16 & {\cellcolor{cbBlue!15}} 0.92 \\
\bottomrule
\end{tabular}
}
\end{table*}

\begin{table*}
\caption{Mean F1 difference across languages, models, steering methods, and layers aggregated over tasks, with \textcolor{cbBlue}{positive} and \textcolor{cbOrange}{negative} differences compared to a \textbf{few-shot prompt} with no steering. Best alpha selected per layer as the sum of differences.}
\label{tab:steering_results_icl_alpha_per_layer}
\tiny
\resizebox{\textwidth}{!}{%
\begin{tabular}{lllccccccccccc||c}
\toprule
Method & LLM & Layer & am & cs & da & de & he & id & jv & mt & sk & sl & su & Average \\
\midrule
\midrule
\multirow{12}{*}{\textit{Language}} & \multirow{3}{*}{Gemma 2 27b} & L10 {\scriptsize$(\alpha=25.0)$} & {\cellcolor{cbBlue!27}} 1.58 & {\cellcolor{cbBlue!38}} 2.48 & {\cellcolor{cbBlue!17}} 0.63 & {\cellcolor{cbBlue!29}} 1.73 & {\cellcolor{cbBlue!47}} 3.25 & {\cellcolor{cbBlue!25}} 1.36 & {\cellcolor{cbOrange!20}} -0.96 & {\cellcolor{cbBlue!35}} 2.27 & {\cellcolor{cbOrange!14}} -0.39 & {\cellcolor{cbBlue!50}} 3.55 & {\cellcolor{cbOrange!30}} -1.76 & {\cellcolor{cbBlue!24}} 1.25 \\
& & L22 {\scriptsize$(\alpha=50.0)$} & {\cellcolor{cbOrange!17}} -0.64 & {\cellcolor{cbBlue!38}} 2.48 & {\cellcolor{cbBlue!10}} 0.00 & {\cellcolor{cbOrange!16}} -0.59 & {\cellcolor{cbBlue!31}} 1.85 & {\cellcolor{cbBlue!21}} 1.04 & {\cellcolor{cbOrange!23}} -1.19 & {\cellcolor{cbBlue!10}} 0.00 & {\cellcolor{cbOrange!12}} -0.20 & {\cellcolor{cbBlue!72}} 5.51 & {\cellcolor{cbOrange!25}} -1.35 & {\cellcolor{cbBlue!17}} 0.63 \\
& & L34 {\scriptsize$(\alpha=25.0)$} & {\cellcolor{cbBlue!21}} 0.97 & {\cellcolor{cbBlue!39}} 2.58 & {\cellcolor{cbBlue!13}} 0.34 & {\cellcolor{cbBlue!13}} 0.35 & {\cellcolor{cbOrange!12}} -0.26 & {\cellcolor{cbBlue!36}} 2.29 & {\cellcolor{cbOrange!17}} -0.69 & {\cellcolor{cbBlue!28}} 1.66 & {\cellcolor{cbOrange!18}} -0.75 & {\cellcolor{cbBlue!36}} 2.32 & {\cellcolor{cbOrange!19}} -0.87 & {\cellcolor{cbBlue!18}} 0.72 \\
\cmidrule(lr){2-15}
& \multirow{3}{*}{Gemma 2 9b} & L9 {\scriptsize$(\alpha=25.0)$} & {\cellcolor{cbOrange!57}} -4.14 & {\cellcolor{cbBlue!13}} 0.34 & {\cellcolor{cbBlue!12}} 0.24 & {\cellcolor{cbBlue!26}} 1.48 & {\cellcolor{cbOrange!16}} -0.54 & {\cellcolor{cbOrange!27}} -1.54 & {\cellcolor{cbBlue!15}} 0.51 & {\cellcolor{cbOrange!32}} -1.97 & {\cellcolor{cbBlue!33}} 2.06 & {\cellcolor{cbOrange!23}} -1.16 & {\cellcolor{cbOrange!15}} -0.49 & {\cellcolor{cbOrange!15}} -0.47 \\
\bottomrule
\end{tabular}
}
\end{table*}

\begin{table*}[t]
\centering
\tiny
\caption{Boxplot visualizations of relative increases/decreases in \% for F1 metrics for LLM + steering methods for alpha values in \textbf{zero-shot} setting.  The X axis in the figures represents the different alpha values for that LLM + steering combination aggregated over languages and tasks, and the Y axis represents the relative difference increase for F1 downstream model performance.}
\label{tab:results_zero_shot_alphas}
\renewcommand{\arraystretch}{1.2}

\begin{tabular}{>{\centering\arraybackslash}m{0.1\textwidth}
                >{\centering\arraybackslash}m{0.2\textwidth}
                >{\centering\arraybackslash}m{0.2\textwidth}
                >{\centering\arraybackslash}m{0.2\textwidth}}
\toprule
\textbf{LLM + steering} &
\textbf{Early layer} &
\textbf{Middle layer} &
\textbf{Late layer}  \\
\midrule

Gemma 2 27b  (lang.) &
\includegraphics[width=0.2\textwidth]{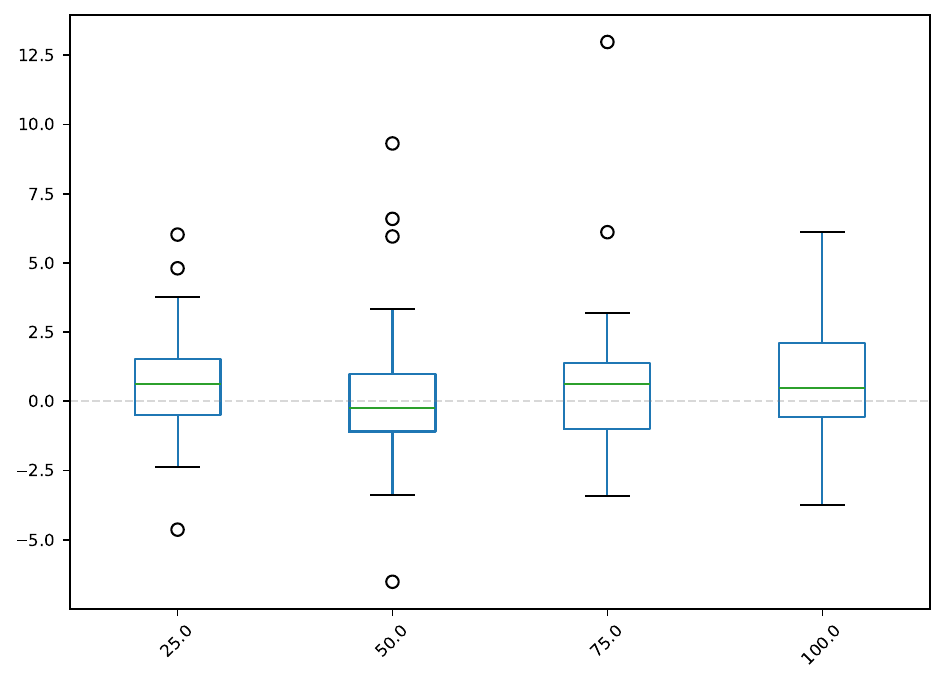} &
\includegraphics[width=0.2\textwidth]{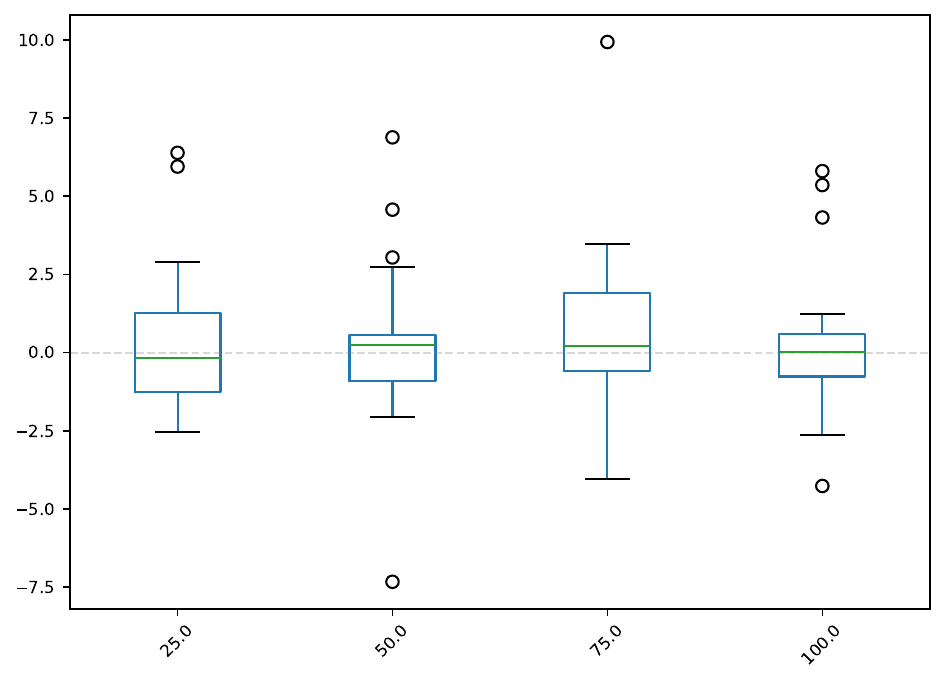} &
\includegraphics[width=0.2\textwidth]{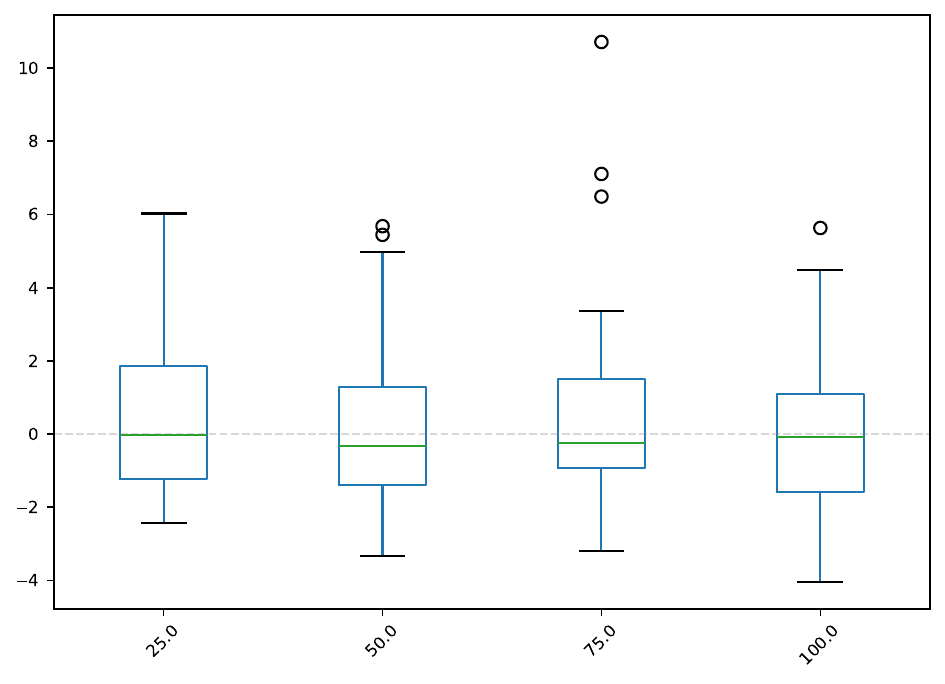} \\

Gemma 2 27b  (qual.) &
\includegraphics[width=0.2\textwidth]{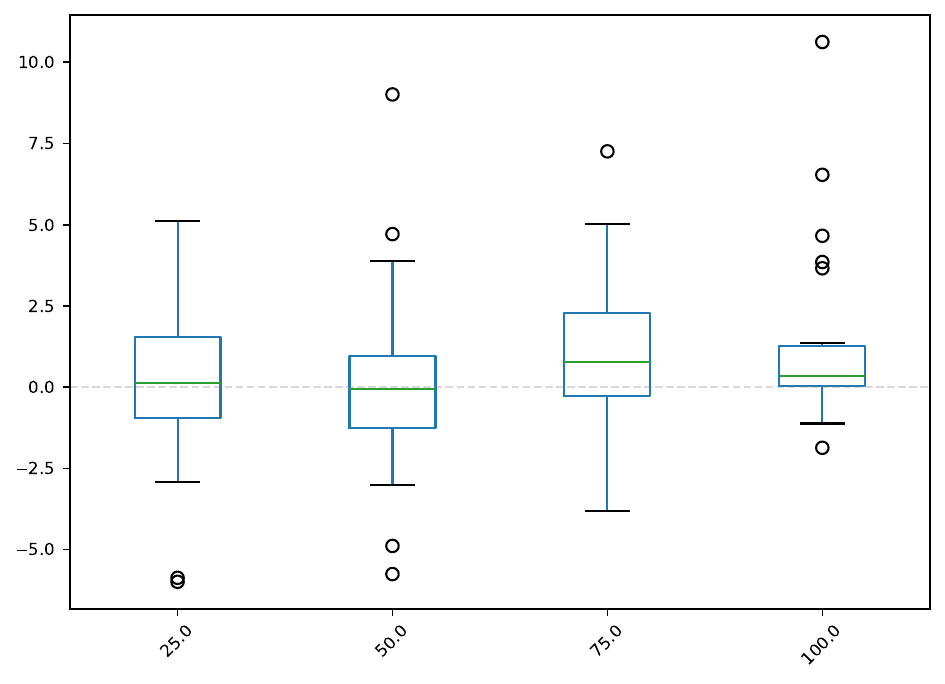} &
\includegraphics[width=0.2\textwidth]{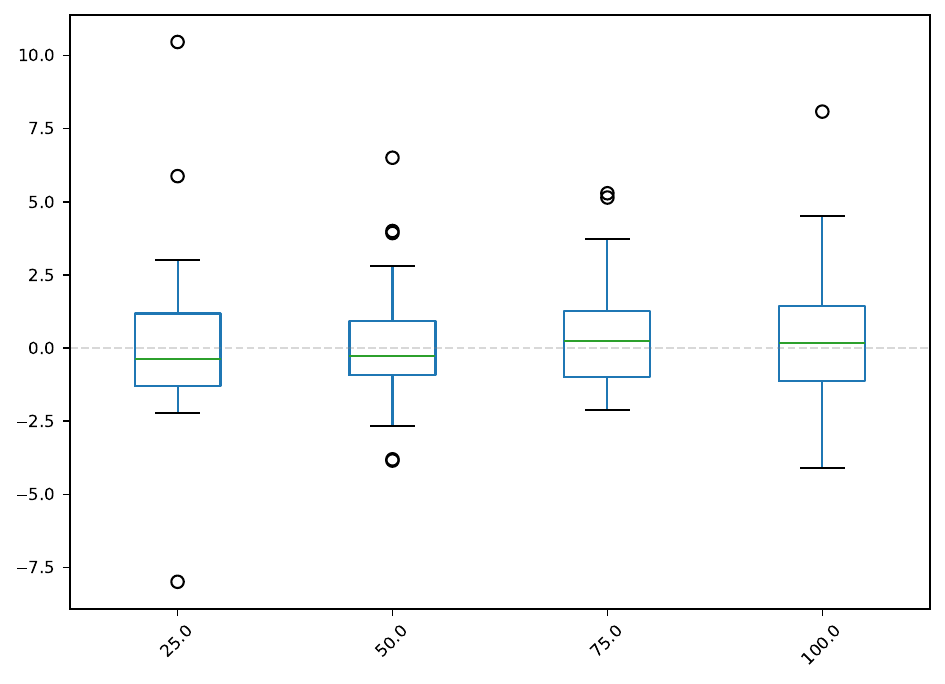} &
\includegraphics[width=0.2\textwidth]{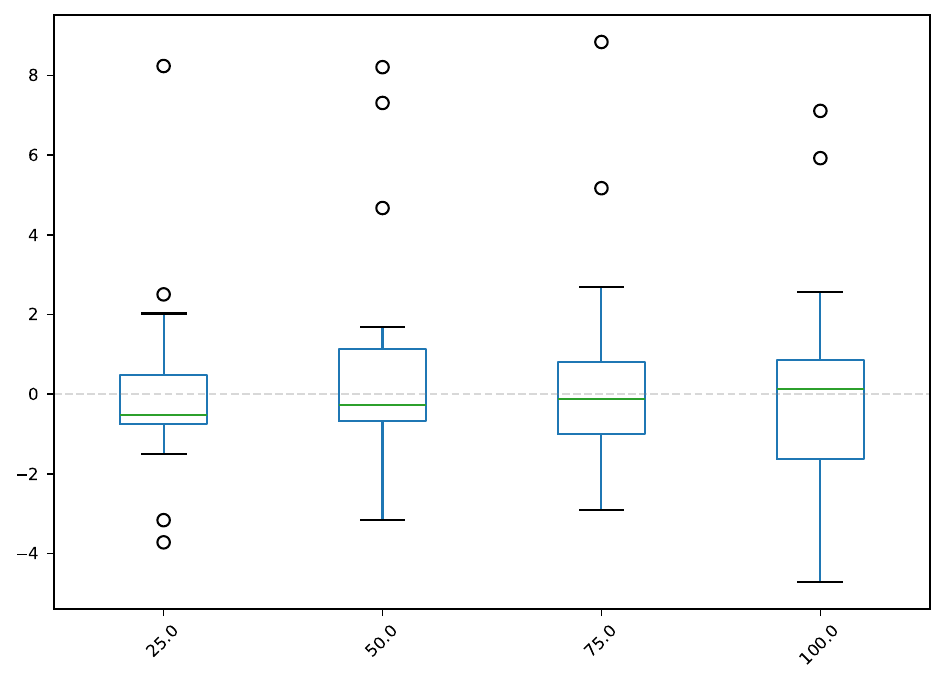} \\

Gemma 2 9b  (lang.) &
\includegraphics[width=0.2\textwidth]{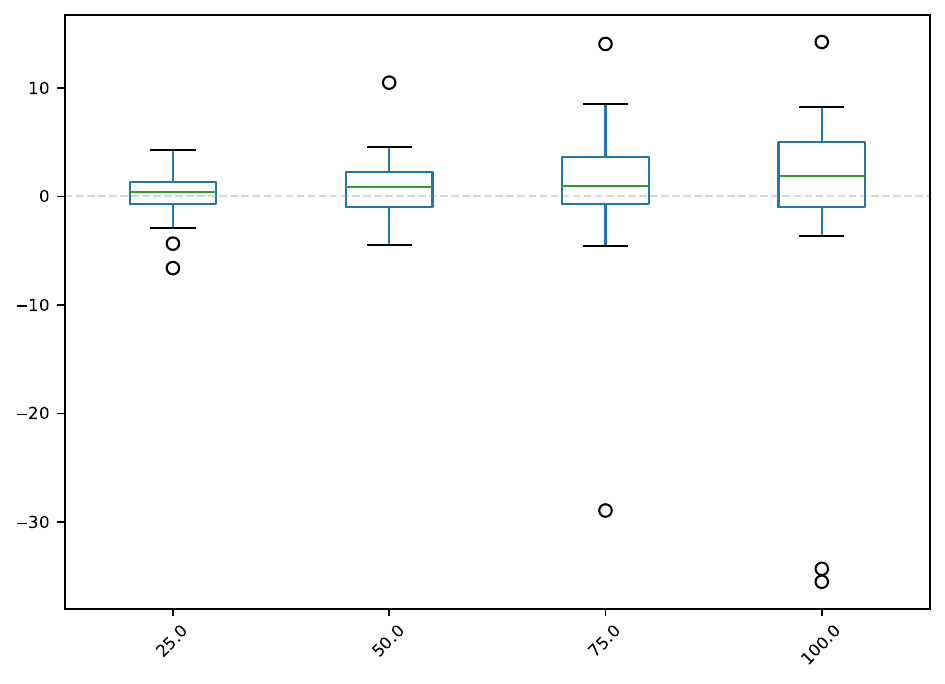} &
\includegraphics[width=0.2\textwidth]{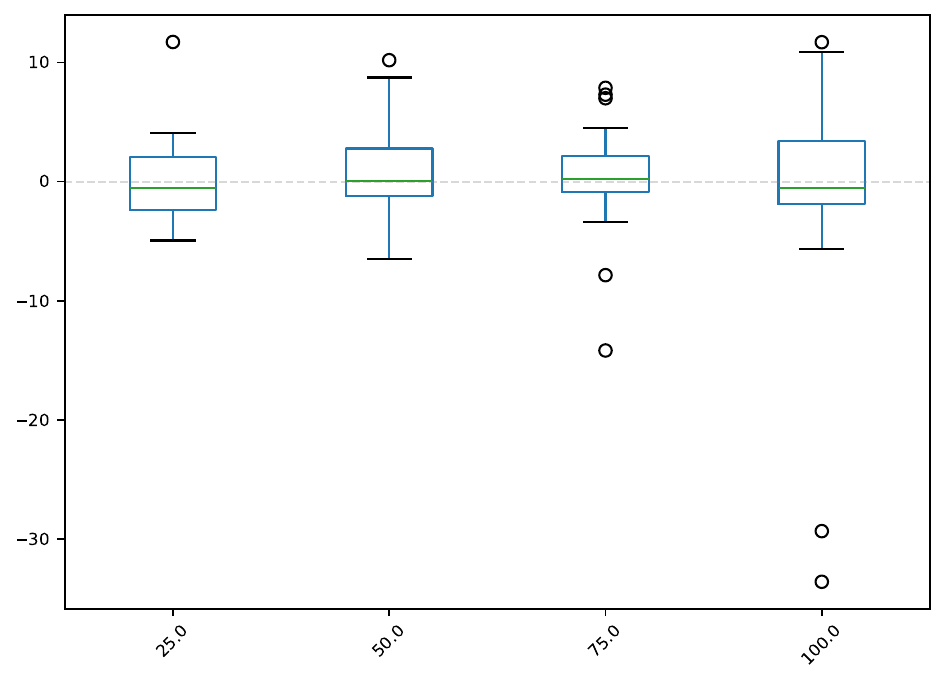} &
\includegraphics[width=0.2\textwidth]{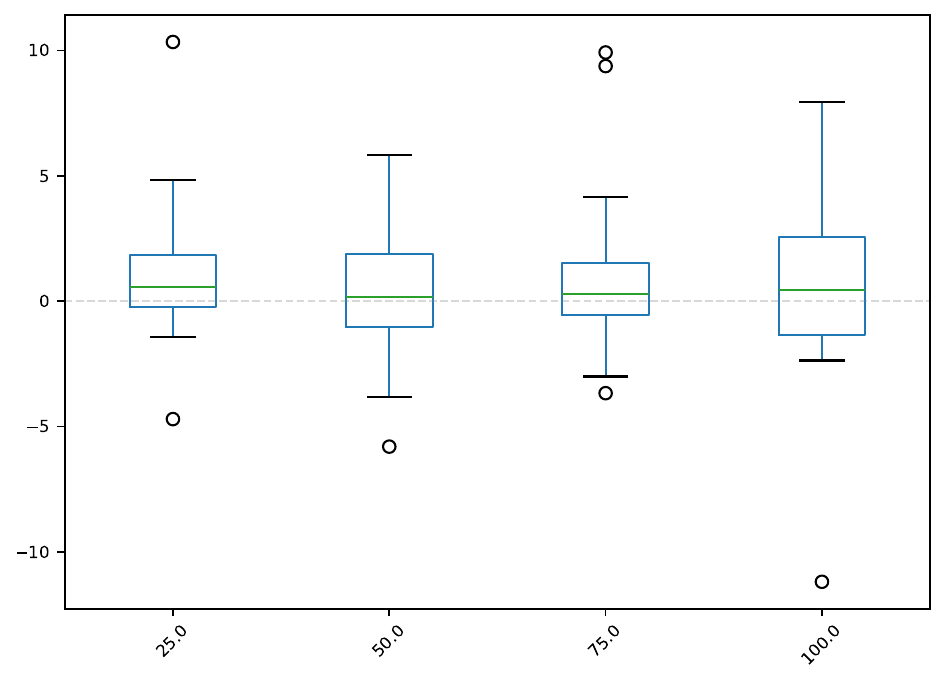} \\

Gemma 2 9b  (qual.) &
\includegraphics[width=0.2\textwidth]{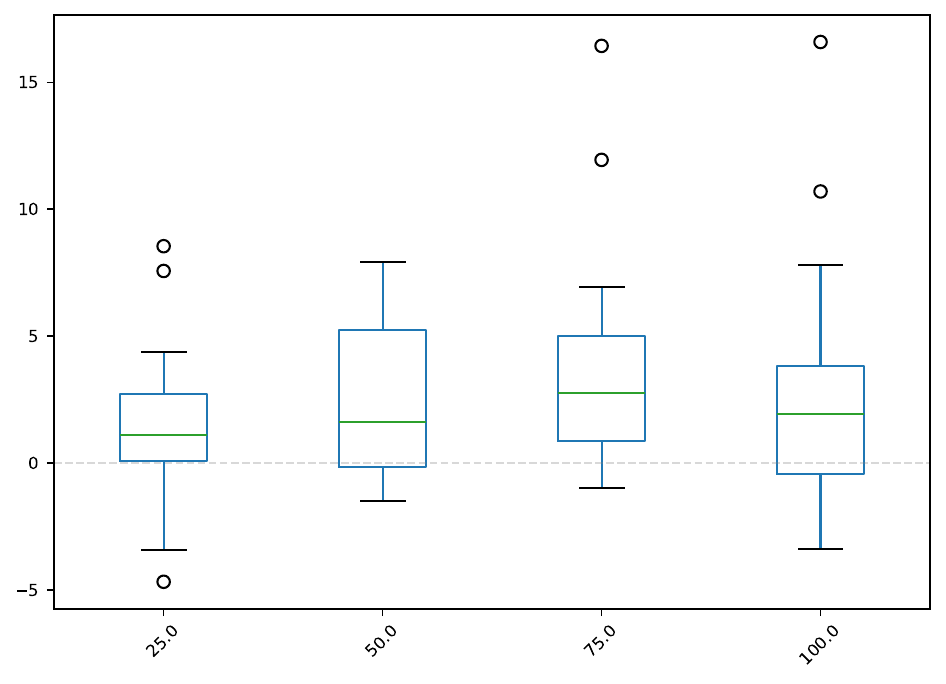} &
\includegraphics[width=0.2\textwidth]{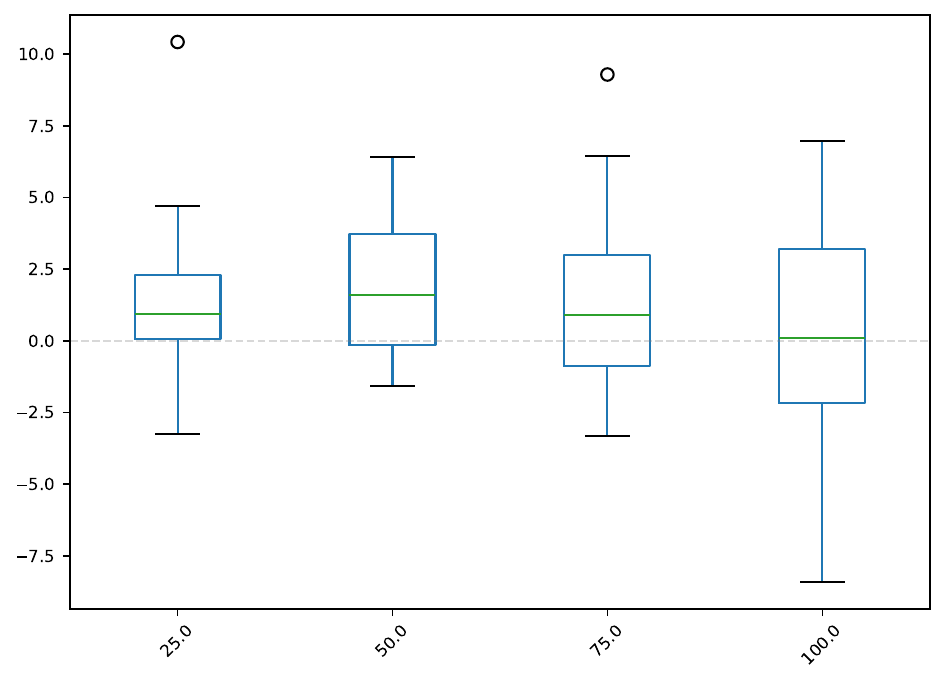} &
\includegraphics[width=0.2\textwidth]{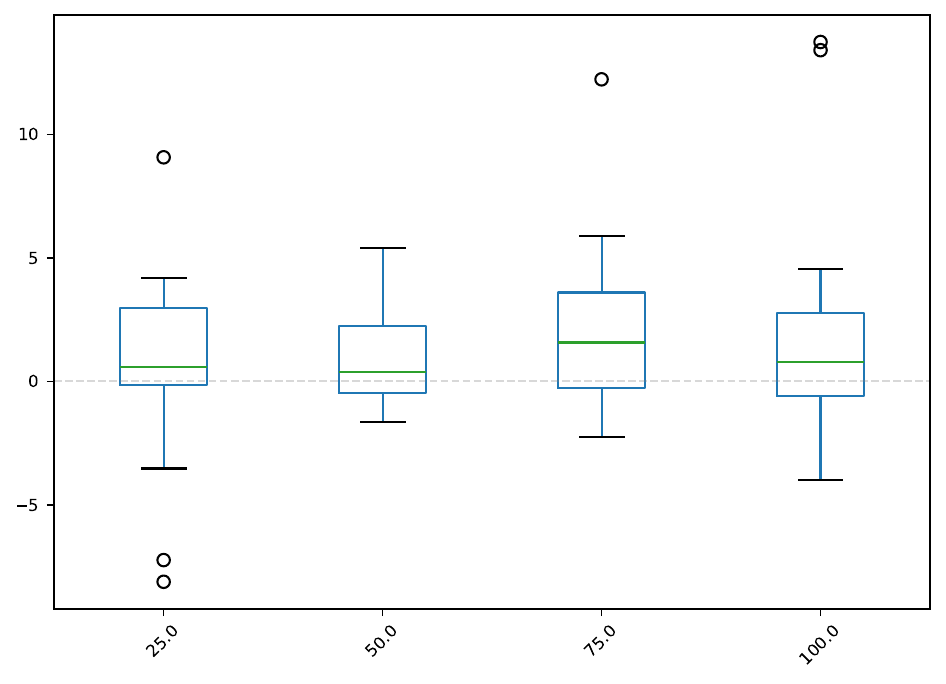} \\

Llama 3.1 70b  (lang.) &
\includegraphics[width=0.2\textwidth]{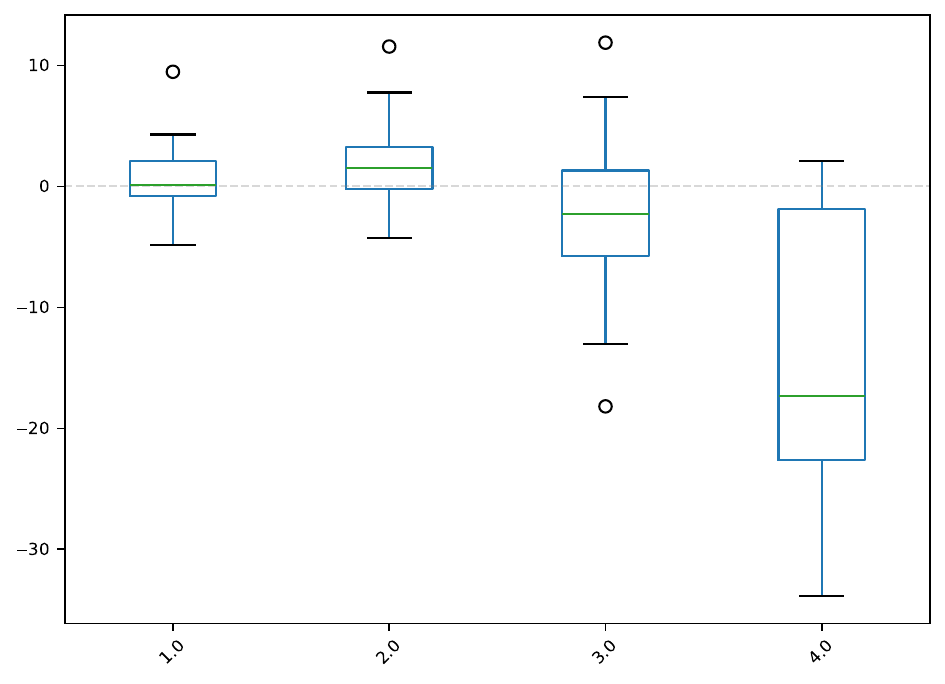} &
\includegraphics[width=0.2\textwidth]{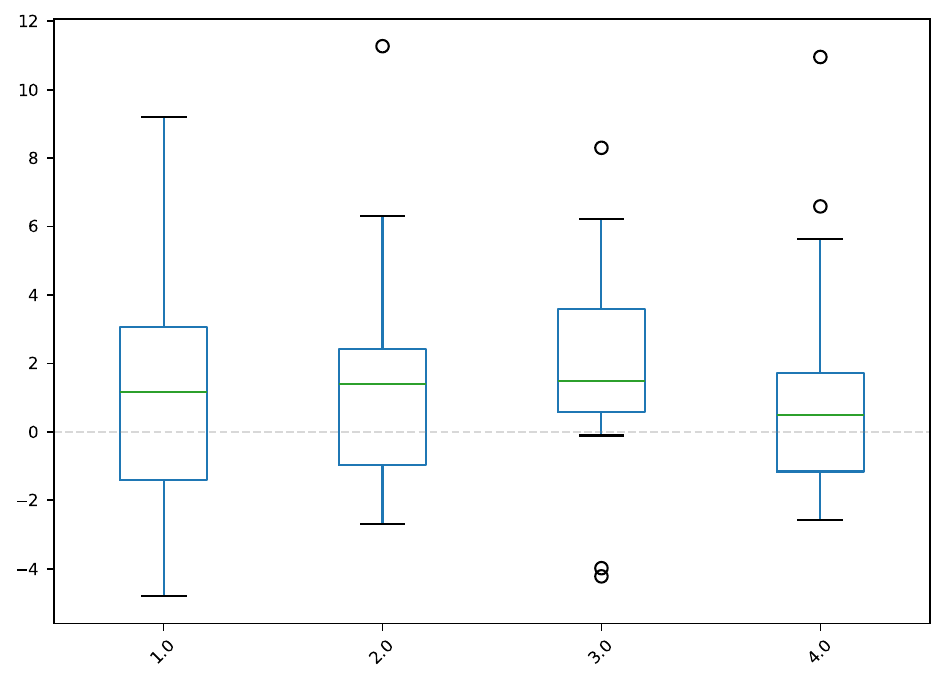} &
\includegraphics[width=0.2\textwidth]{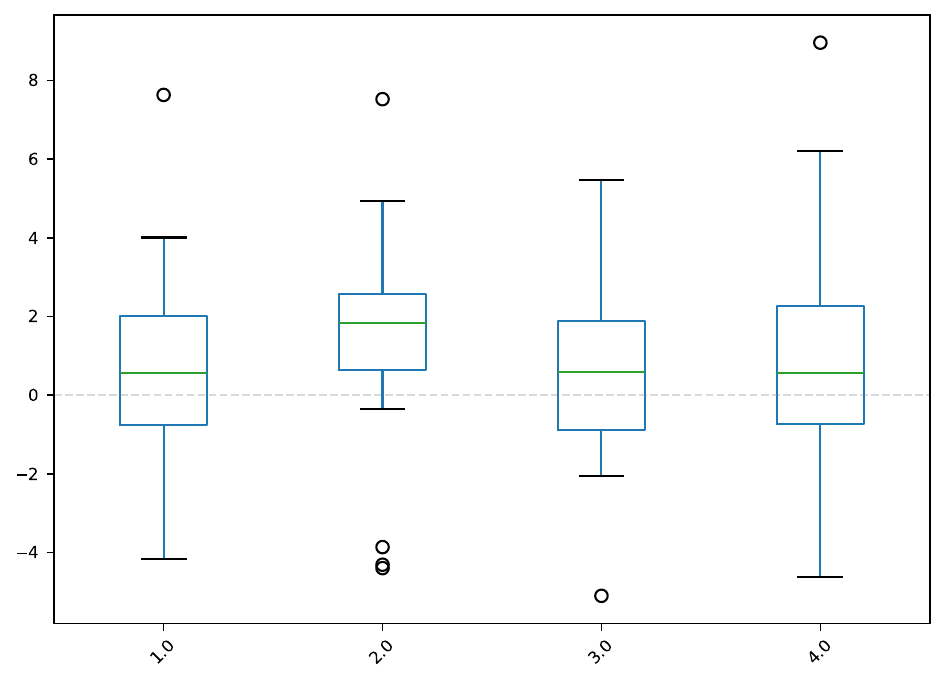} \\
\bottomrule
\end{tabular}
\end{table*}

\begin{table*}[t]
\centering
\tiny
\caption{Boxplot visualizations of relative increases/decreases in \% for F1 metrics for LLM + steering methods for alpha values in \textbf{few-shot} setting.  The X axis in the figures represents the different alpha values for that LLM + steering combination aggregated over languages and tasks, and the Y axis represents the relative difference increase for F1 downstream model performance.}
\label{tab:results_few_shot_alphas}
\renewcommand{\arraystretch}{1.2}

\begin{tabular}{>{\centering\arraybackslash}m{0.1\textwidth}
                >{\centering\arraybackslash}m{0.2\textwidth}
                >{\centering\arraybackslash}m{0.2\textwidth}
                >{\centering\arraybackslash}m{0.2\textwidth}}
\toprule
\textbf{LLM + steering} &
\textbf{Early layer} &
\textbf{Middle layer} &
\textbf{Late layer}  \\
\midrule

Gemma 2 27b  (lang.) &
\includegraphics[width=0.2\textwidth]{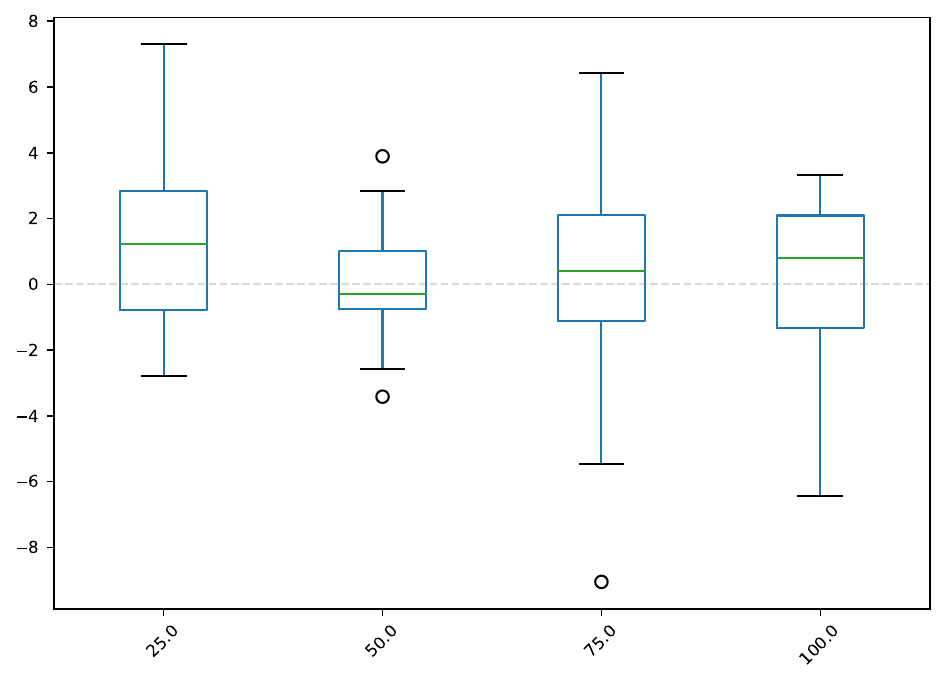} &
\includegraphics[width=0.2\textwidth]{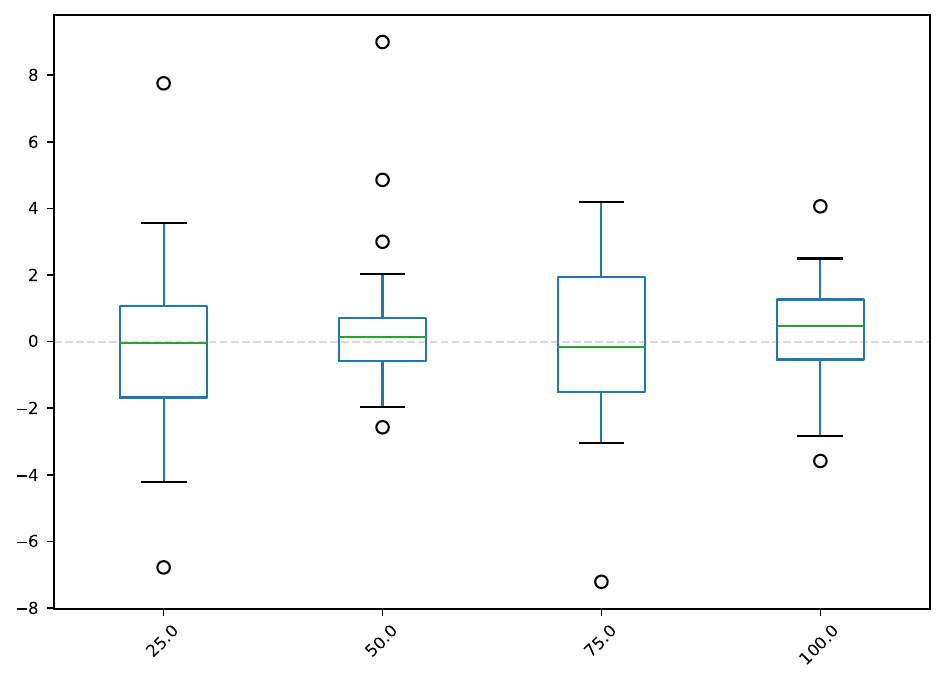} &
\includegraphics[width=0.2\textwidth]{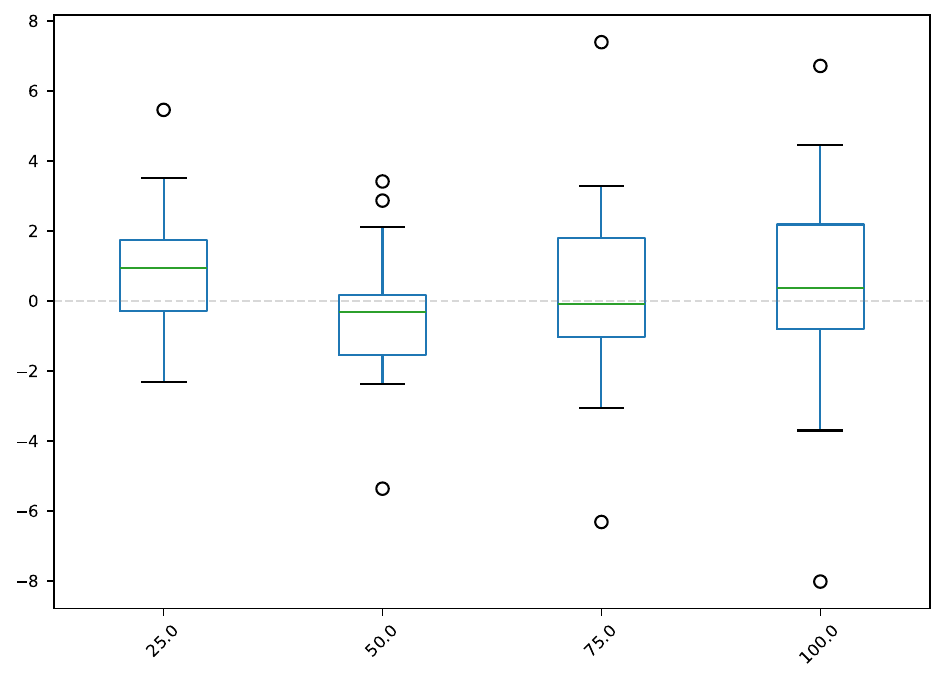} \\

Gemma 2 27b  (qual.) &
\includegraphics[width=0.2\textwidth]{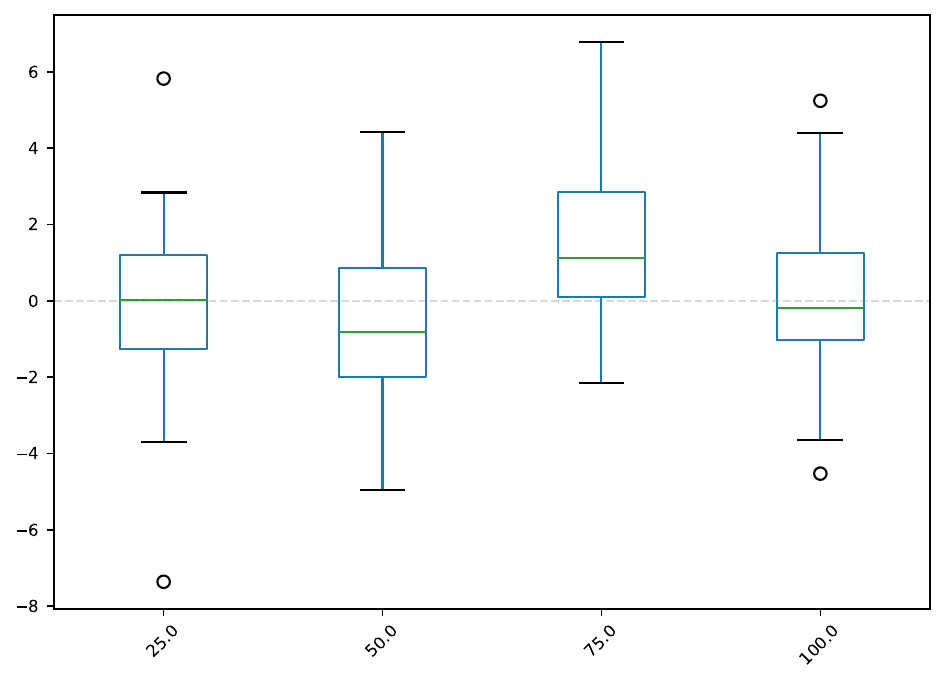} &
\includegraphics[width=0.2\textwidth]{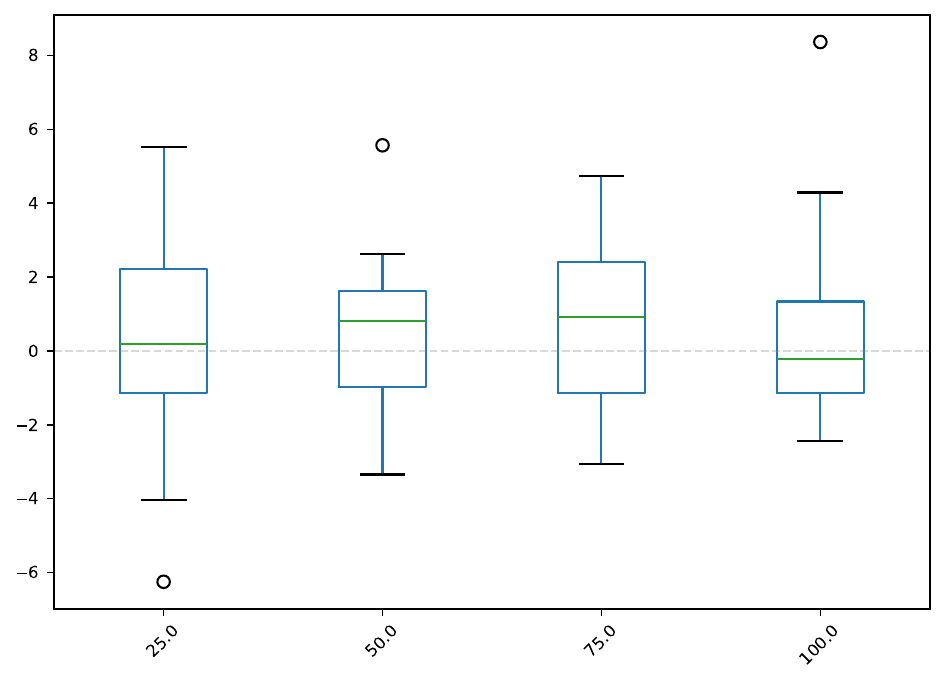} &
\includegraphics[width=0.2\textwidth]{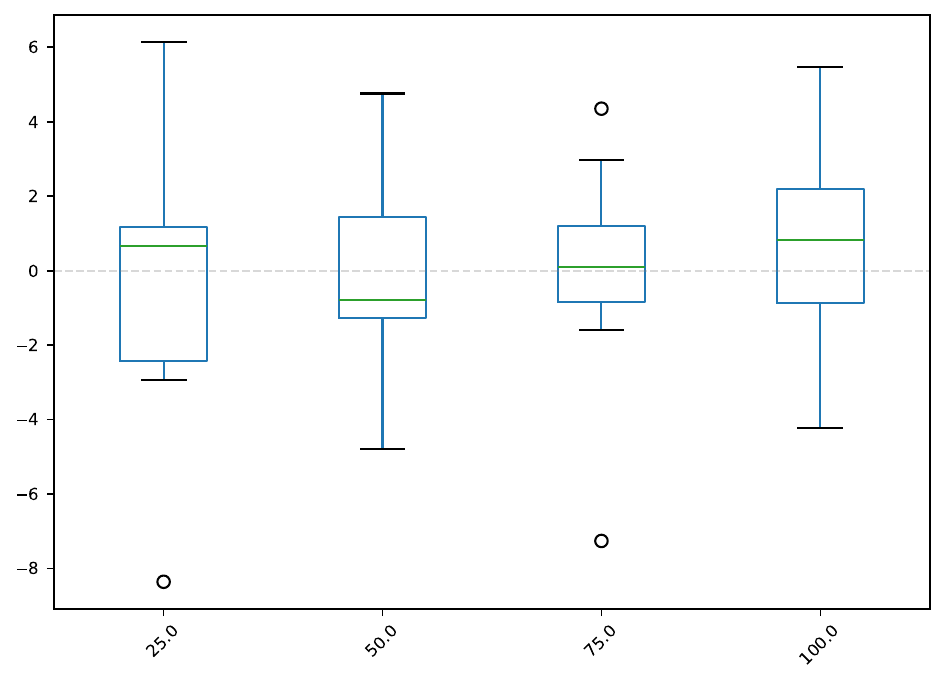} \\

Gemma 2 9b  (lang.) &
\includegraphics[width=0.2\textwidth]{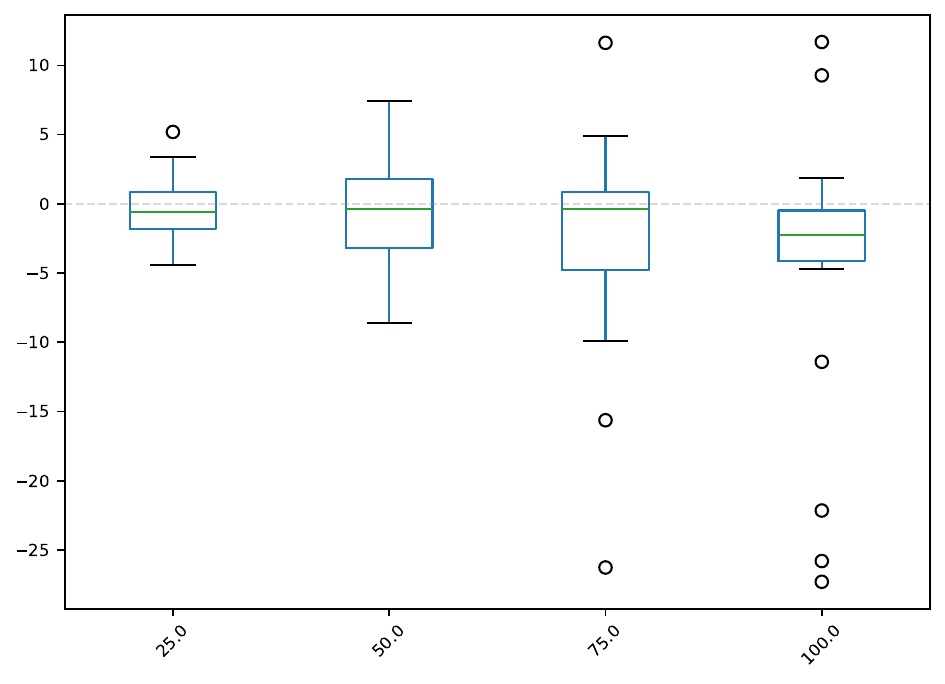} &
\includegraphics[width=0.2\textwidth]{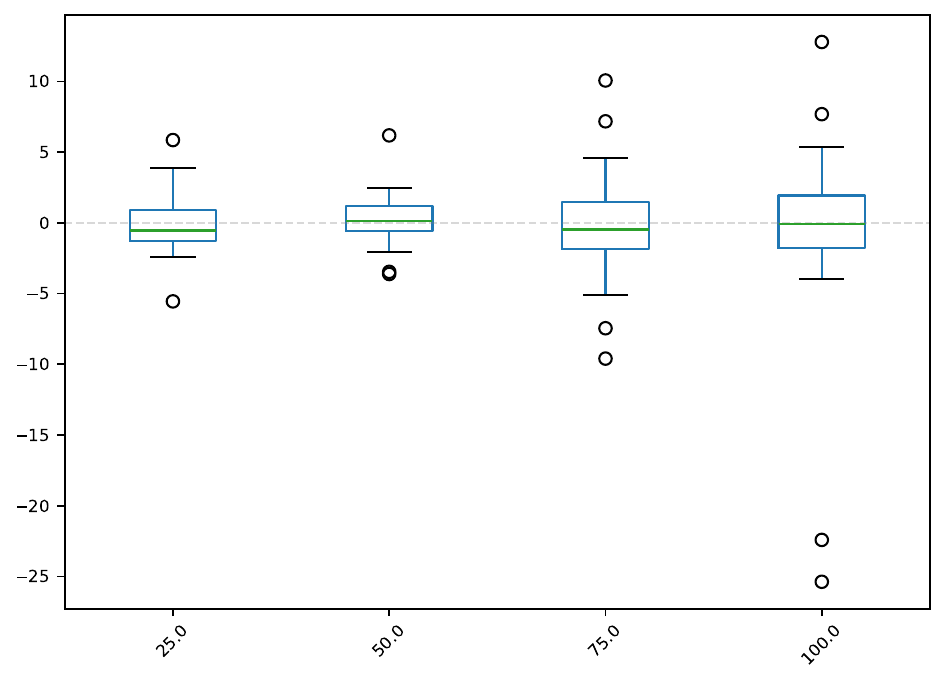} &
\includegraphics[width=0.2\textwidth]{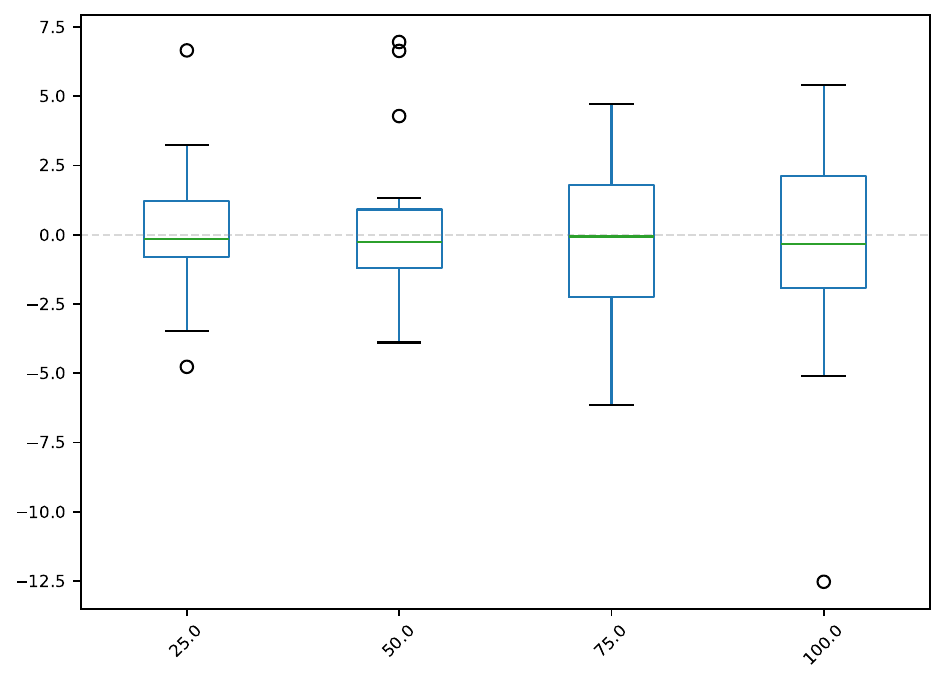} \\

Gemma 2 9b  (qual.) &
\includegraphics[width=0.2\textwidth]{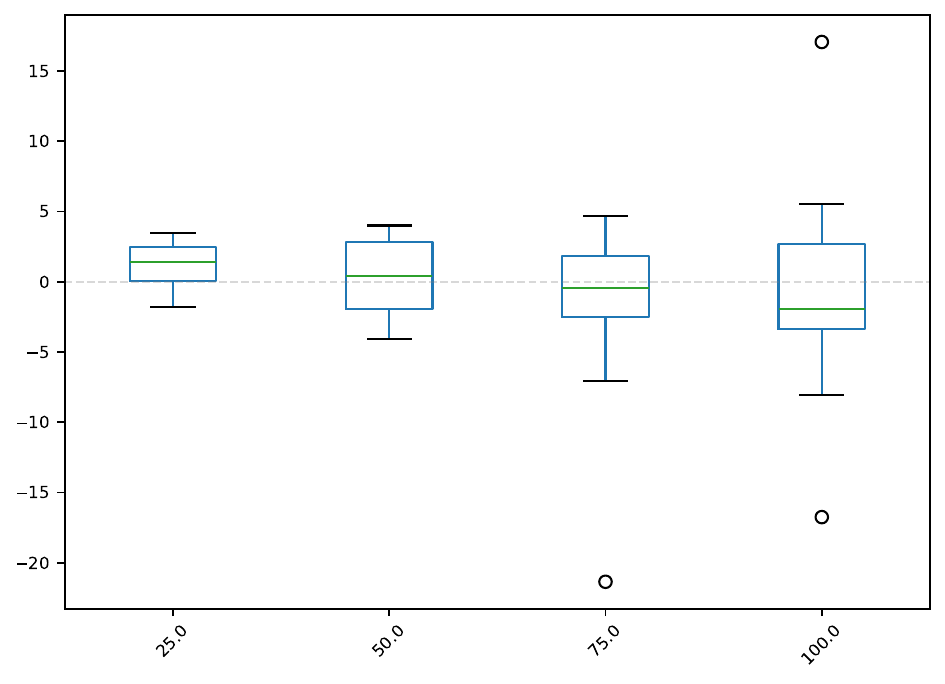} &
\includegraphics[width=0.2\textwidth]{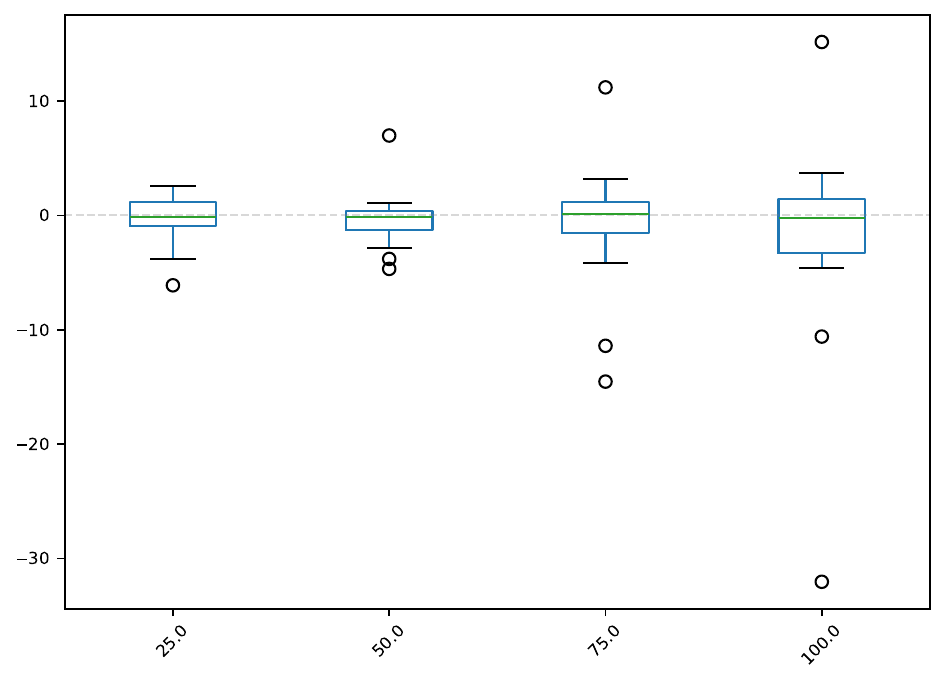} &
\includegraphics[width=0.2\textwidth]{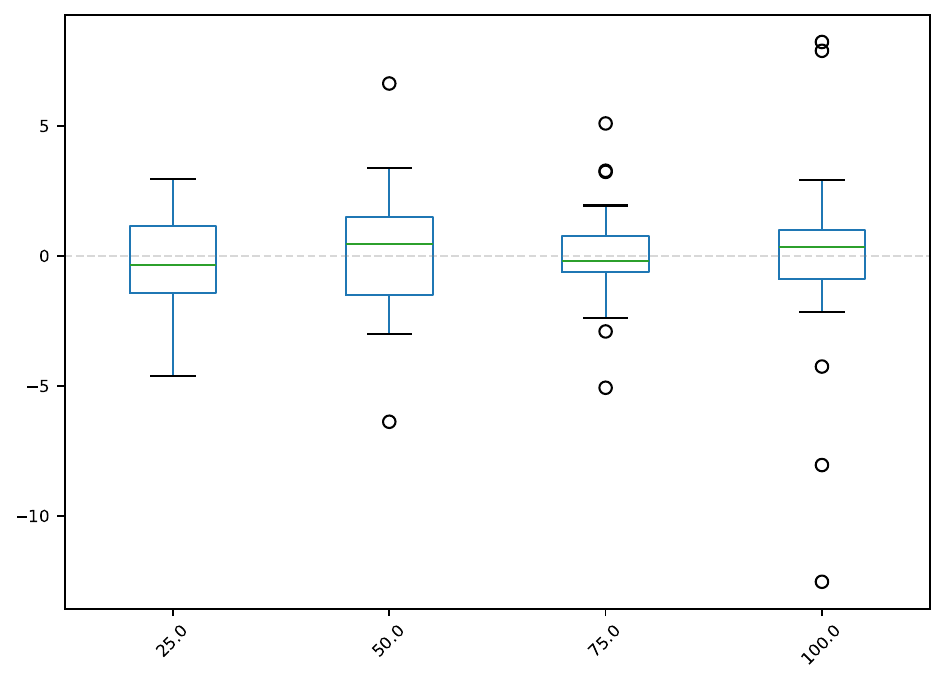} \\

Llama 3.1 70b  (lang.) &
\includegraphics[width=0.2\textwidth]{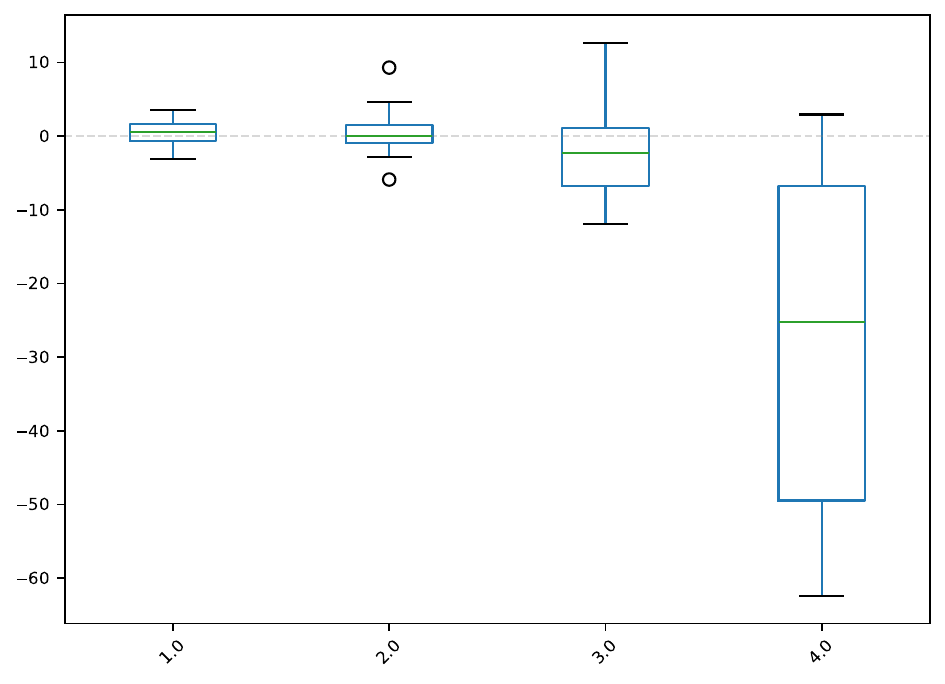} &
\includegraphics[width=0.2\textwidth]{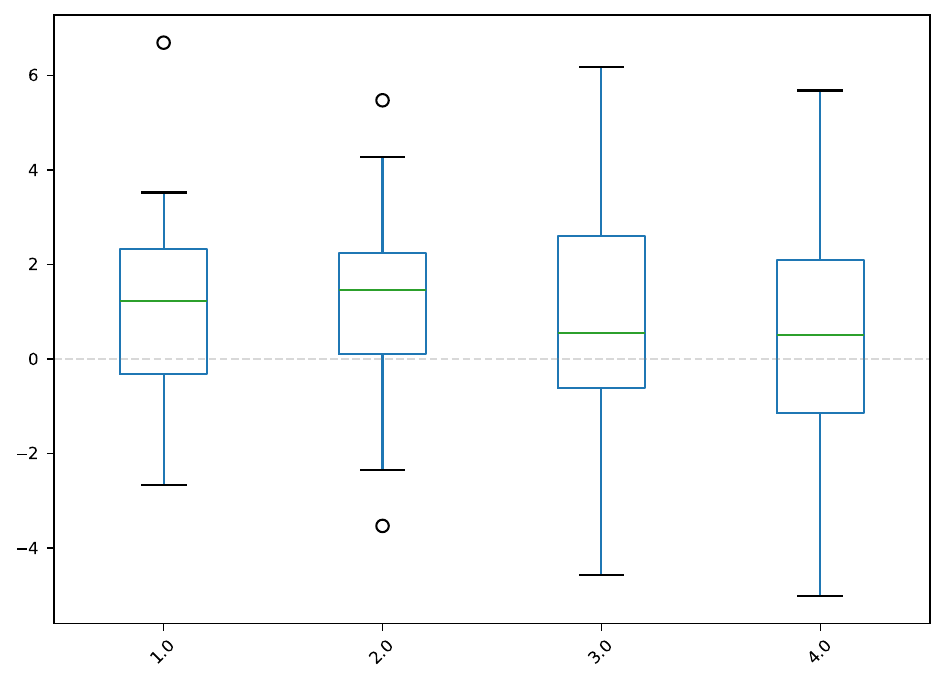} &
\includegraphics[width=0.2\textwidth]{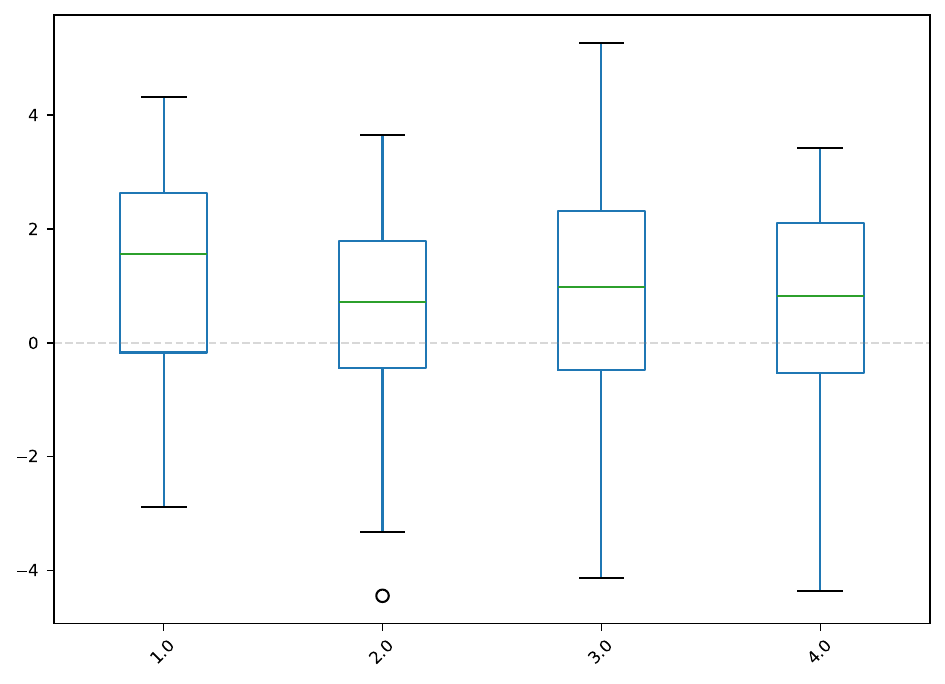} \\
\bottomrule
\end{tabular}
\end{table*}

\begin{table*}[t]
\centering
\tiny
\caption{Boxplot visualizations of relative increases/decreases in \% for F1 metrics for LLM + steering methods for various languages in \textbf{zero-shot} setting. The boxplots are for $\alpha$ values from Table~\ref{tab:steering_results}. The X axis in the figures represents the languages, and the Y axis represents the relative difference increase for F1 downstream model performance.}
\label{tab:results_zero_shot_all}
\renewcommand{\arraystretch}{1.2}
\begin{tabular}{>{\centering\arraybackslash}m{0.1\textwidth}
                >{\centering\arraybackslash}m{0.2\textwidth}
                >{\centering\arraybackslash}m{0.2\textwidth}
                >{\centering\arraybackslash}m{0.2\textwidth}}
\toprule
\textbf{LLM + steering} &
\textbf{Early layer} &
\textbf{Middle layer} &
\textbf{Late layer}  \\
\midrule

Gemma 2 27b  (lang.) &
\includegraphics[width=0.2\textwidth]{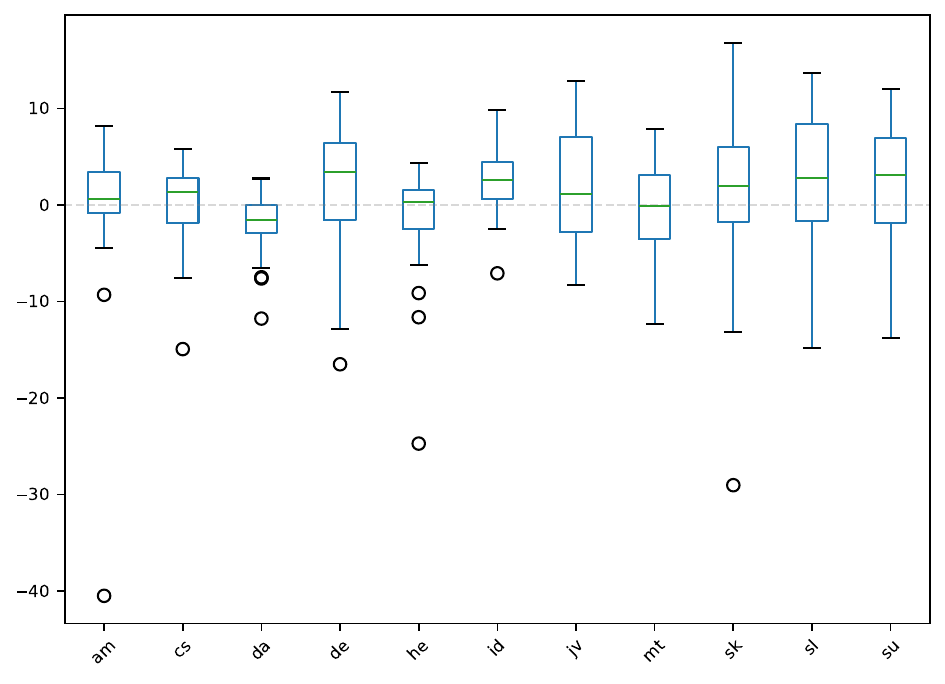} &
\includegraphics[width=0.2\textwidth]{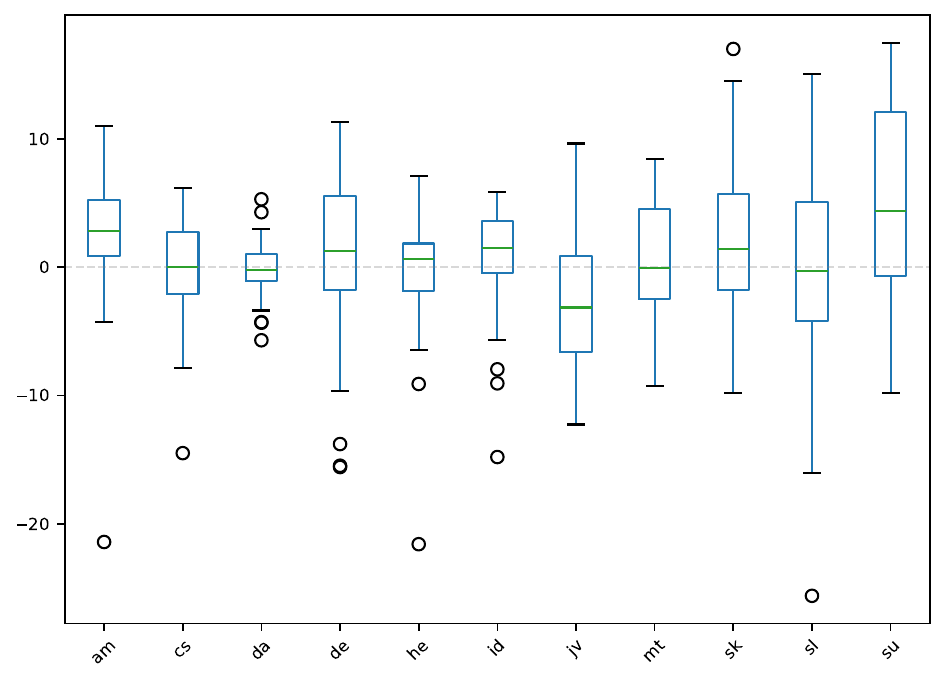} &
\includegraphics[width=0.2\textwidth]{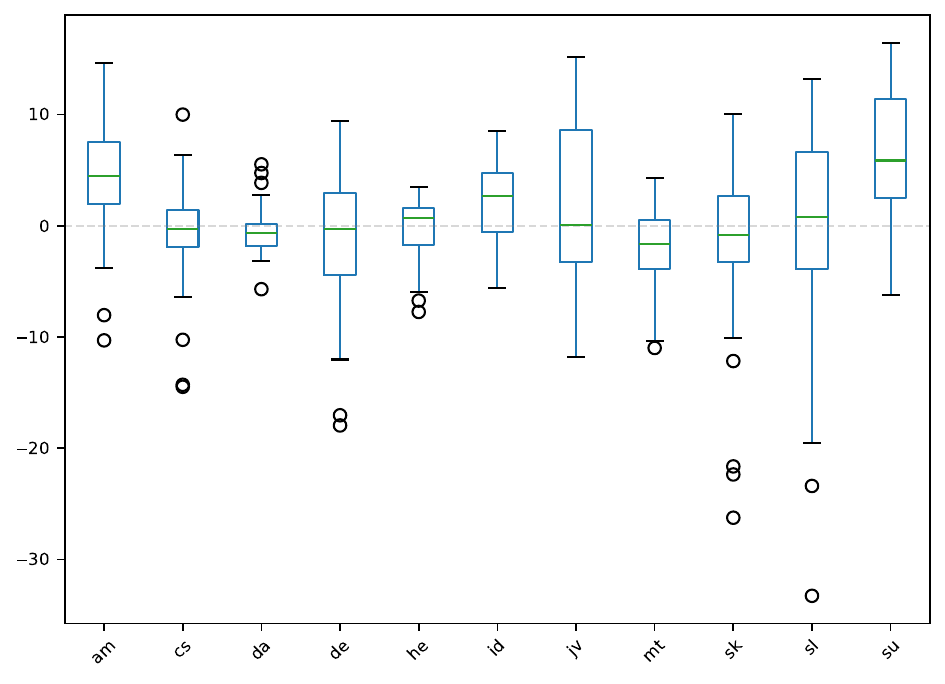} \\

Gemma 2 27b  (qual.) &
\includegraphics[width=0.2\textwidth]{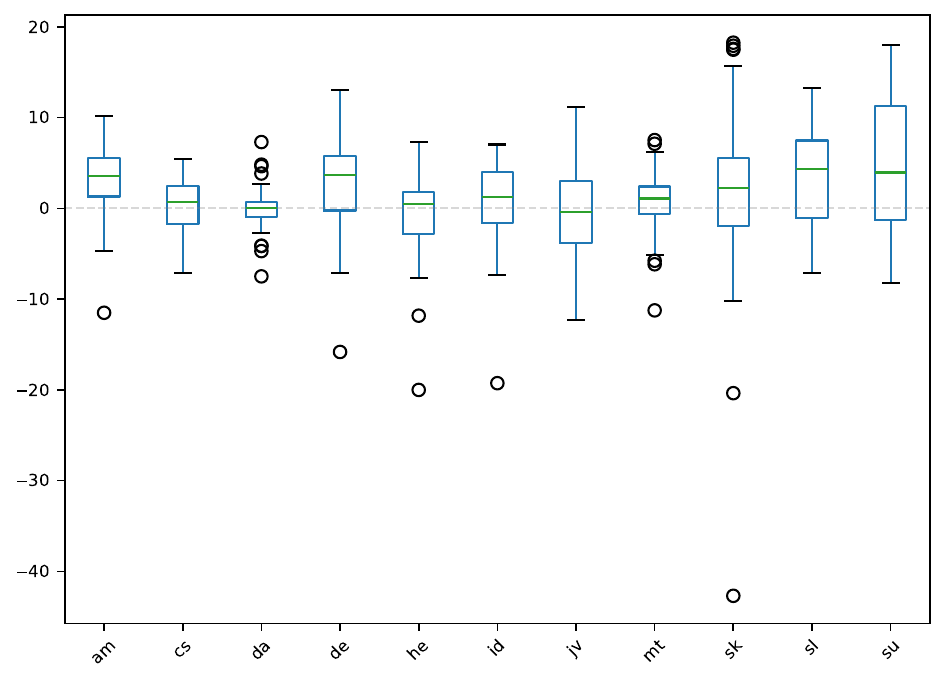} &
\includegraphics[width=0.2\textwidth]{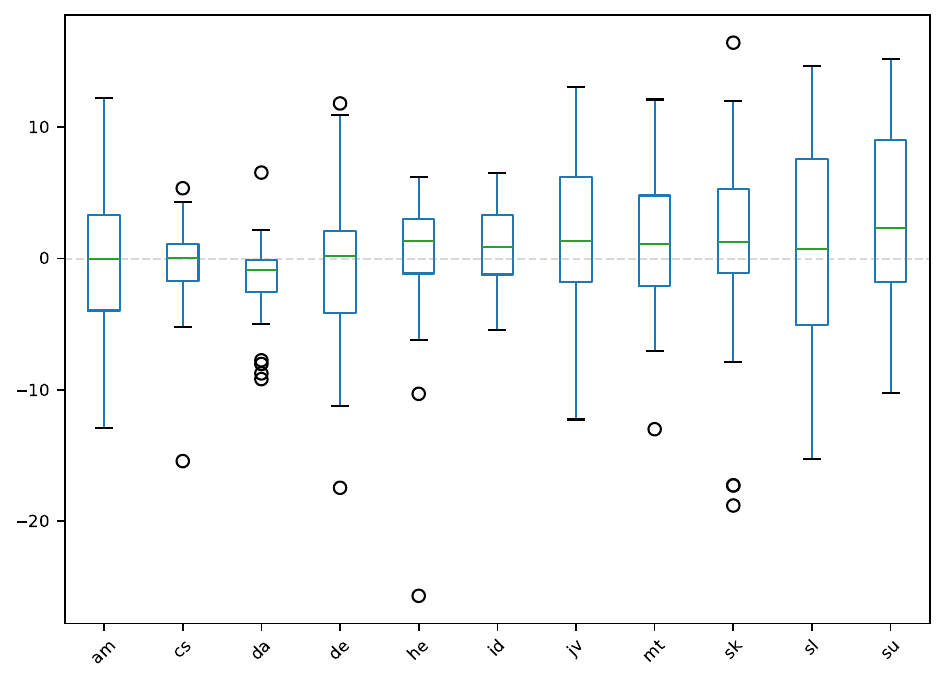} &
\includegraphics[width=0.2\textwidth]{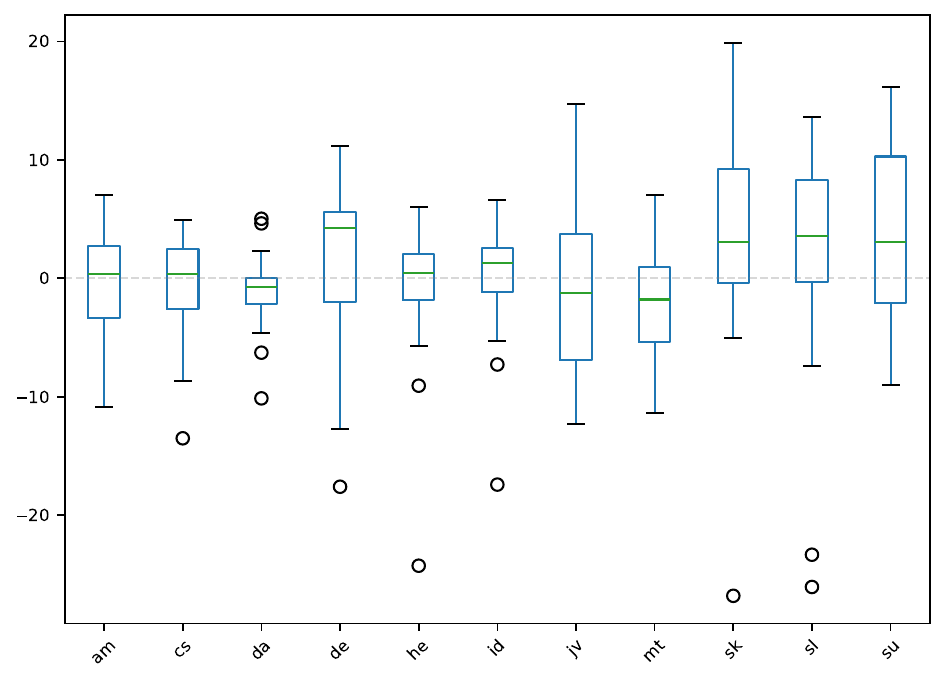} \\

Gemma 2 9b  (lang.) &
\includegraphics[width=0.2\textwidth]{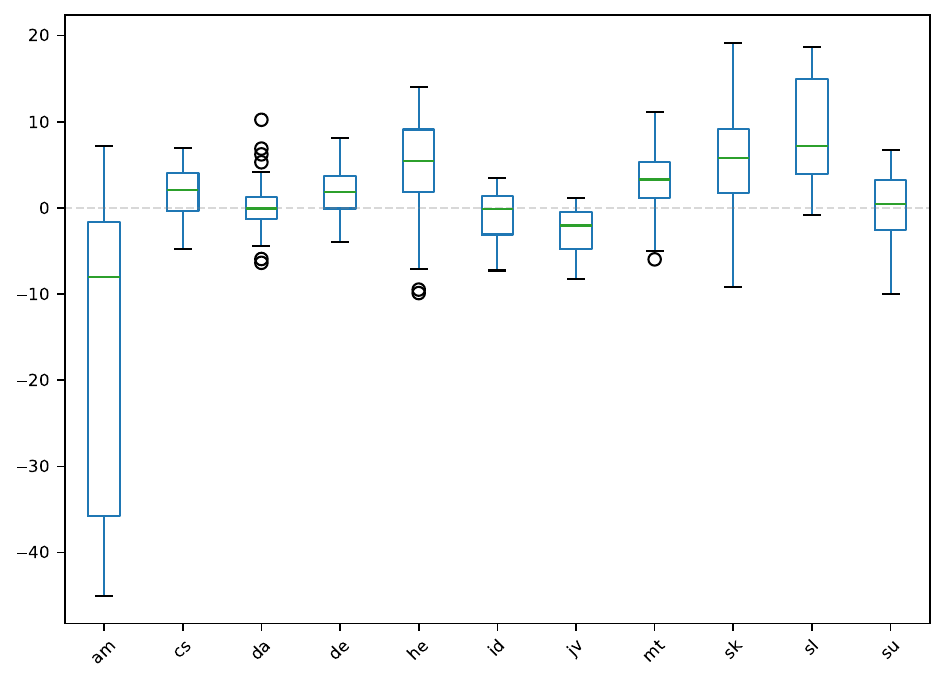} &
\includegraphics[width=0.2\textwidth]{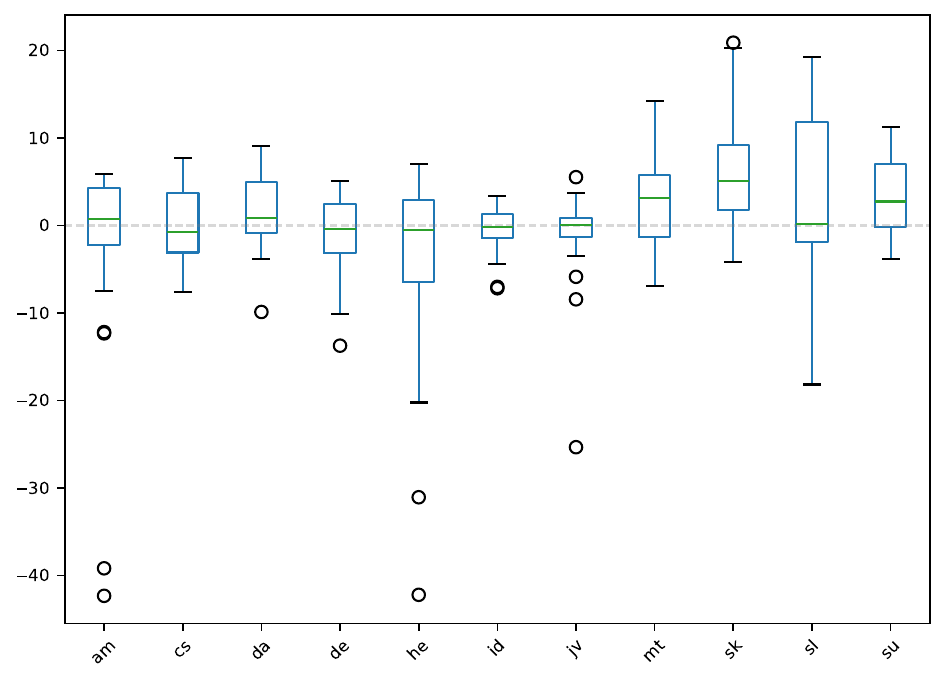} &
\includegraphics[width=0.2\textwidth]{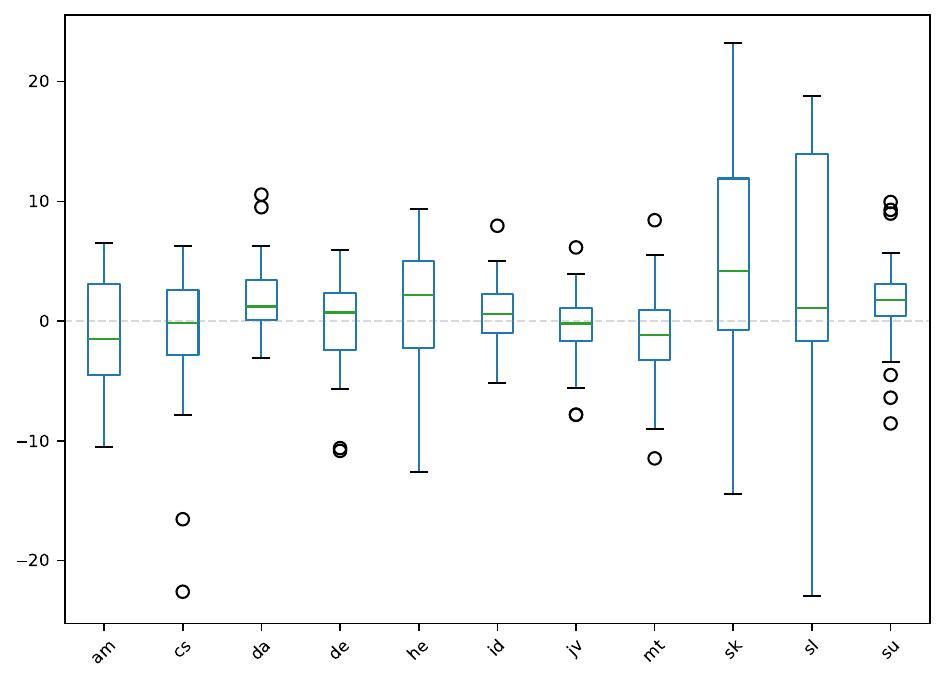} \\

Gemma 2 9b  (qual.) &
\includegraphics[width=0.2\textwidth]{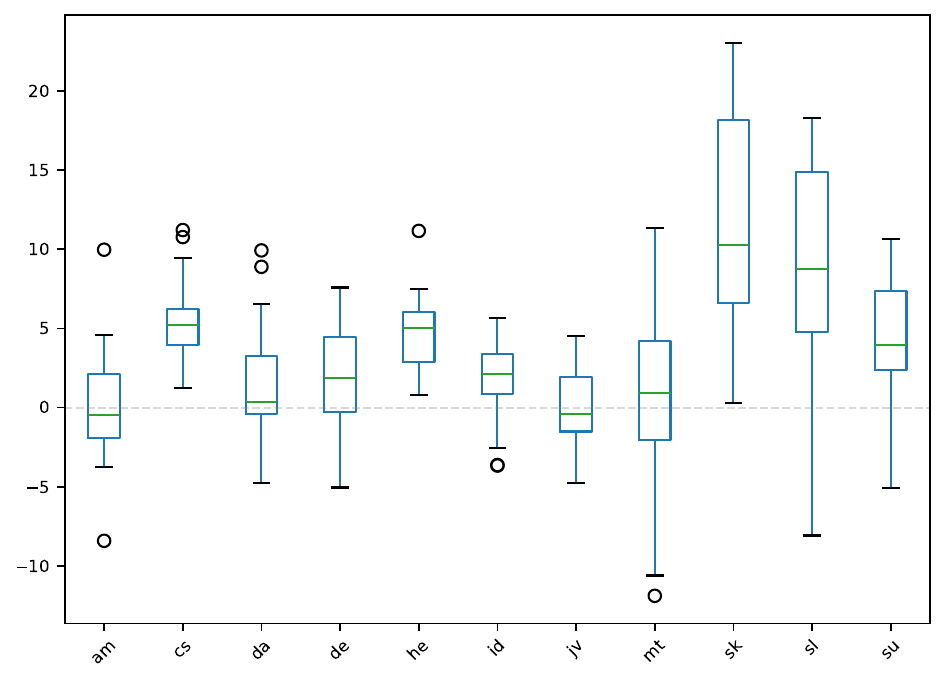} &
\includegraphics[width=0.2\textwidth]{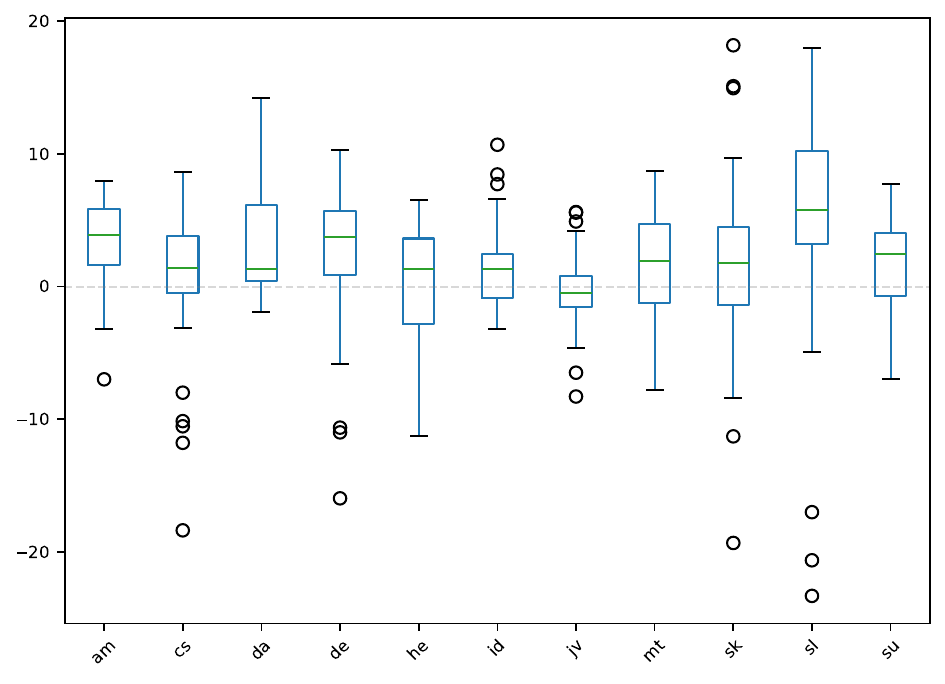} &
\includegraphics[width=0.2\textwidth]{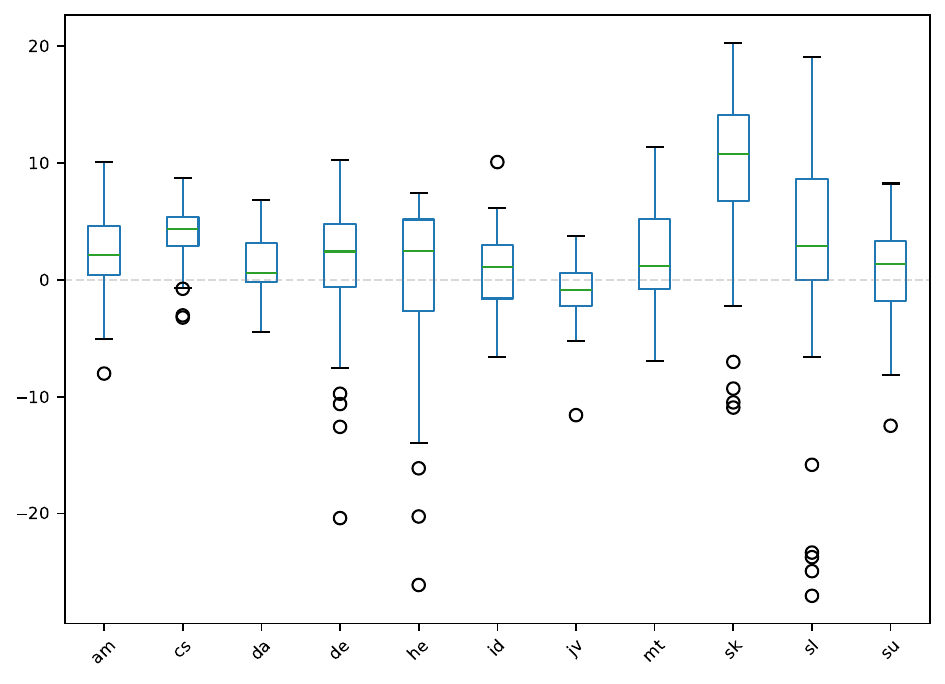} \\

Llama 3.1 70b  (lang.) &
\includegraphics[width=0.2\textwidth]{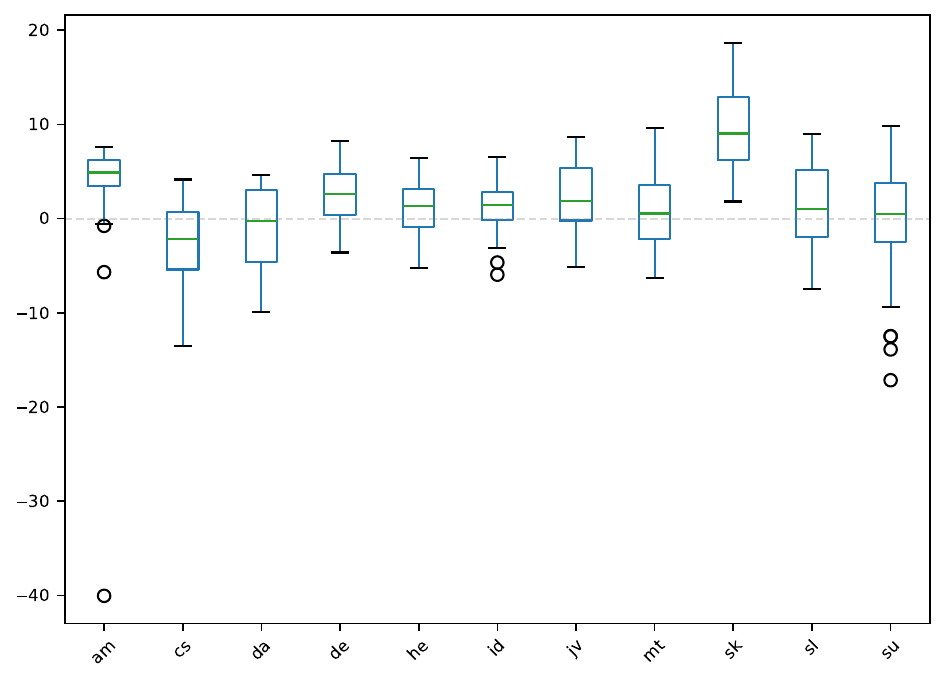} &
\includegraphics[width=0.2\textwidth]{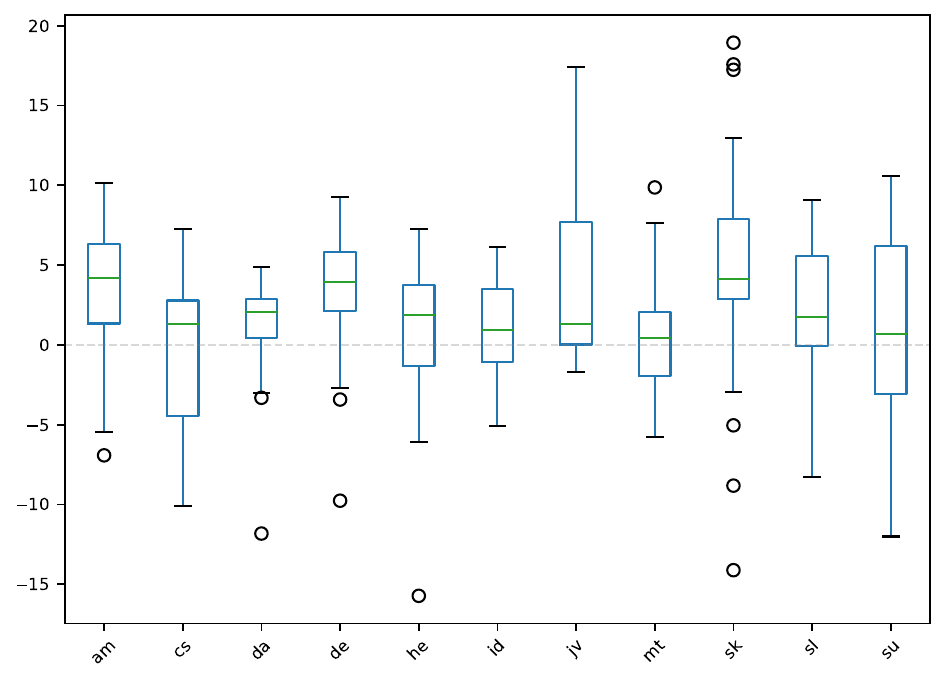} &
\includegraphics[width=0.2\textwidth]{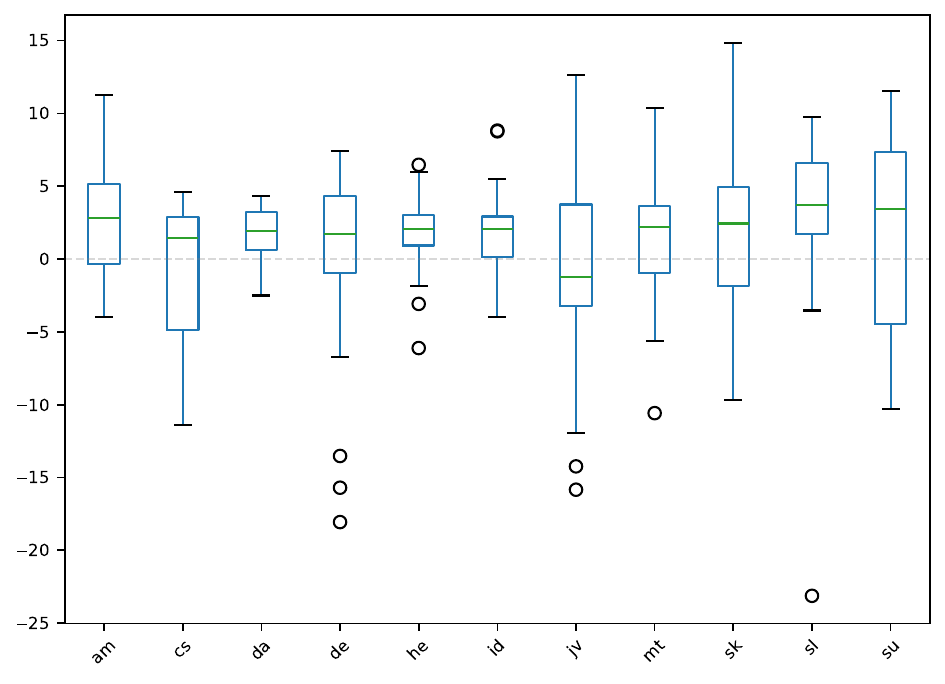} \\
\bottomrule
\end{tabular}
\end{table*}

\begin{table*}[t]
\centering
\tiny
\caption{Boxplot visualizations of relative increases/decreases in \% for F1 metrics for LLM + steering methods for various languages in \textbf{few-shot} setting. The boxplots are for $\alpha$ values from Table~\ref{tab:steering_results}. The X axis in the figures represents the languages, and the Y axis represents the relative difference increase for F1 downstream model performance.}
\label{tab:results_few_shot_all}
\renewcommand{\arraystretch}{1.2}

\begin{tabular}{>{\centering\arraybackslash}m{0.1\textwidth}
                >{\centering\arraybackslash}m{0.2\textwidth}
                >{\centering\arraybackslash}m{0.2\textwidth}
                >{\centering\arraybackslash}m{0.2\textwidth}}
\toprule
\textbf{LLM + steering} &
\textbf{Early layer} &
\textbf{Middle layer} &
\textbf{Late layer}  \\
\midrule

Gemma 2 27b  (lang.) &
\includegraphics[width=0.2\textwidth]{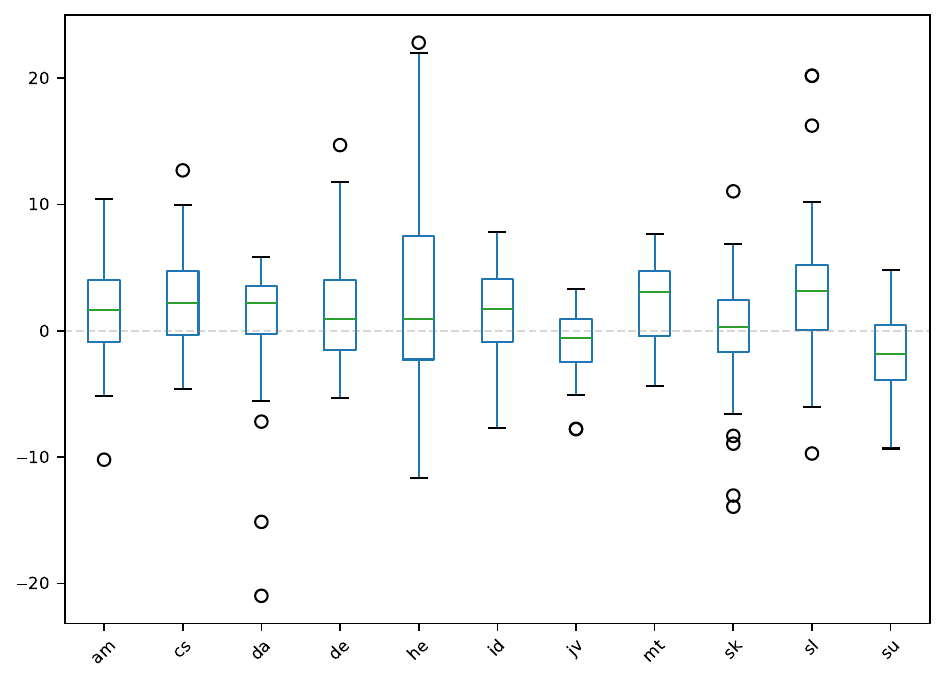} &
\includegraphics[width=0.2\textwidth]{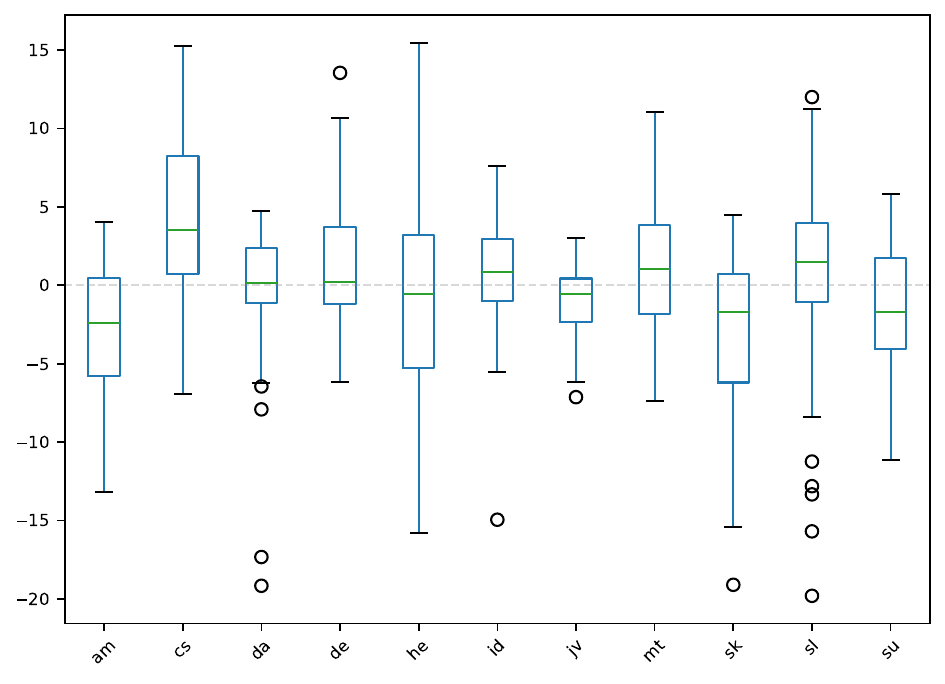} &
\includegraphics[width=0.2\textwidth]{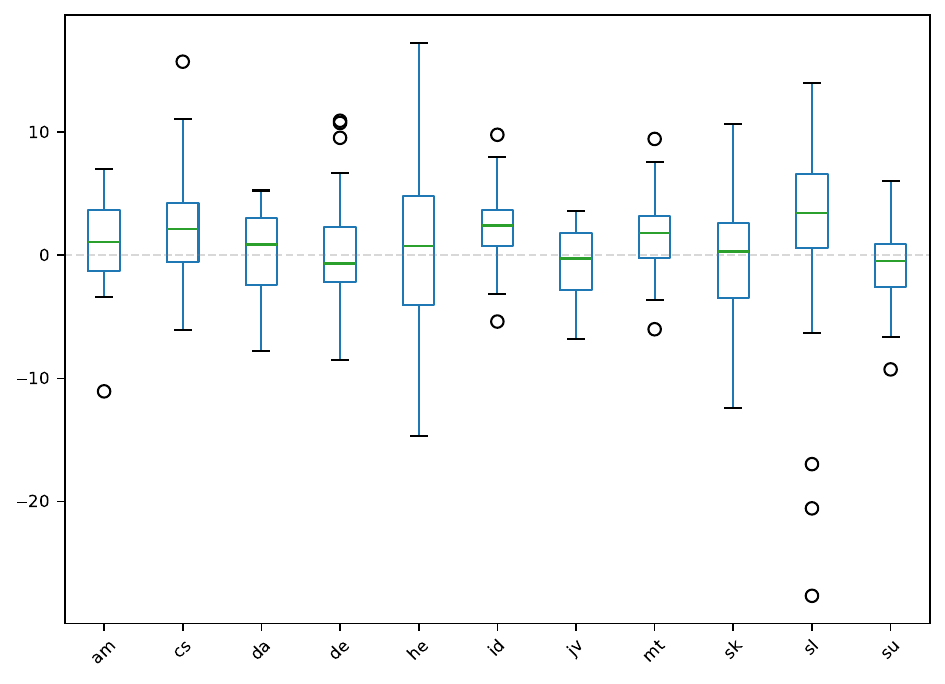} \\

Gemma 2 27b  (qual.) &
\includegraphics[width=0.2\textwidth]{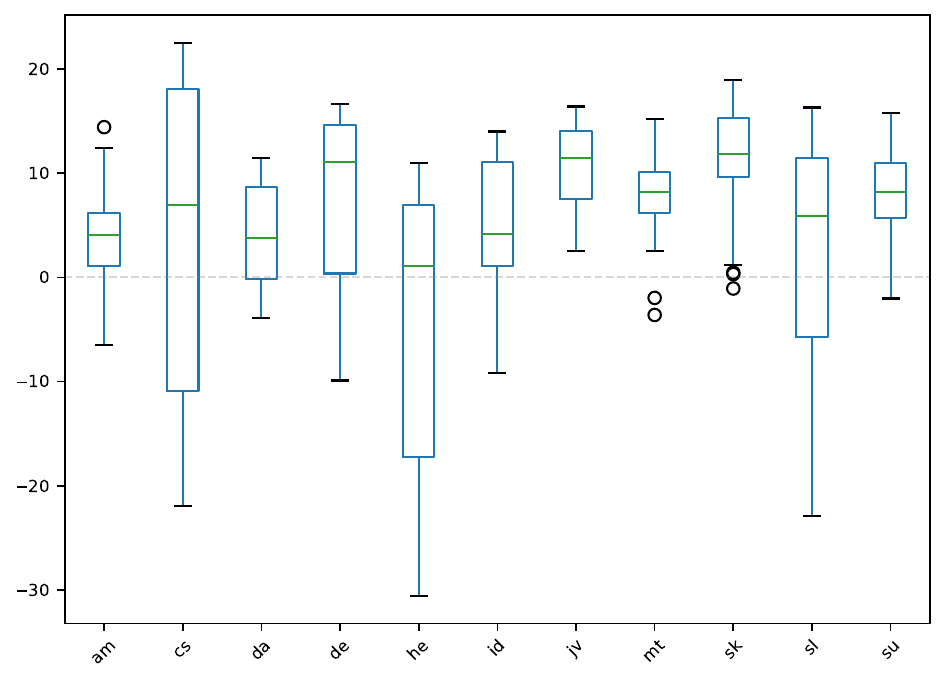} &
\includegraphics[width=0.2\textwidth]{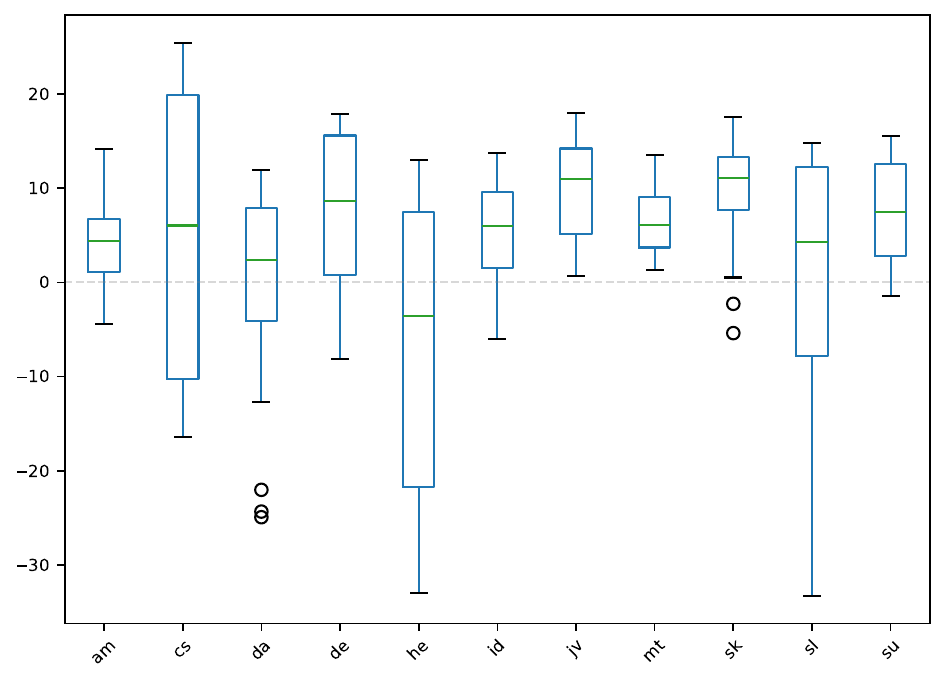} &
\includegraphics[width=0.2\textwidth]{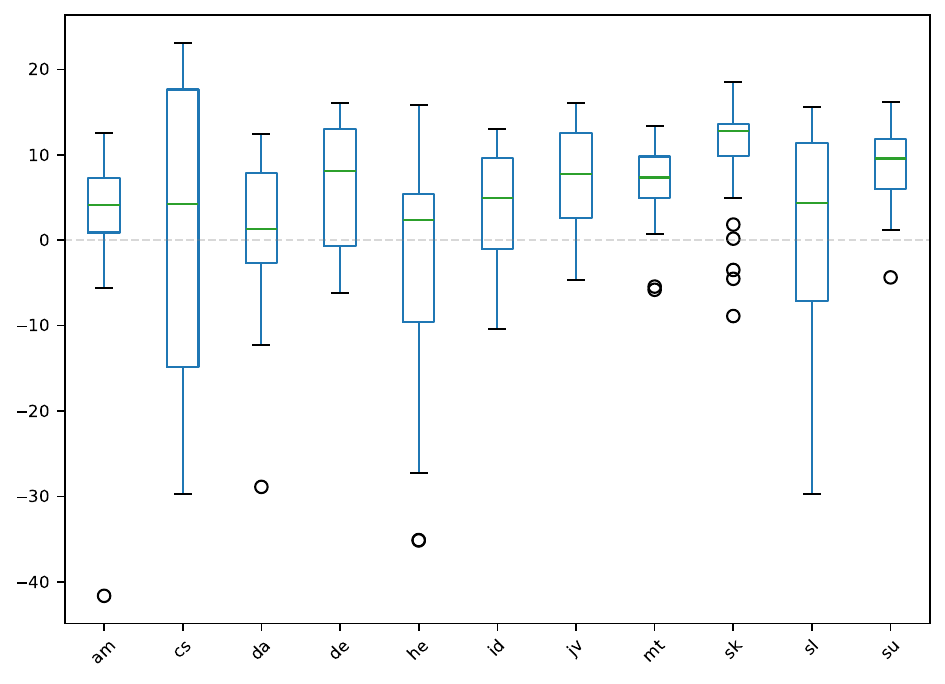} \\

Gemma 2 9b  (lang.) &
\includegraphics[width=0.2\textwidth]{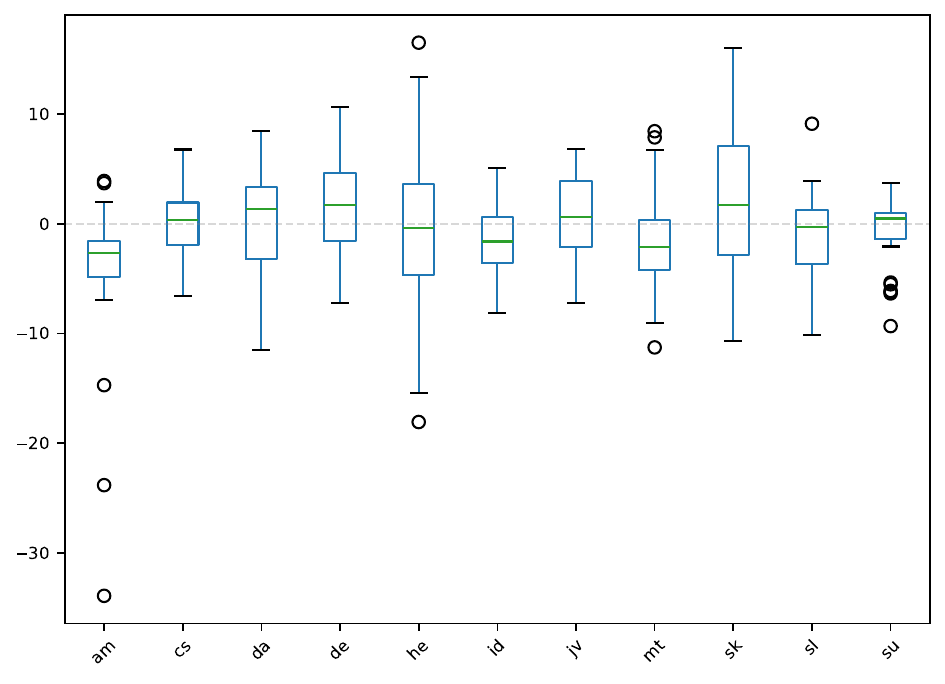} &
\includegraphics[width=0.2\textwidth]{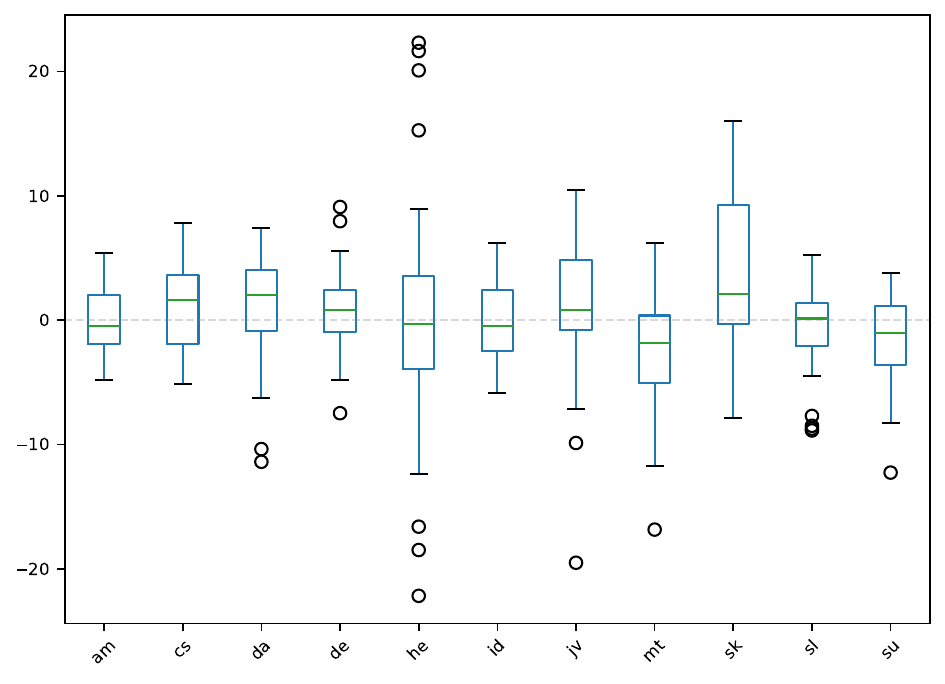} &
\includegraphics[width=0.2\textwidth]{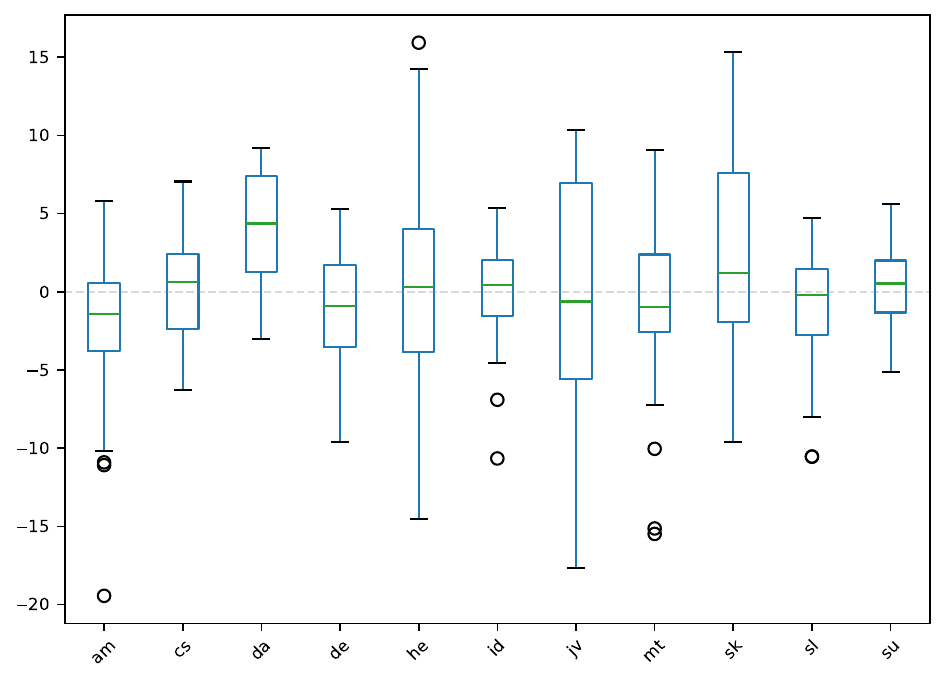} \\

Gemma 2 9b  (qual.) &
\includegraphics[width=0.2\textwidth]{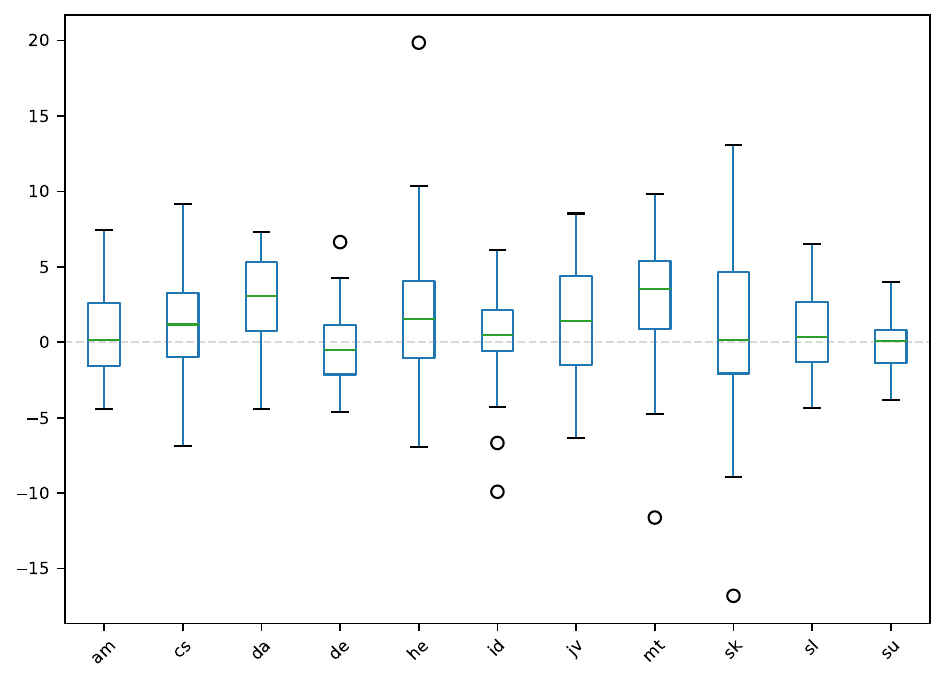} &
\includegraphics[width=0.2\textwidth]{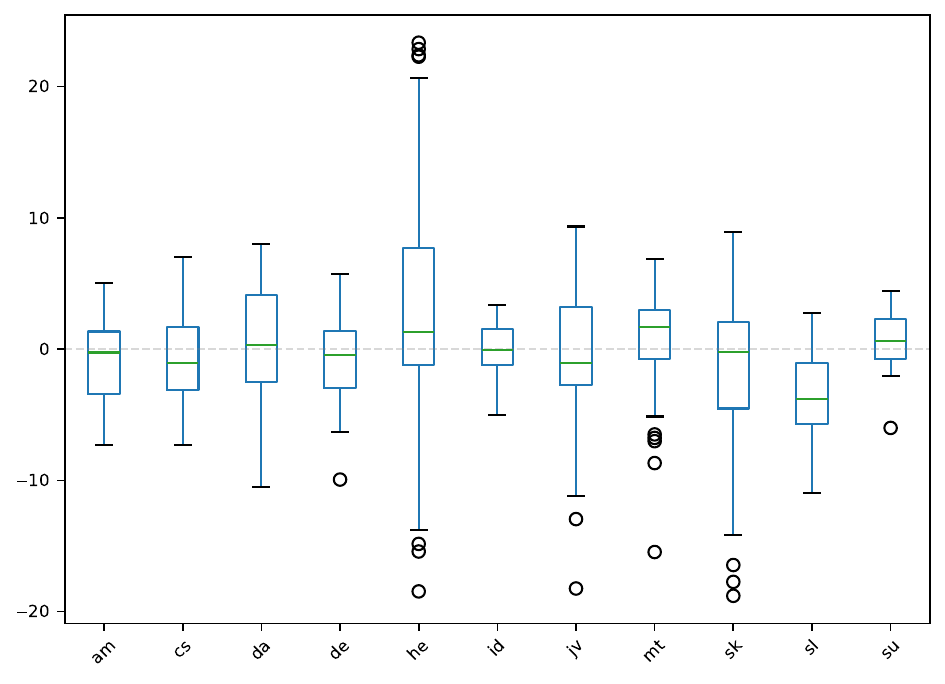} &
\includegraphics[width=0.2\textwidth]{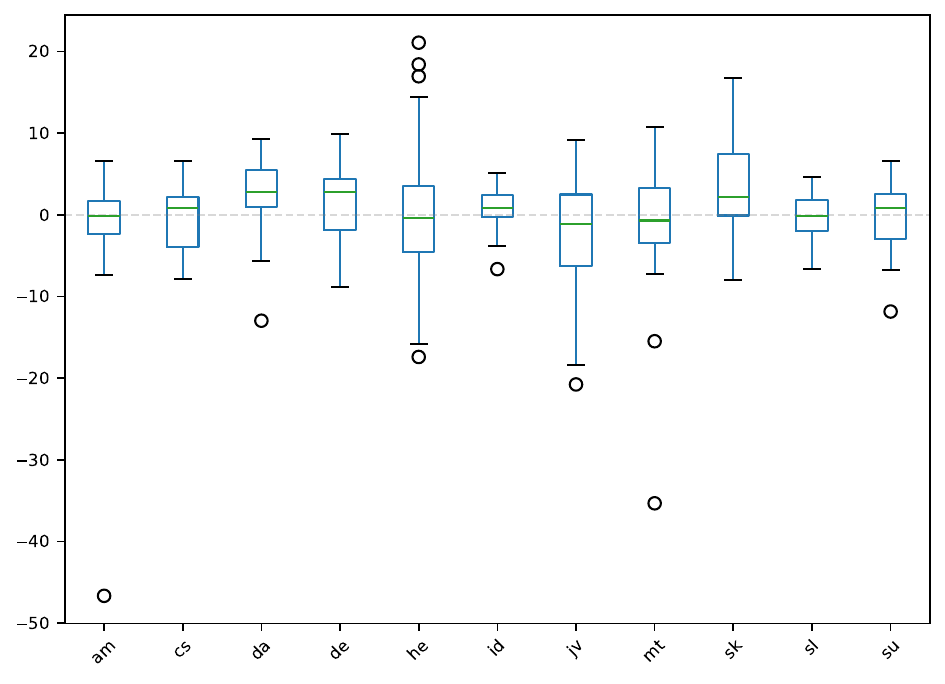} \\

Llama 3.1 70b  (lang.) &
\includegraphics[width=0.2\textwidth]{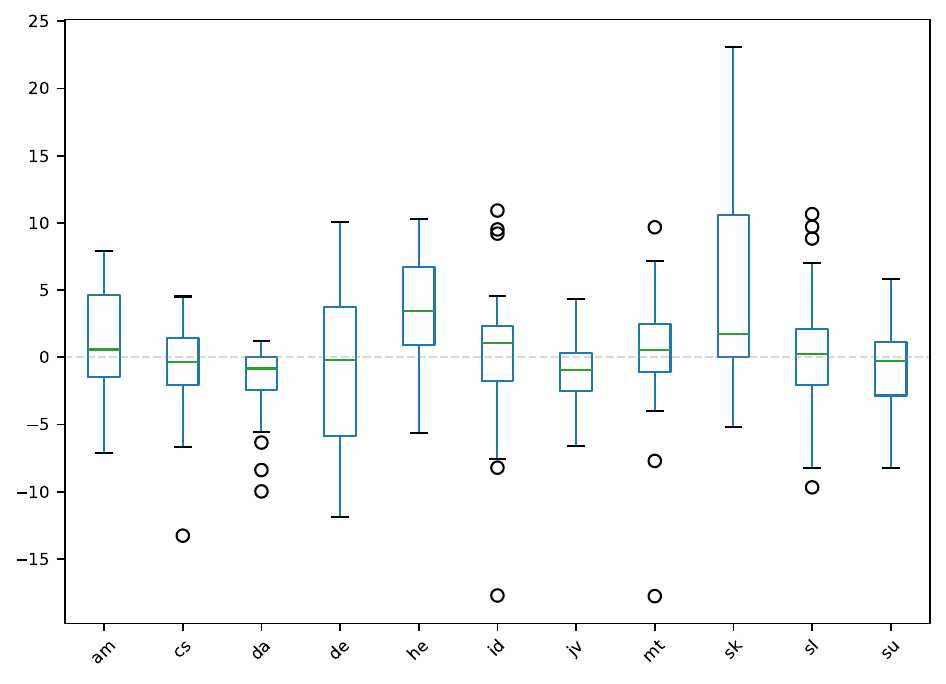} &
\includegraphics[width=0.2\textwidth]{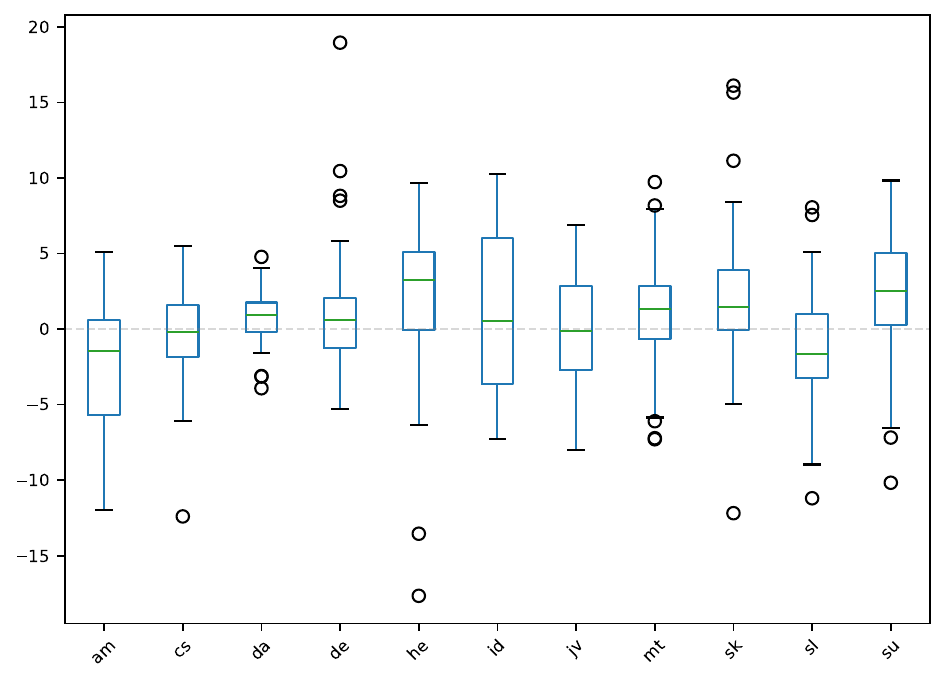} &
\includegraphics[width=0.2\textwidth]{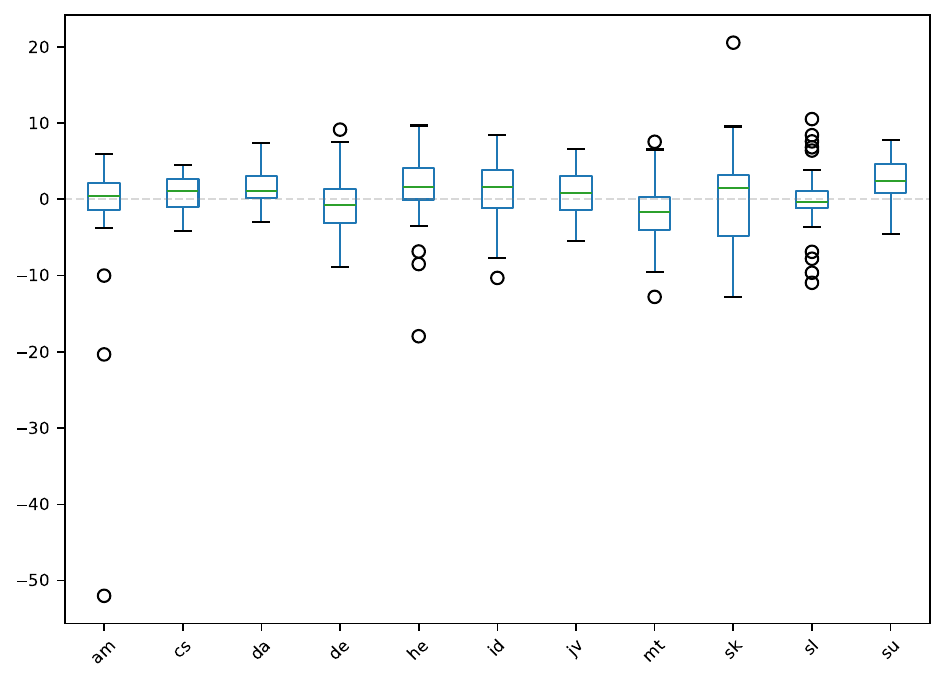} \\
\bottomrule
\end{tabular}
\end{table*}

\begin{table*}[t]
\centering
\tiny
\caption{Relative increases/decreases in \% for diversity metrics aggregated over LLMs and alphas for various languages and layers for \textbf{zero-shot} baseline. The X axis in the figures represents the languages, and the Y axis represents the relative difference increase for that given metric.}
\label{tab:diversity_all_zero}
\renewcommand{\arraystretch}{1.2}

\begin{tabular}{>{\centering\arraybackslash}m{0.1\textwidth}
                >{\centering\arraybackslash}m{0.18\textwidth}
                >{\centering\arraybackslash}m{0.18\textwidth}
                >{\centering\arraybackslash}m{0.18\textwidth}
                >{\centering\arraybackslash}m{0.18\textwidth}}
\toprule
\textbf{Layers + steering} &
\textbf{Lexical div.} &
\textbf{Embbeding div.} &
\textbf{Isocontour rad.} &
\textbf{Homogeneity}  \\
\midrule

Early layers  (qual.) &
\includegraphics[width=0.18\textwidth]{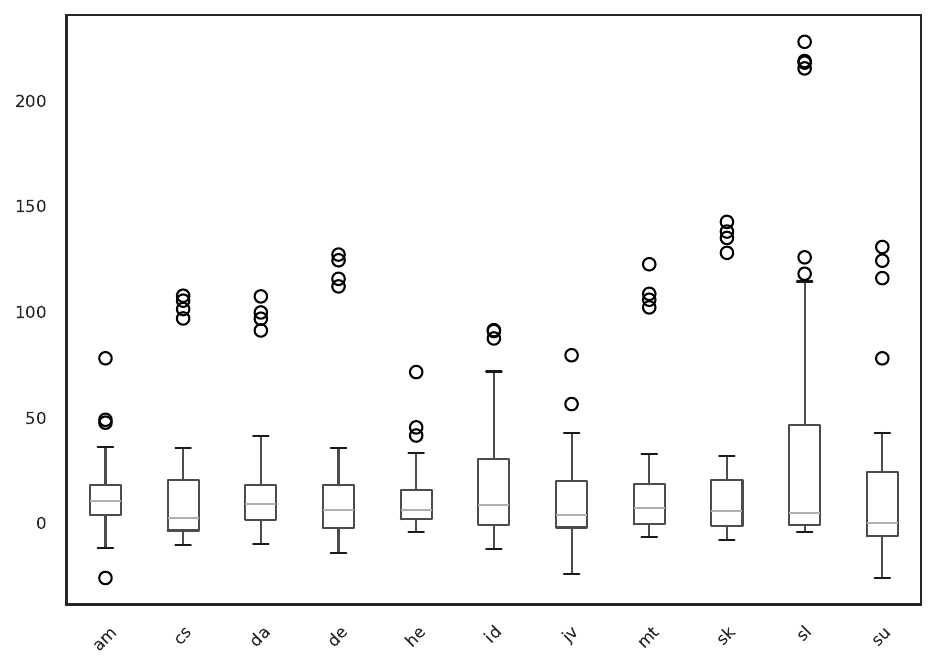} &
\includegraphics[width=0.18\textwidth]{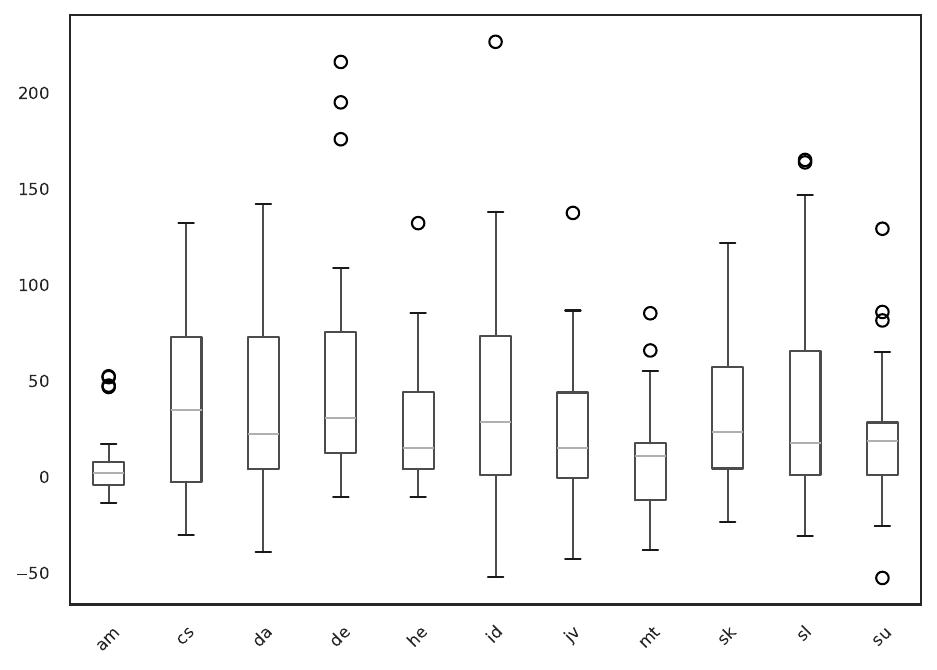} &
\includegraphics[width=0.18\textwidth]{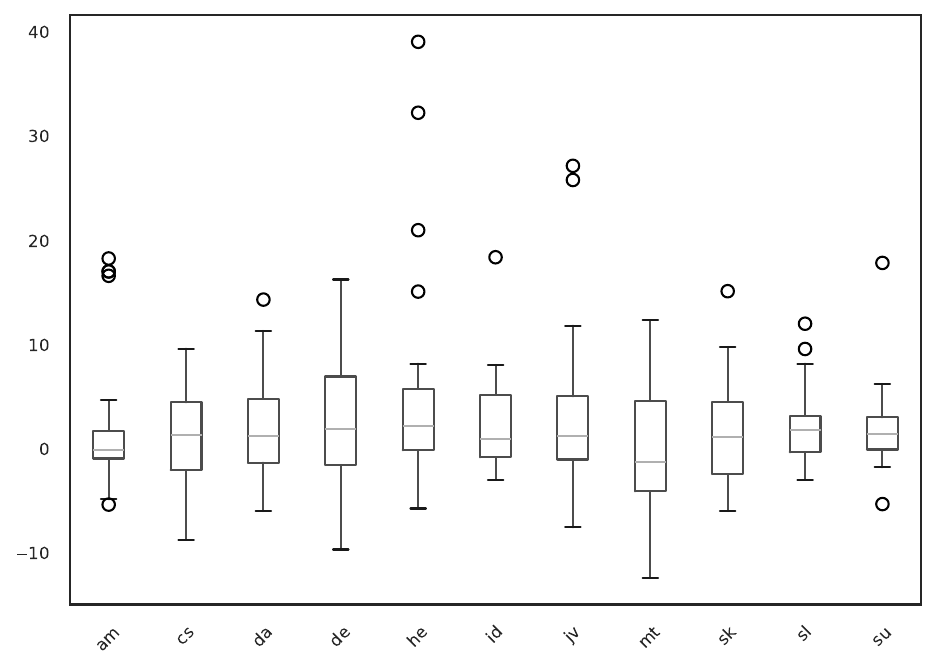} &
\includegraphics[width=0.18\textwidth]{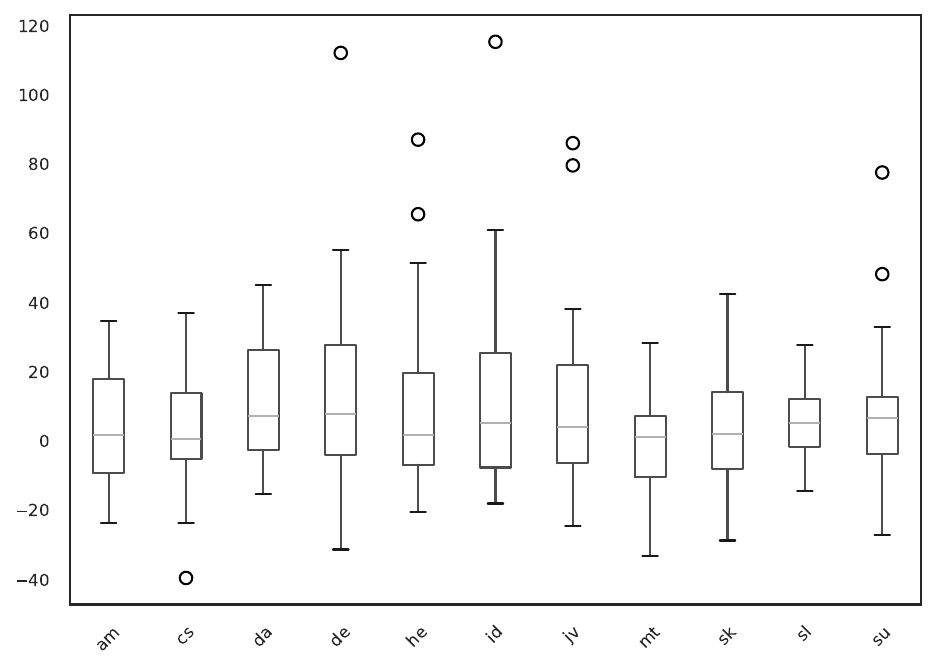}\\

Early layers  (lang.) &
\includegraphics[width=0.18\textwidth]{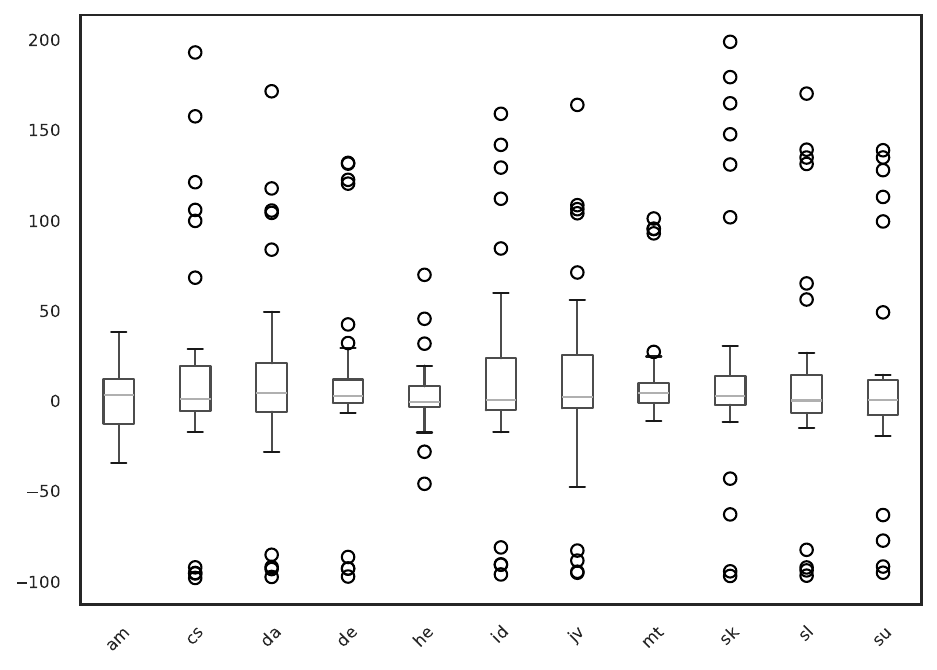} &
\includegraphics[width=0.18\textwidth]{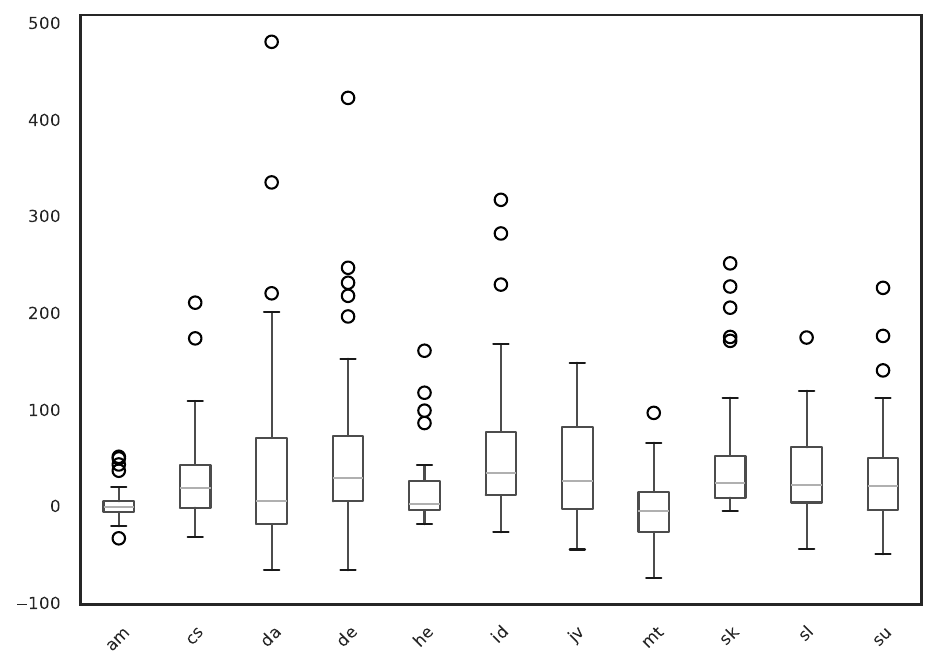} &
\includegraphics[width=0.18\textwidth]{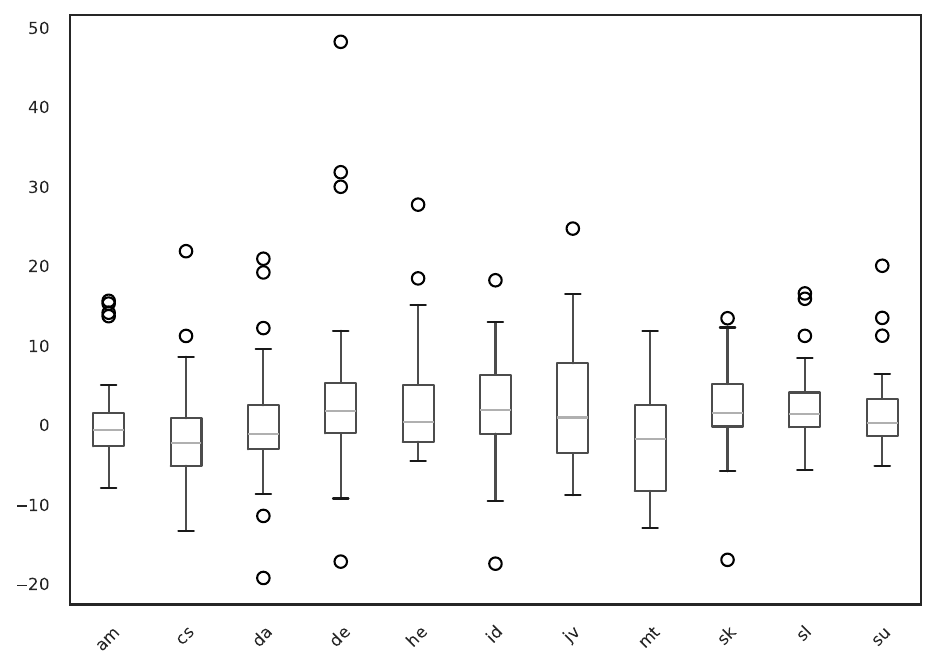} &
\includegraphics[width=0.18\textwidth]{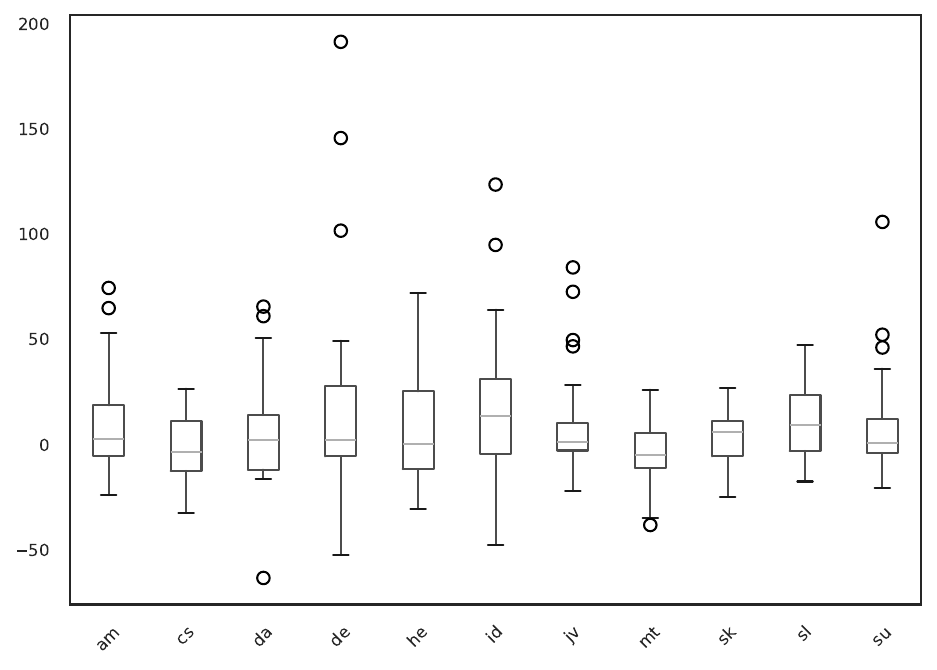}\\

Middle layers  (qual.) &
\includegraphics[width=0.18\textwidth]{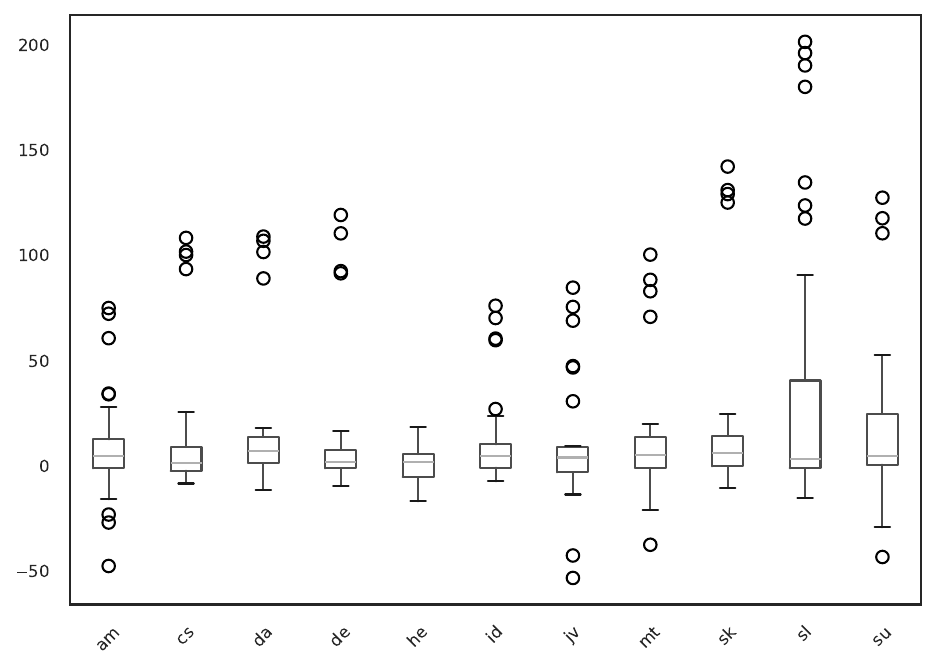} &
\includegraphics[width=0.18\textwidth]{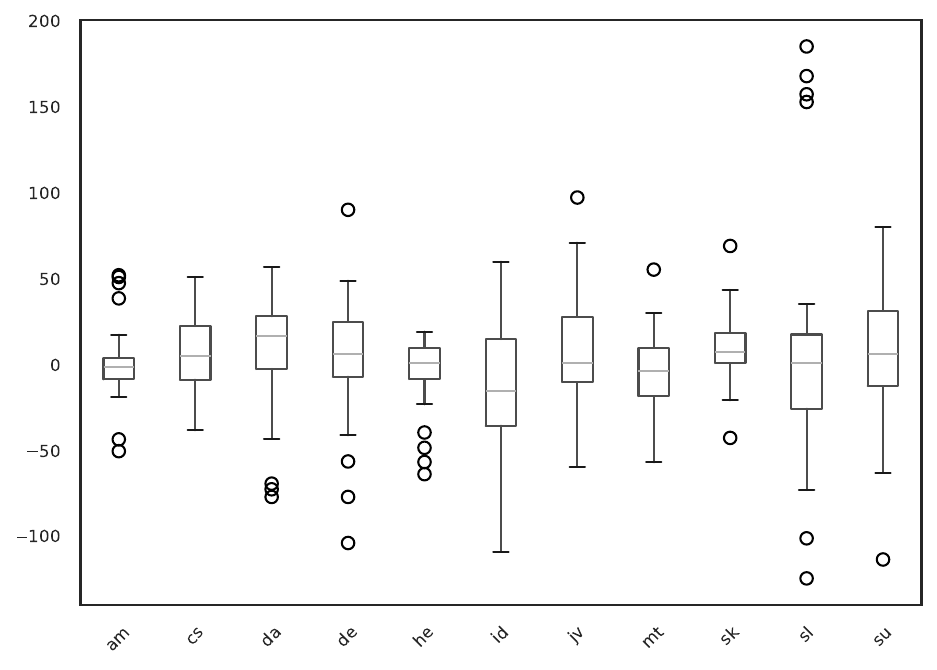} &
\includegraphics[width=0.18\textwidth]{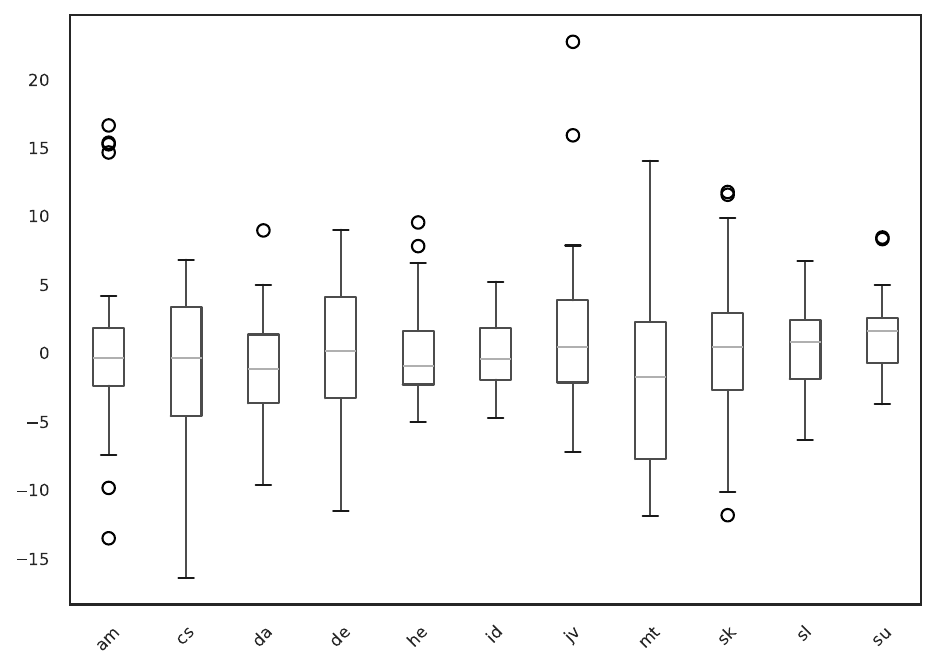} &
\includegraphics[width=0.18\textwidth]{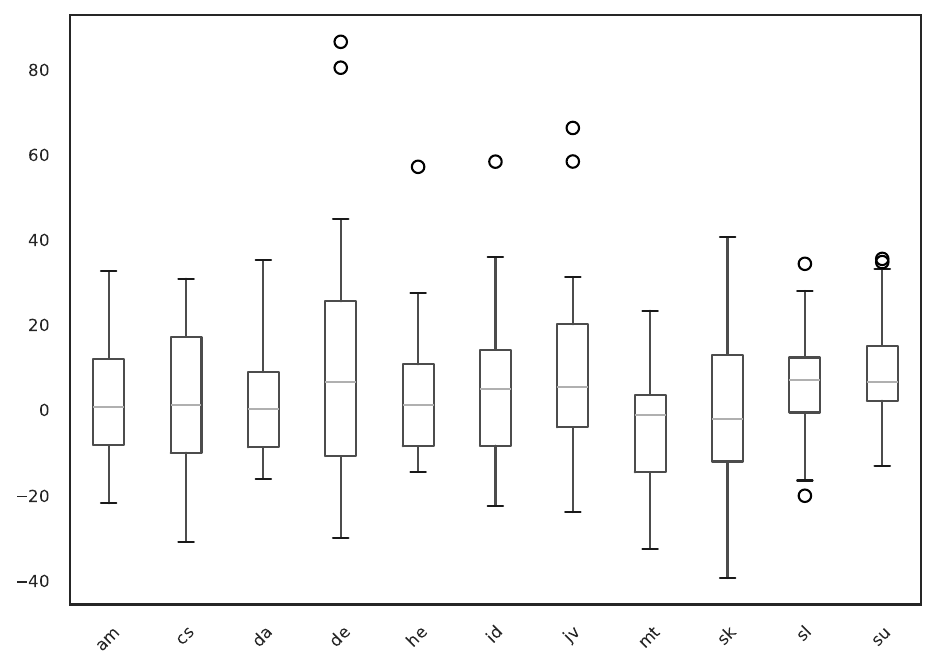}\\
\bottomrule
\end{tabular}
\end{table*}

\begin{table*}[t]
\centering
\tiny
\caption{Relative increases/decreases in \% for diversity metrics aggregated over LLMs and alphas for various languages and layers for \textbf{few-shot} baseline. The X axis in the figures represents the languages, and the Y axis represents the relative difference increase for that given metric.}
\label{tab:diversity_all_few}
\renewcommand{\arraystretch}{1.2}

\begin{tabular}{>{\centering\arraybackslash}m{0.1\textwidth}
                >{\centering\arraybackslash}m{0.18\textwidth}
                >{\centering\arraybackslash}m{0.18\textwidth}
                >{\centering\arraybackslash}m{0.18\textwidth}
                >{\centering\arraybackslash}m{0.18\textwidth}}
\toprule
\textbf{Layers + steering} &
\textbf{Lexical div.} &
\textbf{Embbeding div.} &
\textbf{Isocontour rad.} &
\textbf{Homogeneity}  \\
\midrule

Early layers  (qual.) &
\includegraphics[width=0.18\textwidth]{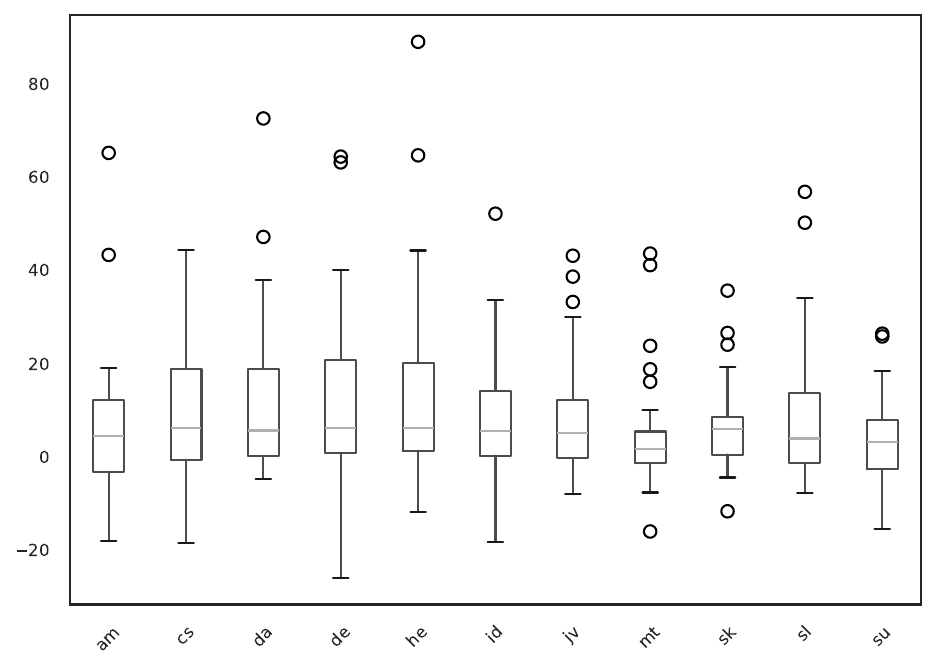} &
\includegraphics[width=0.18\textwidth]{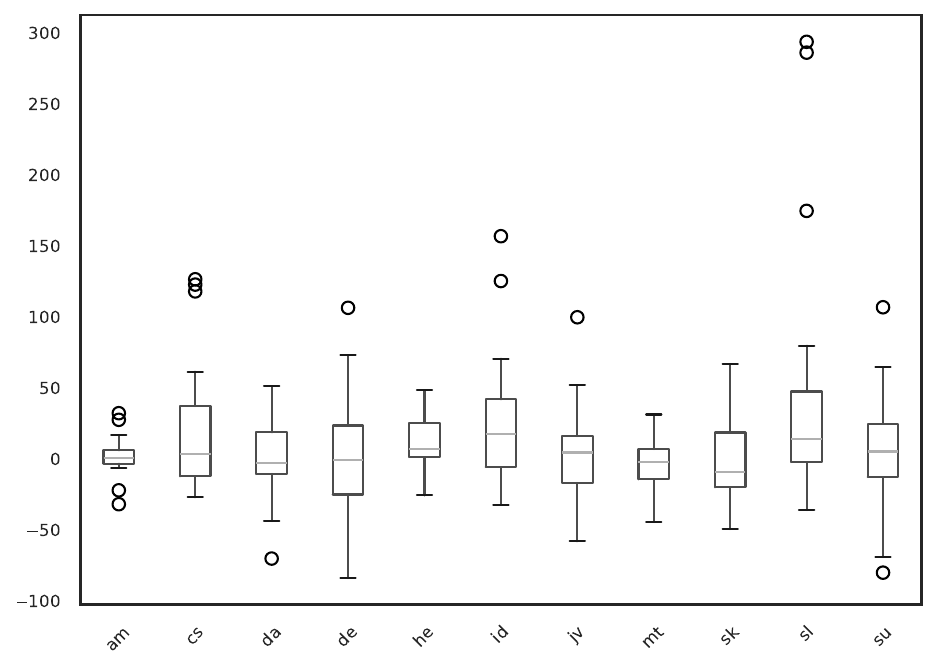} &
\includegraphics[width=0.18\textwidth]{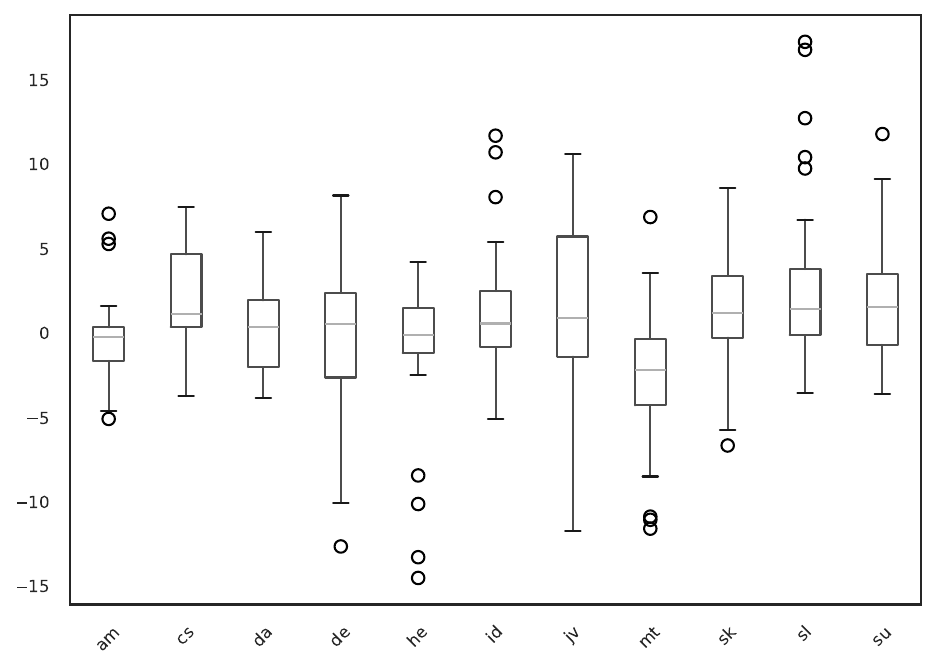} &
\includegraphics[width=0.18\textwidth]{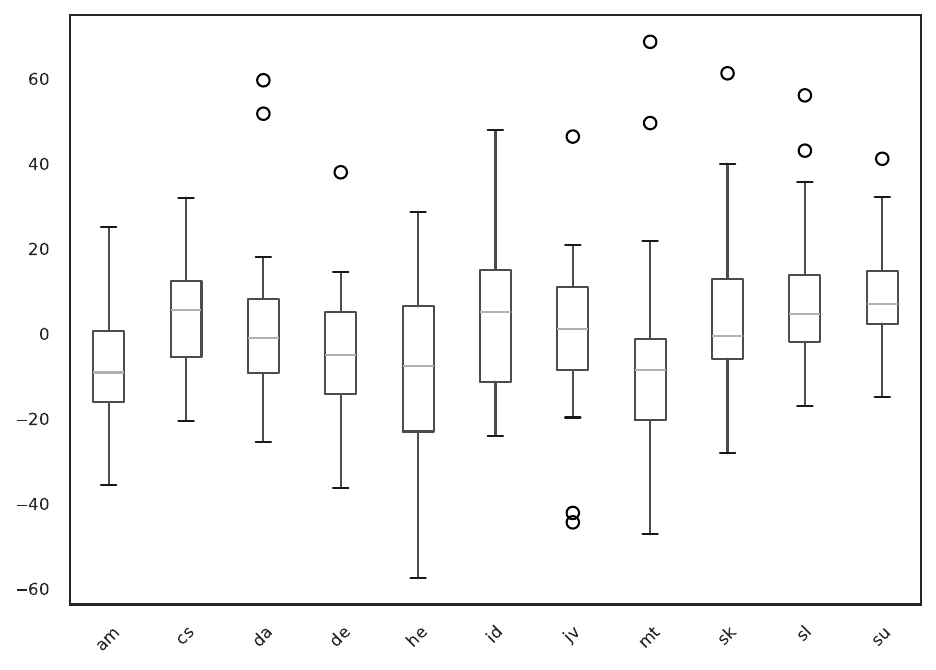}\\

Early layers  (lang.) &
\includegraphics[width=0.18\textwidth]{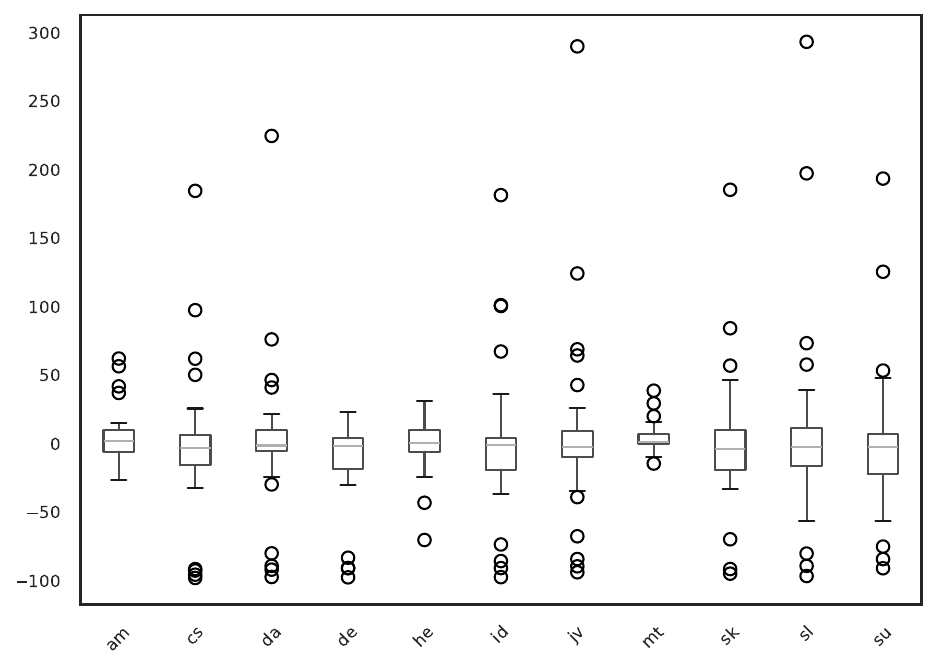} &
\includegraphics[width=0.18\textwidth]{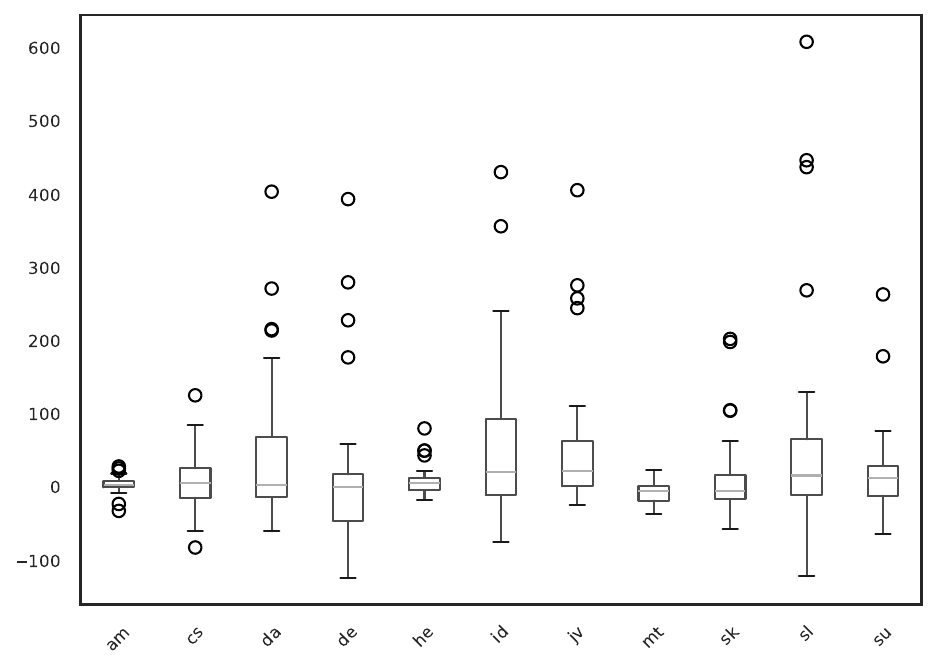} &
\includegraphics[width=0.18\textwidth]{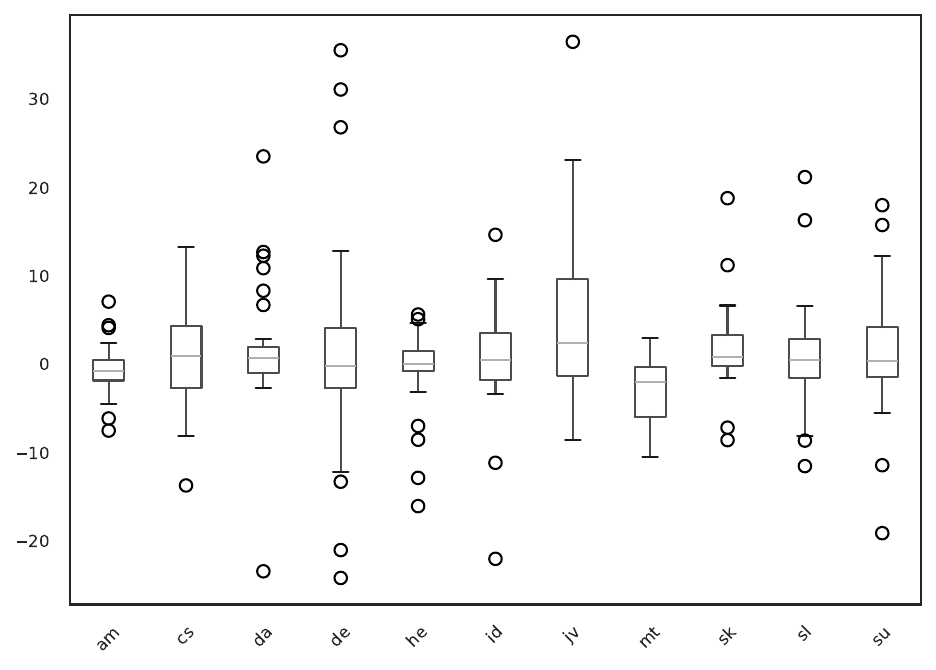} &
\includegraphics[width=0.18\textwidth]{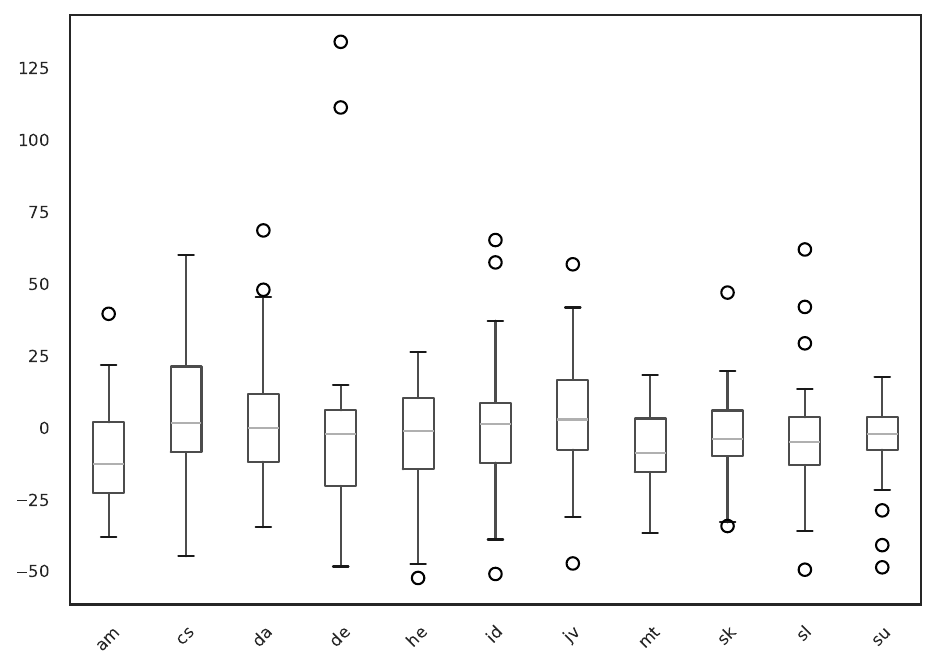}\\

Middle layers  (qual.) &
\includegraphics[width=0.18\textwidth]{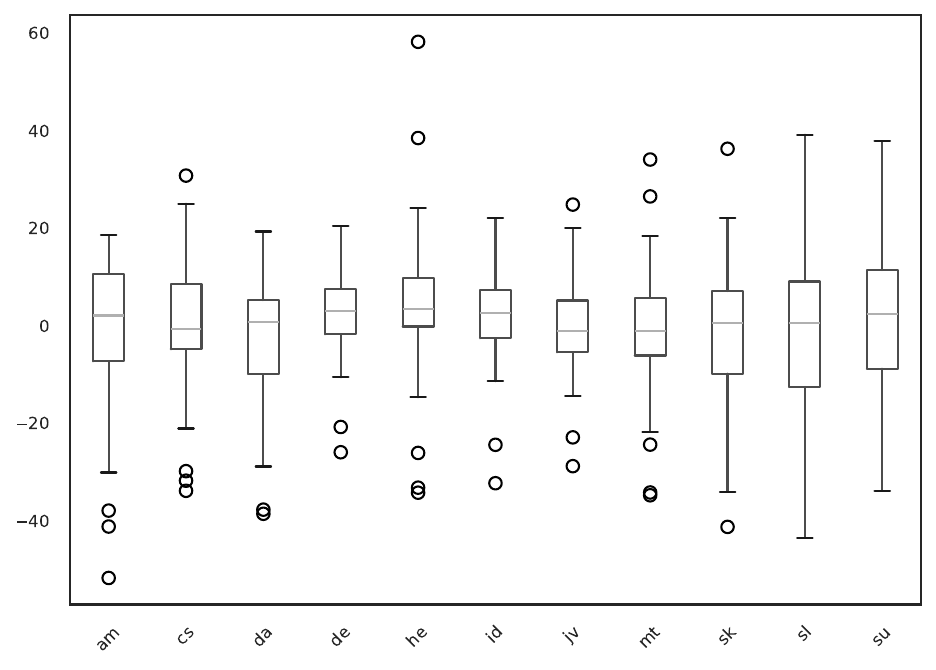} &
\includegraphics[width=0.18\textwidth]{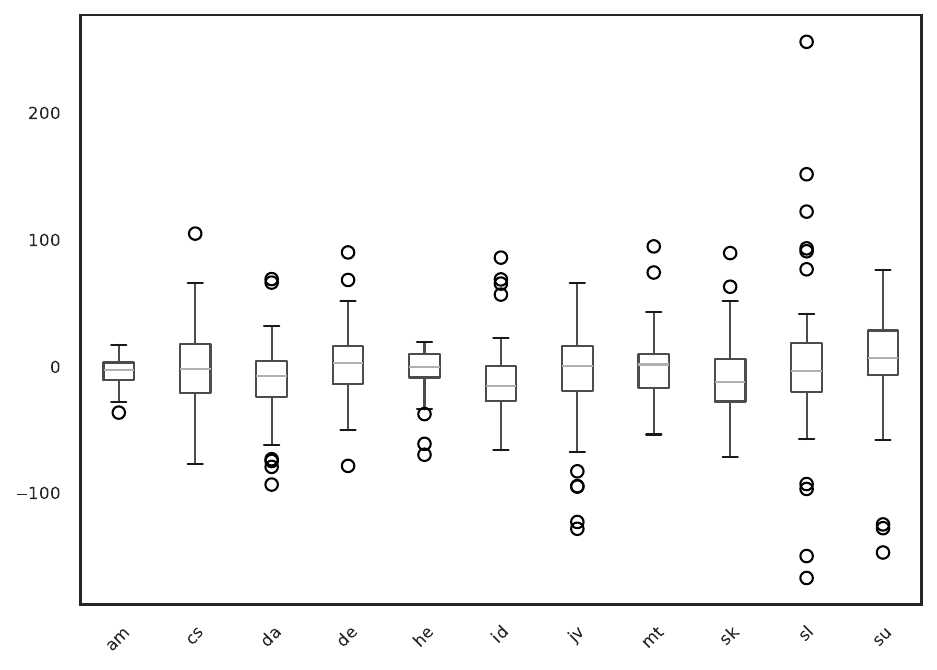} &
\includegraphics[width=0.18\textwidth]{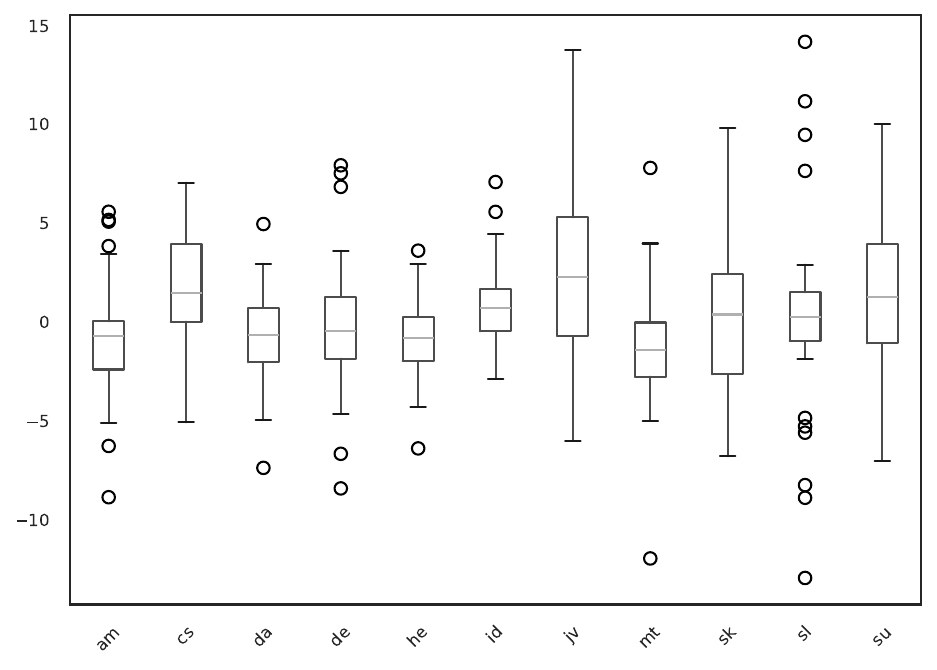} &
\includegraphics[width=0.18\textwidth]{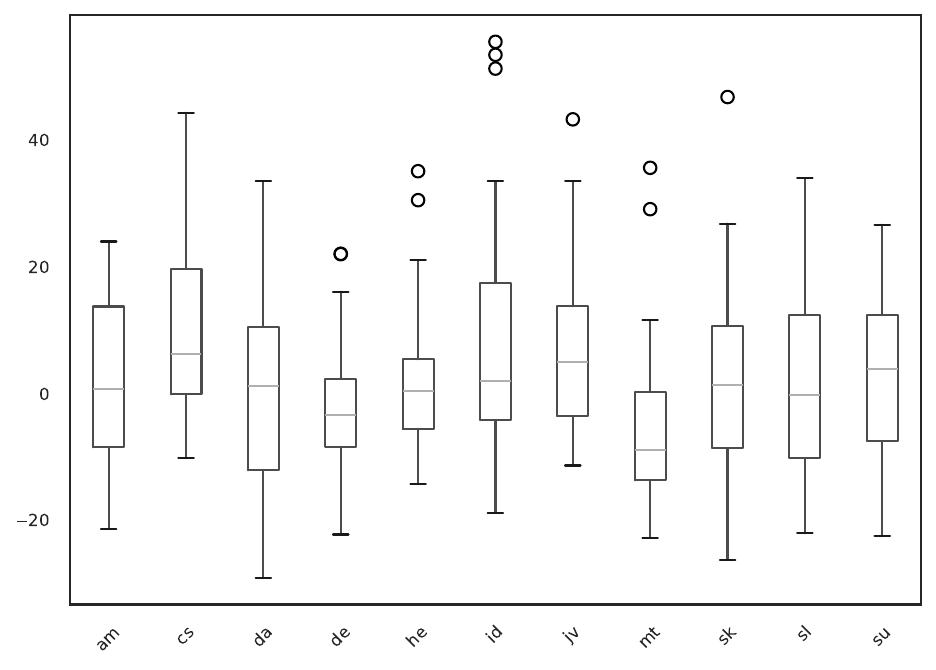}\\
\bottomrule
\end{tabular}
\end{table*}

\end{document}